\documentclass[11pt]{article}
\usepackage[preprint]{acl}    \usepackage{times}
\usepackage{latexsym}
\usepackage[T1]{fontenc}
\usepackage[utf8]{inputenc}
\usepackage{microtype}

\newif\ifJOURNAL  \JOURNALfalse
\newcommand{\journalonly}[1]{\ifJOURNAL #1\fi}      \newcommand{\aclonly}[1]{\ifJOURNAL\else #1\fi}     \newcommand{\shortlong}[2]{\ifJOURNAL #2\else #1\fi}

\usepackage{booktabs}
\usepackage{amssymb}
\usepackage{array}
\usepackage{longtable}  \newcolumntype{L}[1]{>{\raggedright\arraybackslash\hyphenpenalty=10000\exhyphenpenalty=10000}p{#1}}

\usepackage{newunicodechar}
\newunicodechar{—}{\textemdash}
\newunicodechar{–}{\textendash}
\newunicodechar{…}{\ensuremath{\dots}}
\newunicodechar{→}{\ensuremath{\rightarrow}}
\newunicodechar{←}{\ensuremath{\leftarrow}}
\newunicodechar{×}{\ensuremath{\times}}
\newunicodechar{÷}{\ensuremath{\div}}
\newunicodechar{≈}{\ensuremath{\approx}}
\newunicodechar{≥}{\ensuremath{\geq}}
\newunicodechar{≤}{\ensuremath{\leq}}
\newunicodechar{±}{\ensuremath{\pm}}
\newunicodechar{∼}{\ensuremath{\sim}}
\newunicodechar{κ}{\ensuremath{\kappa}}
\newunicodechar{α}{\ensuremath{\alpha}}
\newunicodechar{β}{\ensuremath{\beta}}
\newunicodechar{Δ}{\ensuremath{\Delta}}
\newunicodechar{°}{\ensuremath{^\circ}}
\newunicodechar{’}{'}
\newunicodechar{‘}{`}
\newunicodechar{“}{``}
\newunicodechar{”}{''}

\usepackage{enumitem}

\usepackage{amsmath}
\usepackage[capitalise,nameinlink]{cleveref}

\usepackage{graphicx}
\usepackage{xcolor}
\usepackage[most]{tcolorbox}

\usepackage{morefloats}
\usepackage{float}     \usepackage{seqsplit}  \usepackage{xurl}      

\newtcolorbox{examplebox}[1][]{
  breakable,
  enhanced,
  boxrule=0.4pt,
  colback=black!2,
  colframe=black!55,
  fonttitle=\bfseries\small,
  fontupper=\small,
  left=4pt,right=4pt,top=3pt,bottom=3pt,
  title={#1}
}

\newcommand{\sk}[1]{\texttt{\small #1}}
\newcommand{\tw}[1]{\textbf{#1}}

\newcommand{\lexen}{lexEN}
\newcommand{\sensebench}{SenseBench}
\newcommand{\maru}{Maru2022}
\newcommand{\wn}{WordNet}
\newcommand{\glite}{\shortlong{CoarseWN}{Glite}}
\newcommand{\lens}{\shortlong{LENS}{Glite LENS}}
\newcommand{\limref}{\hyperref[sec:limitations]{the Limitations section}}

 \JOURNALtrue                   

\title{English Word Sense Disambiguation in 2026: When the Labels Become the Bottleneck}
\author{
  \textbf{Vassili Philippov}$^{1}$ \quad \textbf{Amro Salman}$^{1}$ \quad \textbf{Dmitrii Andreev}$^{1}$ \quad \textbf{Penny Hands}$^{2}$ \\
  \textbf{Emil Kaiumov}$^{1}$ \quad \textbf{Pavel Katunin}$^{1}$ \quad \textbf{Anton Nikolaev}$^{3}$ \\[3pt]
  {\normalfont\small $^{1}$Glite, \texttt{\{vassili,amro,dmitrii,emil,pavel\}@glite.ai}} \\[1pt]
  {\normalfont\small $^{2}$Penny Hands Editorial Services} \\[1pt]
  {\normalfont\small $^{3}$School of Biosciences, The University of Sheffield, UK, \texttt{a.nikolaev@sheffield.ac.uk}}
}

\begin{document}
\maketitle

\begin{abstract}
\noindent In English all-words word sense disambiguation (WSD), the labels, not the models, have
become the bottleneck: frontier LLMs are accurate enough that the errors surviving in the gold
standard decide benchmark rankings---in the test sets we score on and, as we show causally, in the
corpus we train on. We release \lexen{}, a WSD evaluation benchmark built as a conservative,
human-adjudicated correction layer over \maru{}'s \textsc{all\_new} benchmark (211 labels changed,
56 removed), and \sensebench{}, an auditable LLM
WSD evaluation harness and living leaderboard (57 models, 192 runs). The task is
inventory-constrained multiple choice (the model picks from the supplied WordNet senses), so the
reported accuracies are a ceiling on what models achieve without that help. On \lexen{}-v1 the
frontier LLMs converge near 95\% (best, 95.6\%), the top three families are statistically
indistinguishable, and accuracy trades off against reasoning effort and cost across a
${\sim}2{,}500\times$ price span. Relabeling SemCor with frontier models and retraining BEM, ESCHER,
and ConSeC unchanged lifts them by several F1 points on test sets the relabeling never touched; we
release the relabeled corpora and \lens{}, a 298M bi-encoder trained on the
repaired labels---to our knowledge the strongest reported (83.6 Raganato \textsc{all}, 87.4
\maru{} \textsc{all\_new})---serving at ${\sim}\$0.13$ per million items. On hard items, fine-grained WordNet senses are
partly ill-posed even for experts (three-reviewer Fleiss $\kappa=0.537$); coarsening raises annotator
agreement and model accuracy together across four inventories, placing a top model inside the expert
agreement band at coarse granularity (statistically equivalent under three of four) but significantly
below it at fine. The binding constraint is now cost.
\end{abstract}

\section{Introduction}\label{sec:intro}

\shortlong{Word sense disambiguation (WSD) asks a system to pick, for a word in context, the intended sense
from a fixed inventory \citep{navigli2009}---a clean diagnostic of lexical semantics, scored against a
sense inventory built by lexicographers. For most of the last decade English all-words WSD looked
stubbornly hard. On the unified Raganato framework \citep{raganato2017}, gloss-aware and
knowledge-infused supervised systems \citep{huang2019,blevins2020,bevilacqua2020,barba2021a,barba2021}
climbed slowly and then plateaued in a narrow band around 79--83\% on the \textsc{ALL} test set,
close enough that architecture, not the data, appeared to be the limit.}{Word sense disambiguation (WSD) asks a system to pick, for a word in context, the intended sense
from a fixed inventory \citep{navigli2009}. It is a clean diagnostic of lexical semantics: it
isolates the one decision that distinguishes \emph{bank} the riverside from \emph{bank} the
institution, and it scores that decision against a sense inventory built by lexicographers. For most
of the last decade English all-words WSD looked stubbornly hard. On the unified Raganato evaluation framework
\citep{raganato2017}, gloss-aware and knowledge-infused supervised systems
\citep{huang2019,blevins2020,bevilacqua2020,barba2021a,barba2021} climbed slowly and then plateaued:
the strongest published numbers on the Raganato \textsc{ALL} test set sat in a narrow band around
79--83\% for years, close enough together that architecture, not the data, appeared to be the limit.
\journalonly{The field treated the gold labels as ground truth and competed on models. That
assumption was reasonable while systems were far enough below the labels that their errors swamped
any errors in the labels.}}

\shortlong{As frontier language models approached and then matched the supervised band (GPT-4o ties ConSeC at ${\sim}$83
F1; \citealp{meconi2025}), the few percent of mislabeled gold items stopped being background noise and
started deciding the ranking---and, as we show, on \lexen{}-v1 the frontier clusters near 95\%.
A model that ``misses'' a
mislabeled item is penalized for being right, and two models a fraction of a point apart can swap
places on annotation errors neither made \citep{maru2022}. This is not WSD-specific---pervasive label
errors destabilize benchmark rankings once models near the label ceiling \citep{northcutt2021}---and
the natural response, measuring at a reproducible granularity, has a long WSD history
\citep{palmer2007,navigli2007}. \maru{} re-annotated the hard core of Raganato, shifting accuracy by
several points; but a regime where a few labels decide the outcome demands verification an
inexpensive pass does not reach. To rank systems that agree to within a point, each corrected label
must be defensible item by item: who changed it, on what evidence, with how much expert agreement.}{As frontier language models approached and then matched the supervised band (GPT-4o ties ConSeC at ${\sim}$83
F1; \citealp{meconi2025}), the few percent of mislabeled gold items stopped being background noise and
started deciding the ranking---and, as we show, on \lexen{}-v1 the frontier clusters near 95\%.
A model that ``misses'' a mislabeled item is penalized for being right, and two models a fraction of a
point apart can swap places entirely on the strength of annotation errors neither of them made
\citep{maru2022}. This is not a WSD peculiarity: pervasive label errors have been shown to
destabilize the rankings of widely used machine-learning benchmarks once models approach the label
ceiling \citep{northcutt2021}, and the natural response---measuring at a granularity experts can
reproduce---has a long history in WSD, from OntoNotes sense grouping to the SemEval coarse-grained
all-words task \citep{palmer2007,navigli2007}. \maru{} recognized this and re-annotated the hard
core of the Raganato data, which
shifts measured accuracy by several points. But correcting labels for a regime in which a few labels
decide the outcome demands a level of verification that an inexpensive re-annotation pass does not
reach. To rank systems that already agree to within a point, the corrected layer must itself be
defensible item by item: who changed each label, on what evidence, and with how much agreement among
independent experts.}
\shortlong{What the field lacks is an auditable correction layer with full
lineage, an auditable harness that current models can be measured on, and baselines from the systems
people actually run today.}{What the field lacks, in short, is three things at once: an auditable
correction layer with full lineage rather than a single re-annotation pass; an auditable evaluation
harness on which today's rapidly changing proprietary and open models can be measured and
re-measured; and a current set of baselines spanning both the new frontier systems and the classic
supervised ones, scored under identical conditions.}

\journalonly{The stakes of getting this right are not academic. When the measurement instrument is noisier than
the differences it is asked to resolve, leaderboards reward annotation idiosyncrasies, ablations
chase artifacts, and the field's sense of progress detaches from anything a downstream user would
notice. The remedy is not a better model. It is a better-characterized benchmark: one that says
explicitly which items are decidable, how much experts agree on the rest, and how far apart the
leading systems really are once the labels are trustworthy.}

\shortlong{This paper supplies those three things and asks a sharper question than ``what is the best model?'':
where the remaining error actually lives. Frontier families cluster near 95\% on
\lexen{}-v1 (best, 95.6\%; an inventory-constrained multiple-choice task, selecting among the
supplied WordNet senses rather than open generation); the top three families are statistically
indistinguishable, on the full set as on the far smaller hard-reviewed subset. WSD is still not
solved---at fine granularity neither models nor expert lexicographers agree on the hard items, so much
of the residual disagreement coincides with sense distinctions the inventory draws more finely than
its annotators reproduce. We make six contributions.}{This paper supplies those three things and uses them to answer a sharper question than ``what is the
best model?'' We ask where the remaining error actually lives. The answer reframes the headline
number: frontier families cluster near 95\% on \lexen{}-v1 (best, 95.6\%; an
inventory-constrained multiple-choice task, not open generation), and the top three families are
statistically indistinguishable under paired tests --- on the full set as on the far smaller
hard-reviewed subset. WSD is still not solved, because at fine granularity neither models nor expert
lexicographers agree on the hard items, and much of the residual disagreement coincides with sense
distinctions the inventory draws more finely than its own annotators reproduce. We make six contributions.}

\begin{enumerate}[leftmargin=1.4em,itemsep=2pt,topsep=2pt]
  \item \textbf{\lexen{}}, a conservative, auditable correction and coarsening layer over
  \maru{}. \shortlong{A model panel selects suspicious contested items; three lexicographers review
  them independently; a frozen two-of-three rule adjudicates. It changes 211 labels and removes 56
  unanswerable items, with full lineage, a coarse-sense mapping}{A model panel selects a hard, suspicious tail; three independent lexicographers
  review the selected items independently; a frozen two-of-three rule adjudicates. The result changes 211 labels and
  removes 56 unanswerable items, with full lineage, a coarse-sense mapping for granularity analysis} (\Cref{sec:lexen}).
  \item \textbf{\sensebench{}}, an auditable LLM WSD evaluation harness: immutable registered prompts, raw
  call artifacts re-verified in continuous integration, per-item cost tracking, bootstrap confidence
  intervals, and a living public leaderboard (\Cref{sec:sensebench}).
  \item \textbf{A measurement study} of the current frontier on WSD: \shortlong{label-noise impact on
  rankings, cross-family agreement on the corrected tail, reasoning effort as a first-class
  axis, and a cost/accuracy Pareto spanning $>{}1{,}200\times$ in the high-accuracy slice}{the impact of label noise
  on rankings, cross-family agreement on the corrected tail, reasoning effort as a
  first-class axis (comparisons are only fair within a tier), and a cost/accuracy Pareto
  spanning more than $1{,}200\times$ within the high-accuracy slice} (\Cref{sec:results-noise}).
  \item \textbf{A controlled training-label intervention} (\Cref{sec:results-training}): relabeling the
  SemCor training corpus with frontier models and retraining the classic supervised systems (BEM,
  ESCHER, ConSeC) with no architectural change lifts them by several F1 points on test sets the
  relabeling never touched---the controlled causal complement to the observational measurement study
  above, and a route to distilling frontier-quality labels into cheap models. We release both
  relabeled corpora, \textbf{SemCor-GPT5.5} and \textbf{SemCor-Gemma}.
  \item \textbf{\lens{}, a released state-of-the-art bi-encoder} (\Cref{sec:rt-encoder}): a modern
  298M dual encoder built from standard components and trained solely on the relabeled corpus.
  It reaches 83.6 on Raganato \textsc{all} and 87.4 on Maru
  \textsc{all\_new}---to our knowledge the strongest bi-encoder reported, above every published
  SemCor-only supervised system and level with ConSeC+WNGE, which trains on ${\sim}3\times$ the
  labelled data---with 90.5 on
  \lexen{}-v1 itself (confirmatory, since the training labels share a model with the relabeling),
  and serves at roughly \$0.13 per million items. The accompanying labels$\times$architecture
  decomposition shows the two axes contribute comparably and compose; we release code and weights.
  \item \textbf{A granularity and human-ceiling analysis} \shortlong{relating the residual
  disagreement to sense granularity and inter-annotator agreement, and identifying cost as
  the next binding constraint}{that relates the residual disagreement to sense granularity and to
  inter-annotator agreement, and identifies cost as
  the next binding constraint} (\Cref{sec:results-ceiling}).
\end{enumerate}

\shortlong{Our thesis is deliberately careful. On the hardest items, the best models sit just below the expert
reviewers at fine WordNet granularity; at the coarse, practically meaningful granularity that gap is no
longer statistically detectable, and coarse-graining raises both inter-annotator agreement and model
accuracy robustly across every sense inventory we test. At the fine level the residual disagreement is
shared by humans and models alike, because the inventory over-specifies distinctions competent readers
do not reliably reproduce. Benchmark quality --- the labels and the sense granularity --- has become the
binding constraint on measurement, and we provide an auditable layer built to that standard. Where that constraint is label noise it is repairable---correcting the training labels lifts even the classic supervised systems (\Cref{sec:results-training})---and where it is sense granularity it is shared by models and experts alike.}{Our thesis is deliberately careful. On the hardest items, the best models sit just below the expert
reviewers at fine WordNet granularity; at the coarse, practically meaningful granularity that gap is no
longer statistically detectable under our inventory or the public ones, and coarse-graining raises both
inter-annotator agreement and model accuracy robustly across every sense inventory we test. At the fine
level the residual disagreement is shared by humans and models alike, because the inventory
over-specifies distinctions that competent readers do not reliably reproduce. Benchmark quality --- the
labels and the sense granularity at which the task is scored --- has become the binding constraint on
measurement, and we provide an auditable layer built to that standard. Where that constraint is label noise it is repairable---correcting the training labels lifts even the classic supervised systems (\Cref{sec:results-training})---and where it is sense granularity it is shared by models and experts alike.}

\shortlong{%
The motivation is applied as well as scientific: this work grew out of building a concept-based
English dictionary for language learners (at a company named in the de-anonymized version), where
each sense---not each word---is the unit of teaching and of learner-knowledge estimation, and where
authentic media must be sense-tagged at production scale; that is why cost per million items is a
first-class axis throughout.%
}{%
The motivation is applied as well as scientific: this work grew out of building a concept-based
English dictionary for language learners at Glite, where each sense---not each word---is the unit of
teaching and of learner-knowledge estimation, and where authentic media (film and video subtitles)
must be sense-tagged at production scale. That is why cost per million items is a first-class axis
throughout, and why the \glite{} coarse inventory of \Cref{sec:lexen} exists.}

\begin{figure*}[t]
\centering
\includegraphics[width=\shortlong{\linewidth}{0.85\linewidth}]{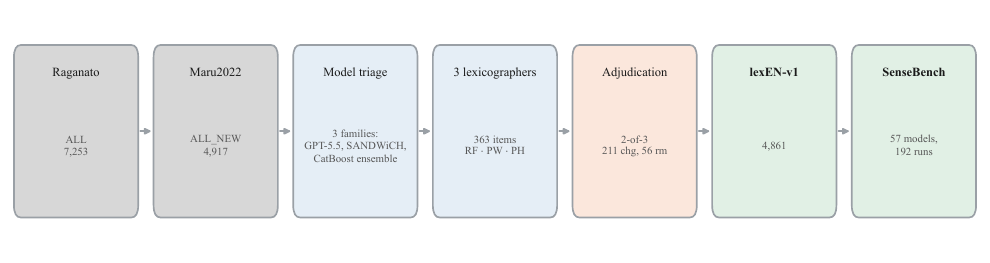}
\caption{The \lexen{} construction and \sensebench{} evaluation pipeline. Raganato \textsc{ALL}
\citep{raganato2017} is re-annotated by \maru{} \citep{maru2022}; a model panel selects
contested items; three lexicographers review them independently; a frozen two-of-three rule yields
\lexen{}-v1, an auditable human-adjudicated correction layer over the model-flagged tail, which \sensebench{} scores under immutable prompts with full artifact
provenance.}
\label{fig:f1}
\end{figure*}
 \section{Related Work}\label{sec:related}

The three gaps \S\ref{sec:intro} identifies---an auditable correction layer, an
auditable harness, and current frontier baselines---each have a literature; we
organize prior work by the role it plays here rather than
chronologically: the evaluation framework we correct (\S\ref{sec:rel-eval}), the
specialized systems we benchmark against (\S\ref{sec:rel-systems}), the LLM-WSD
studies our leaderboard subsumes (\S\ref{sec:rel-llm}), the sense-granularity
work that motivates our coarse layer (\S\ref{sec:rel-gran}), and the
annotation-quality and governance work that motivates a verified, auditable
benchmark (\S\ref{sec:rel-gov}).\journalonly{ Table~\ref{tab:t11} summarizes each line and the
role it serves here.}

\journalonly{\begin{table*}[t]
\centering
\small
\caption{Related work by role in this paper.}
\label{tab:t11}
\begin{tabular}{L{0.26\linewidth}lL{0.56\linewidth}}
\toprule
\textbf{Work} & \textbf{Year} & \textbf{Role} \\
\midrule
Raganato et al. & 2017 & unified English all-words WSD framework (the labels we correct) \\
Maru et al. & 2022 & ALL\_NEW corrected base + hardEN; our starting point \\
Huang et al. (GlossBERT) / Blevins (BEM) & 2019 / 2020 & gloss-aware supervised baselines \\
Barba et al. (ESCHER, ConSeC) & 2021 & comprehension-based supervised SOTA; our baselines \\
Kocon et al. & 2023 & GPT-3.5 73.3 Macro-F1 on Raganato ALL \\
Yae et al. & 2025 & GPT-4 70.4\% MC; BabelNet \\
Basile et al. & 2025 & zero-shot LLMs + FT Llama-3.1-8B 86.5 (XL-WSD) \\
Meconi et al. & 2025 & GPT-4o 83.2 on Maru2022; most comprehensive prior LLM-WSD \\
Navigli & 2026 & survey: WSD redefined, not dead \\
Lacerra et al. (CSI) / Kikuchi & 2020/24 & coarse sense inventories (motivates Glite) \\
Ng / Murray & 1999/2004 & WSD inter-annotator agreement \\
\bottomrule
\end{tabular}
\end{table*}
}

\subsection{Unified WSD evaluation and its corrections}\label{sec:rel-eval}

\shortlong{English all-words WSD is standardly evaluated on the framework of
\citet{raganato2017}, which unifies five Senseval/SemEval test sets against
WordNet~3.0 \citep{miller1995,fellbaum1998}. \citet{maru2022} showed it still
contains non-trivial annotation errors and released a manually refined version
(ALL\_NEW) with a challenge subset (hardEN) all supervised systems miss. \lexen{}
takes ALL\_NEW as its starting point and adds professional lexicographer review
(\S\ref{sec:lexen}), inheriting the Raganato/Maru lineage while arguing that, in
the frontier-LLM regime, even Maru's corrections leave residual label noise that
distorts rankings.}{English all-words WSD is standardly evaluated on the framework of
\citet{raganato2017}, which unifies five Senseval/SemEval test sets against the
WordNet~3.0 sense inventory \citep{miller1995,fellbaum1998}. \citet{maru2022}
showed that this framework still contains a non-trivial number of annotation
errors, and released a manually refined version (ALL\_NEW) together with a
challenge subset (hardEN) of instances that all supervised systems miss. \lexen{}
takes ALL\_NEW as its starting point and adds a further layer of professional
lexicographer review (\S\ref{sec:lexen}); we thus inherit the Raganato/Maru
lineage while arguing that, in the frontier-LLM regime, even Maru's corrections
leave residual label noise that materially distorts rankings.}

\subsection{Specialized WSD systems}\label{sec:rel-systems}

\shortlong{A decade of supervised, gloss-aware systems raised English all-words F1 from the
65--70 range toward the low 80s: GlossBERT \citep{huang2019}, sense embeddings
(LMMS) \citep{loureiro2019}, the bi-encoder BEM \citep{blevins2020}, the
knowledge-injecting EWISER that first broke the 80\% ``glass ceiling''
\citep{bevilacqua2020}, and the comprehension-based ESC and ConSeC
\citep{barba2021a,barba2021}; the cross-encoder SANDWiCH reports a new supervised
state of the art \citep{guzmanolivares2025}. We score MFS, BEM, ESCHER, and ConSeC
as reference baselines on the same items (\S\ref{sec:results-noise}), and return to
the bi-encoder design in \Cref{sec:rt-encoder}, where a modernized descendant of BEM
trained on relabeled SemCor becomes the strongest member of the family. The recurring
plateau near 79--83\% fine-grained F1 frames our central question: is the remaining
error in the models or in the labels?}{A decade of supervised, gloss-aware systems steadily raised English all-words F1
from the 65--70 range toward the low 80s: GlossBERT \citep{huang2019}, sense
embeddings such as LMMS \citep{loureiro2019}, the gloss-informed bi-encoder BEM
\citep{blevins2020}, the knowledge-injecting EWISER that first broke the 80\%
``glass ceiling'' \citep{bevilacqua2020}, and the comprehension-based ESC and
ConSeC \citep{barba2021a,barba2021}. More recently, the cross-encoder SANDWiCH
reports a new supervised state of the art \citep{guzmanolivares2025}. We use MFS,
BEM, ESCHER, and ConSeC as reference baselines scored on the same items
(\S\ref{sec:results-noise}), and treat these systems as the point of comparison
against which LLM performance and label-noise sensitivity are measured; we return
to the bi-encoder design in \Cref{sec:rt-encoder}, where a modernized descendant of
BEM trained on relabeled SemCor becomes the strongest member of the family. A
recurring observation in this line of work---that progress had plateaued near
79--83\% fine-grained F1---frames our central question: whether the remaining
error is in the models or in the labels.}

\subsection{LLMs for WSD}\label{sec:rel-llm}

\shortlong{A growing body of work evaluates instruction-tuned LLMs on WSD, but, as we
document in \S\ref{sec:results-noise}, these studies are mutually
\emph{incomparable} along five axes: sense inventory, dataset/subset, task
format, metric, and model version. Reported points range from GPT-3.5 at 73.3
macro-F1 \citep{kocon2023} and GPT-4 at 70.4\% accuracy \citep{yae2025} to a
fine-tuned Llama-3.1-8B at 86.5 F1 on a BabelNet-based XL-WSD reformulation
\citep{basile2025}. The most comprehensive study, \citet{meconi2025}, finds GPT-4o
and DeepSeek-V3 match specialized systems on Maru's ALL\_NEW (GPT-4o few-shot 83.2 vs.\
ConSeC 83.0 F1) while staying far more robust on hardEN and cross-domain data, yet
still trailing a human expert (82.5 vs.\ 91.25 F1, both on a random 400-item subset of the same
benchmark); they also report ${\sim}5$~F1
prompt sensitivity and positional bias. Prompt-engineering (GlossGPT)
\citep{sumanathilaka2025,sumanathilaka2024}, LLM distillation \citep{ming2025},
and robustness probing \citep{zhang2025} round out the line; \citet{navigli2026}
argues WSD is not obsolete but ``redefined'' as a diagnostic of lexical
competence.}{A growing body of work evaluates instruction-tuned LLMs on WSD, but, as we
document quantitatively in \S\ref{sec:results-noise}, these studies are mutually
\emph{incomparable} along five axes: sense inventory (WordNet vs.\ BabelNet vs.\
Wiktionary/FEWS), dataset and subset (full Raganato ALL including SemEval-2007
vs.\ Maru's ALL\_NEW vs.\ XL-WSD vs.\ single subsets), task format (definition
selection vs.\ index multiple-choice vs.\ free generation), metric (micro-F1 vs.\
macro-F1 vs.\ accuracy), and model version. \citet{kocon2023} report GPT-3.5 at
73.3 macro-F1 on the full Raganato framework; \citet{yae2025} report GPT-4 at
70.4\% accuracy in a multiple-choice setting; \citet{basile2025} evaluate open
LLMs zero-shot on a BabelNet-based XL-WSD reformulation and show that a
fine-tuned Llama-3.1-8B reaches 86.5 F1 on English. The most comprehensive study,
\citet{meconi2025}, finds that GPT-4o and DeepSeek-V3 match specialized systems
on Maru's ALL\_NEW (GPT-4o few-shot 83.2 vs.\ ConSeC 83.0 F1) while remaining far more
robust on hardEN and cross-domain data, yet still trailing a human expert---on a random 400-item subset of the same benchmark, GPT-4o
scores 82.5 against the expert's 91.25 F1; they also report substantial prompt sensitivity (a
${\sim}5$~F1 swing across templates) and positional bias from definition
ordering. Prompt-engineering approaches such as GlossGPT add few-shot
chain-of-thought and gloss retrieval \citep{sumanathilaka2025,sumanathilaka2024},
and LLMs have been used as teachers for distillation \citep{ming2025} and probed
for robustness \citep{zhang2025}. Synthesizing these results, \citet{navigli2026}
argues that WSD is not obsolete but ``redefined'' as a diagnostic lens on lexical
competence.}

\shortlong{Two gaps motivate our work. First, this literature's knowledge of \emph{frontier}
LLM behavior is effectively frozen at GPT-4o and DeepSeek-V3 (early 2025): to our
knowledge no prior study reports a standard-benchmark WSD result for the Claude,
Gemini, GPT-4.1/5.x, DeepSeek-V4, Qwen3, GLM-5, Kimi, or Llama-4 families. Second,
no shared protocol exists, so the numbers cannot be placed on a common scale---a
problem compounded by known LLM sensitivities to option order \citep{zheng2023}
and prompt formatting \citep{he2024b,zhuo2024}. \sensebench{} addresses both: it
evaluates 57 models across a 192-run public leaderboard,
under an immutable prompt registry---the leaderboard is reported per registered
prompt---with raw-artifact re-verification, and, because it re-scores the same
predictions against each gold layer (Raganato, \maru{}, and \lexen{}), separates
label-quality from model gains (\S\ref{sec:results-noise}). We now build the corrected layer
those gaps demand.}{Two gaps motivate our work. First, this literature's knowledge of \emph{frontier}
LLM behavior on WSD is effectively frozen at GPT-4o and DeepSeek-V3 (early 2025):
to our knowledge no prior study reports a standard-benchmark WSD result for the
Claude, Gemini, GPT-4.1/5.x, DeepSeek-V4, Qwen3, GLM-5, Kimi, or Llama-4
families. Second, no shared, reproducible protocol exists, so the reported
numbers cannot be placed on a common scale---a problem compounded by known LLM
sensitivities to multiple-choice option order \citep{zheng2023} and prompt
formatting \citep{he2024b,zhuo2024}. \sensebench{} addresses both: it evaluates
57 models across a 192-run public leaderboard, under an
immutable prompt registry---each registered prompt yields its own comparable
leaderboard---with raw-artifact re-verification, and, because it re-scores the
same predictions against each gold layer in turn (Raganato, \maru{}, and
\lexen{}), it lets us separate label-quality gains from model gains (\S\ref{sec:results-noise}). We now build the
corrected layer that those gaps demand.}

\subsection{Sense granularity and coarse inventories}\label{sec:rel-gran}

\shortlong{WordNet's fine-grained inventory is widely held to over-specify distinctions that
neither systems nor humans reliably reproduce \citep{kilgarriff1997,hanks2000}, and
granularity-aware evaluation was proposed as early as \citet{resnik1999}; coarser, more
reproducible inventories have been proposed to mitigate this \citep{ide2006}, from automatic
sense clustering and merging \citep{navigli2006,snow2007} and the OntoNotes coarse-sense work
\citep{hovy2006,palmer2007,zhong2008} to the SemEval-2007 coarse-grained task \citep{navigli2007} and later
coarse inventories \citep{lacerra2020,kikuchi2024}; the graded-annotation line responds
to the same unreliability by replacing discrete assignment with graded judgments
\citep{erk2013,mccarthy2016}. Closest to our granularity analysis,
\citet{loureiro2021} coarsen the Raganato benchmark with CSI domain labels and find that BERT
approaches the coarse human ceiling on curated nouns, arguing WSD is granularity-relative rather than
solved; we extend this to frontier LLMs on \emph{corrected} all-words labels, quantify the expert
ceiling with professional inter-annotator agreement, and carry the same coarsening logic to the
training corpus (\S\ref{sec:results-training}). Our \glite{} coarsening layer
(\S\ref{sec:lexen}) is in this tradition, and \S\ref{sec:results-ceiling}
quantifies how much residual fine-grained disagreement dissolves under coarsening.}{}\journalonly{WordNet's fine-grained inventory is widely held to over-specify distinctions that
neither systems nor humans reliably reproduce---a position stated most sharply by
lexicographers themselves \citep{kilgarriff1997,hanks2000} and anticipated on the evaluation
side by \citet{resnik1999}, who argued that WSD scoring should weight sense distinctions rather
than treat every fine split as equally real. Two responses developed. The graded-annotation line
replaces discrete sense assignment with graded judgments, finding annotator behavior graded
rather than categorical \citep{erk2013}, with clusterability itself varying by lemma
\citep{mccarthy2016}. The other response---the one this paper takes---moves to coarser, more
reproducible sense distinctions, the practical target articulated by \citet{ide2006}: automatic
sense clustering and learned sense merging \citep{navigli2006,snow2007}; the OntoNotes grouping
effort \citep{hovy2006}, the manual and automatic coarse sense distinctions of \citet{palmer2007},
and the empirical WSD study of those groupings \citep{zhong2008}, which reported the agreement
gains they bring; and the SemEval-2007 coarse-grained all-words task, which standardized
evaluation at that granularity \citep{navigli2007}. More recent coarse inventories continue this program,
including the Coarse Sense Inventory of \citet{lacerra2020} (which targets
${\sim}85\%$-accuracy WSD) and dictionary-derived coarse inventories
\citep{kikuchi2024}. Closest to our own granularity analysis is \citet{loureiro2021}, who coarsen the
unified Raganato benchmark using the CSI domain labels of \citet{lacerra2020} and show that BERT
approaches the coarse-grained human ceiling on curated noun sets under ideal training data,
concluding that ``solved'' is granularity-relative and that WSD is not solved even at the coarse
level. We reach a convergent conclusion from a different vantage point: we measure \emph{frontier}
LLMs on \emph{lexicographer-corrected} all-words labels, quantify the human ceiling with three
professional annotators' inter-annotator agreement rather than a small estimate, and show that the
same coarsening logic, applied to the \emph{training} corpus, repairs the classic supervised systems
(\S\ref{sec:results-training}). Our \glite{} coarsening layer (\S\ref{sec:lexen}) is in this
tradition; \S\ref{sec:results-ceiling} quantifies how much of the residual
fine-grained disagreement---between expert lexicographers as well as models---is
dissolved by coarsening.}

\subsection{Annotation quality, agreement, and governance}\label{sec:rel-gov}

\shortlong{Fine-grained sense disagreement among human annotators is a long-standing finding
\citep{ng1999,murray2004,passonneau2012}, and the broader measurement-quality literature shows
that the resulting label errors, once models near the label ceiling, can
reorder benchmark rankings outright \citep{northcutt2021}---which is why a
high-accuracy benchmark must report inter-annotator agreement and a human ceiling
rather than only a single asserted label. The human-label-variation literature reads such
disagreement as signal to model rather than noise to adjudicate away
\citep{pavlick2019,uma2021,basile2021disagreement,plank2022}; we correct the gold layer but
release every unaggregated reviewer choice, keeping disagreement-aware evaluation possible.
\lexen{} responds with independent
three-lexicographer review, full agreement statistics
(\S\ref{sec:results-ceiling}), and a release contamination canary.}{}\journalonly{That human annotators disagree on fine-grained senses is a long-standing finding
\citep{ng1999,murray2004,passonneau2012}, which is precisely why a high-accuracy benchmark must
report inter-annotator agreement and a human ceiling rather than a single gold
label. This concern is not specific to WSD: the measurement-quality literature
documents that pervasive label errors in standard test sets destabilize
machine-learning benchmarks once model accuracy approaches the noise floor of the
labels themselves \citep{northcutt2021}; WSD, where frontier models now exceed
nine items in ten, is squarely in that regime. The human-label-variation literature draws a
further lesson from such disagreement: that it is signal to be released and modeled rather than
noise to be adjudicated away \citep{pavlick2019,uma2021,basile2021disagreement,plank2022}.
\lexen{} takes the correction route for its gold layer---an audited benchmark needs a
defensible single label---but releases every unaggregated reviewer choice and rationale, so
disagreement-aware evaluation over the same items remains possible. Contamination of public benchmarks
is a further concern in the LLM era, prompting newly curated evaluation sets
\citep{meconi2025}. \lexen{} responds with independent three-lexicographer review,
full agreement statistics
(\S\ref{sec:results-ceiling}), a release contamination canary, and
\sensebench{}'s immutable prompts and re-verifiable raw artifacts
(\S\ref{sec:sensebench})---turning benchmark governance into a first-class
property of the resource.}

\shortlong{Model-assisted benchmark repair has close precedents in the annotation-error-detection line
\citep{klie2023aed}. \citet{alt-etal-2020-tacred} used model
disagreement to focus trained annotator effort on challenging TACRED examples, while
\citet{northcutt2021} used algorithmic flagging followed by human validation to show that label
errors can reorder benchmark rankings. ImageNet-ReaL \citep{beyer2020imagenet} and LLM-era audits
\citep{nahum-etal-2025-llms,gema2025mmlu} similarly use model proposals or ensembles to surface likely
label problems for human reassessment. \lexen{} follows this targeted-audit line: models concentrate
attention, and experts adjudicate.}{Model-assisted benchmark repair has close precedents, and its flagging step sits within the
annotation-error-detection literature surveyed by \citet{klie2023aed}. \citet{alt-etal-2020-tacred} used model
disagreement to focus trained annotator effort on challenging TACRED examples and included a control
sample---Re-TACRED later re-annotated the benchmark wholesale \citep{stoica2021retacred}, and
semi-automatic correction of CoNLL-2003 found test-set error rates comparable to state-of-the-art
model error \citep{reiss2020conll}---while \citet{northcutt2021} used algorithmic flagging followed by human validation to show
that label errors can reorder benchmark rankings. In vision, ImageNet-ReaL \citep{beyer2020imagenet}
used model-derived candidate labels followed by human reassessment; in the LLM era,
\citet{nahum-etal-2025-llms} use LLM ensembles to flag likely label errors and show that human
correction shifts reported performance, and expert re-annotation of MMLU reaches the same
conclusion at LLM-benchmark scale \citep{gema2025mmlu}. \lexen{} follows this targeted-audit line: models concentrate
attention on the suspicious tail, and experts adjudicate the retained labels.}

\shortlong{Our training-side intervention (\S\ref{sec:results-training}) connects two further lines. The
learning-with-noisy-labels literature builds models robust to corrupted \emph{training} annotations
\citep{song2023} and estimates which training labels are themselves wrong \citep{northcutt2021cl}; we take the complementary route of \emph{correcting} the labels at the source and
retraining unchanged systems. And a growing body of work uses LLMs as annotators, finding them
competitive with or better than crowd workers \citep{wang2021gpt3labeling,gilardi2023,ding2023,tornberg2023chatgpt4}; we apply that idea not to
fresh data but to re-annotating an established training corpus, measured through the downstream
supervised systems rather than against a reference annotation.}{Our training-side intervention (\S\ref{sec:results-training}) connects this paper to two further lines
of work. The first is learning with noisy labels, which develops architectures and losses robust to
corrupted \emph{training} annotations \citep{song2023}, and methods to estimate which training
labels are themselves wrong \citep{northcutt2021cl}; rather than tolerate the noise, we correct it
at the source and retrain otherwise-unchanged systems on the cleaned corpus. The second is the use of
LLMs as data annotators, where frontier models have been found competitive with or superior to crowd
workers across annotation tasks \citep{wang2021gpt3labeling,gilardi2023,ding2023,tornberg2023chatgpt4}. Most of that work labels \emph{new} data;
we instead re-annotate an established, widely used training corpus (SemCor) and evaluate the relabeling
not against a held-out reference annotation but through the accuracy of the supervised systems it
produces---an end-to-end test of whether LLM relabeling improves the training signal itself. Relatedly,
the weak-to-strong line asks when models trained on another model's labels can exceed their
supervision \citep{burns2024weaktostrong}; our design asks the complementary question---whether
repaired labels lift fixed students---and reads the answer off untouched test sets.}
 \section{\lexen{}: Constructing an Auditable Correction Layer}\label{sec:lexen}

A benchmark can only diagnose a model as far as its labels are trusted. In the near-saturation regime described in \Cref{sec:intro}, residual label error sets the ranking; \lexen{} is our response: a conservative, fully traced correction and coarsening layer over the standard English all-words evaluation set. It is built to be audited rather than believed. \shortlong{Every step --- which items were inspected, who inspected them, what each reviewer chose, and how disagreements resolved --- is recorded and released. The pipeline of Figure~\ref{fig:f1} runs source (Section~\ref{sec:lexen:source}) $\rightarrow$ triage (Section~\ref{sec:lexen:triage}) $\rightarrow$ review (Section~\ref{sec:lexen:review}) $\rightarrow$ adjudication (Section~\ref{sec:lexen:adjudication}) $\rightarrow$ coarse layer (Section~\ref{sec:lexen:glite}) $\rightarrow$ release (Section~\ref{sec:lexen:release}).}{Every step --- which items were inspected, who inspected them, what each reviewer chose, and how disagreements were resolved --- is recorded and released, so a skeptical reader can reconstruct any single label change. This section walks the construction pipeline of Figure~\ref{fig:f1} end to end: the source we start from (Section~\ref{sec:lexen:source}), the model-assisted triage that decides what to inspect (Section~\ref{sec:lexen:triage}), the lexicographer review (Section~\ref{sec:lexen:review}), the frozen adjudication rule that produces \lexen{}-v1 labels for reviewed retained items (Section~\ref{sec:lexen:adjudication}), the coarse-sense layer (Section~\ref{sec:lexen:glite}), and the release and governance terms (Section~\ref{sec:lexen:release}).} The headline counts are in Table~\ref{tab:t1}.

\subsection{Source}\label{sec:lexen:source}

\shortlong{\citet{raganato2017} unified the historical Senseval and SemEval all-words tasks into a single English evaluation set (ALL) under one WordNet inventory \citep{miller1995,fellbaum1998}. \citet{maru2022} released a manually corrected version, ALL\_NEW, fixing annotation and preprocessing errors and excluding SemEval-2007 and monosemous targets (4{,}917 of ALL's 7{,}253 instances); we take their release as our source layer. Building on the corrected set rather than Raganato ALL is deliberate: \lexen{} then measures the residual noise that survives a competent prior pass, not the easier errors \citet{maru2022} already caught.}{We start from the most carefully maintained version of the standard benchmark rather than the original. \citet{raganato2017} unified the historical Senseval and SemEval all-words tasks into a single English evaluation set (ALL) under one WordNet inventory \citep{miller1995,fellbaum1998}, and it has anchored the field for nearly a decade. \citet{maru2022} re-examined that set and released a manually corrected version, ALL\_NEW, fixing annotation and preprocessing errors; we take their 4{,}917-instance release as our source layer. Building on the corrected set, rather than on Raganato ALL directly, is a deliberate choice: it means \lexen{} measures the residual label noise that survives a competent prior correction pass, not the easier errors \citet{maru2022} already caught. \journalonly{Relative to ALL's 7{,}253 instances, ALL\_NEW excludes SemEval-2007 (conventionally the development set) and monosemous targets, fixes tokenization, lemmatization, and part-of-speech errors, and corrects the gold labels its reviewing linguist found inaccurate, leaving 4{,}917 polysemous instances across the four remaining test sets (Senseval-2 and -3, SemEval-2013 and -2015); we inherit the \maru{} sense keys verbatim and only ever replace, remove, or coarsen them.}} Throughout, an item is a single sense-annotated target token in its sentence context, and a label is the WordNet sense key \citep{miller1995} assigned to it.

\subsection{Model-Assisted Triage}\label{sec:lexen:triage}

\shortlong{Reviewing all 4{,}917 items with three lexicographers would be prohibitive, and most are uncontroversial; the question is where the residual errors are. We use a panel of models to rank items by how strongly automatic predictions disagree with the source label, then send only the most suspicious to human review. The panel draws on three model families: GPT-5.5; SANDWiCH, a distilled supervised WSD system, on its own \citep{guzmanolivares2025}; and a CatBoost ensemble, a learned per-item selector over five WSD systems (ConSeC, ESCHER, BEM, MFS, and GPT-5-mini). The GPT-5.5 family drives the ranking by how concentrated its disagreement with \maru{} is, while SANDWiCH and the CatBoost ensemble enter the modal vote; an item draws the strongest suspicion when the three families independently converge on the same alternative sense. This panel flagged \textbf{363} items for human review; the exact selection rule --- vote thresholds, suspicion-bucket waterfall, and the sensitivity of the yield to each threshold --- is documented, with a reference implementation, in the released \lexen{} repository.}{Reviewing all 4{,}917 items with three professional lexicographers would be prohibitively expensive, and most items are uncontroversial; the question is where the residual errors are. We use a panel of models to rank items by how strongly automatic predictions disagree with the source label, then send only the most suspicious items to human review. The panel draws on three model families: GPT-5.5; SANDWiCH, a distilled supervised WSD system, evaluated on its own \citep{guzmanolivares2025}; and a CatBoost ensemble, a learned per-item selector over five WSD systems (ConSeC, ESCHER, BEM, MFS, and GPT-5-mini). For each item we record whether each panel member agrees with the \maru{} label and, when it disagrees, which alternative sense it prefers. The GPT-5.5 family drives the ranking by how concentrated its disagreement with \maru{} is, while SANDWiCH and the CatBoost ensemble enter the modal vote; an item draws the strongest suspicion when the three families independently converge on the same alternative sense. This panel flagged \textbf{363} items for human review. The exact selection rule is not a tunable score but a fixed decision waterfall --- all eight GPT-5.5 variants vote with equal weight, an item is flagged when at least six of the eight dispute the \maru{} label, and SANDWiCH and CatBoost enter only to grade how strongly the flagged disagreement converges --- and it is documented in full, with every threshold, a reference implementation, and the sensitivity of the 363-item yield to each threshold (loosening the main vote threshold by one step adds 59 items; the loosest any-dissent pool is 867), in the released \lexen{} repository.}

\shortlong{The selection has a property worth stating exactly. Because the panel is built from frontier models, the reviewed subset is \emph{enriched} for items where strong models disagree with \maru{}, and is therefore \emph{not} a random sample of the benchmark. Three consequences follow. First, the corrections are concentrated in this suspicious tail, not spread across the corpus: the triage is efficient, but its yield cannot be read as a corpus-wide \maru{} error rate, and we never report one (\limref). Second, any analysis comparing models on the reviewed subset must account for those items being chosen partly because some models found them hard (Section~\ref{sec:results-ceiling}). Third, the direction of any selection-induced bias is knowable: review could move a label only where the panel disputed it, so every correction moves a label \emph{toward} a reading at least one frontier model preferred --- a structural tailwind for LLM-class systems that we bound rather than ignore, by checking whether families outside the panel replicate the gains (Section~\ref{sec:rn-triple}). The triage is a search heuristic for likely errors, audited downstream by humans --- not itself a source of ground truth.}{The selection has a property worth stating exactly. Because the panel is built from frontier models, the reviewed subset is \emph{enriched} for items where strong models disagree with \maru{}, and is therefore \emph{not} a random sample of the benchmark. Three consequences follow, and we hold to all three for the rest of the paper. First, the corrections are concentrated in this suspicious tail, not spread across the corpus: the triage is efficient --- of the 363 flagged items, 211 received a corrected label --- but this cannot be read as a corpus-wide \maru{} error rate, and we never report one (\limref). Second, any analysis that compares models on the reviewed subset must account for the fact that those items were chosen partly because some models found them hard; we return to this when we use the reviewed items to study human agreement (Section~\ref{sec:results-ceiling}). Third, the direction of any selection-induced bias is knowable and worth stating: review could move a label only where the panel disputed it, so every correction moves a label \emph{toward} a reading at least one frontier model preferred, and the corrected tail therefore carries a structural tailwind for LLM-class systems that a random-audit design would not. We do not net this out; we bound it, by checking whether families outside the panel --- including one released only after the triage ran --- replicate the gains (Section~\ref{sec:rn-triple}) and whether the panel family leads on the contested items (it does not; Section~\ref{sec:results-ceiling}). The triage is a search heuristic for likely errors, audited downstream by humans --- not itself a source of ground truth.}

\subsection{Lexicographer Review}\label{sec:lexen:review}

\shortlong{The 363 triaged items were reviewed by three professional lexicographers (RF, PW, PH) on a dedicated platform that showed each target token in its full sentence context alongside the complete set of WordNet candidates for that lemma and part of speech, with glosses and examples. For each item a reviewer chose the single best-fitting sense, selected an explicit \emph{cannot-answer} response when no candidate fit, and could leave a free-text rationale.}{The 363 triaged items were reviewed by three professional lexicographers, whom we identify by their initials RF, PW, and PH. Review took place on a dedicated annotation platform (\texttt{marureview.com}) that presented each item as the target token highlighted in its full sentence context alongside the complete set of WordNet sense candidates for that lemma and part of speech, each with its gloss and examples. For every item a reviewer chose the single best-fitting sense, selected an explicit \emph{cannot-answer} response when no candidate fit, and could leave a free-text rationale. \journalonly{The cannot-answer response was further typed as \emph{no sense applies}, \emph{inventory inadequate}, or \emph{input defective}, a distinction we exploit in the error typology of Section~\ref{sec:results-ceiling}.}}

\shortlong{The review was independent and provenance-hidden, though not cue-free. Reviewers worked independently and could not see one another's choices, so agreement reflects convergent judgment rather than anchoring on a colleague. They were not shown the provenance of any item --- not the \maru{} label, not which panel member flagged it, not which sense any source preferred. The interface highlighted candidate senses an earlier annotation and the automated triage systems had judged plausible to make long sense lists navigable; hiding provenance reduces explicit source bias but does not remove possible anchoring from the highlights themselves. The brief stated that a highlight is not a hint: the correct sense may be unhighlighted, or the verdict a cannot-answer flag. Reviewers chose from the full candidate list, and we discuss the highlighting and its possible anchoring effect in \limref. The full reviewer brief and interface are in Appendix~\ref{sec:app-brief}.}{The review was independent and provenance-hidden, though not cue-free. Reviewers worked independently and could not see one another's choices, so agreement reflects convergent judgment rather than discussion or anchoring on a colleague. They were not shown the provenance of any item: not the \maru{} label, not which panel member had flagged it, not which sense the source label or any model preferred. The interface did, however, highlight the candidate senses an earlier annotation and the automated triage systems had judged plausible --- two or more per item when those sources disagreed --- to make long sense lists navigable. Hiding the source of each highlight reduces explicit source bias, but does not remove possible anchoring from the highlights themselves. The brief therefore stated that a highlight is not a hint: the correct sense may be unhighlighted, the right verdict may be a cannot-answer flag, and a lone highlight means only that the sources agreed. Reviewers were asked to do their own analysis first and then note whether it landed on a highlighted sense, either outcome being equally valid. The triage thus decided which items a human saw and which senses were surfaced for attention, but never revealed which sense was the source label or a model's pick, so the recorded verdict is the reviewer's own judgment. We do not claim the candidate set was presented free of cues; the highlighting and its possible anchoring effect are discussed in \limref, and the full reviewer brief and interface are reproduced in Appendix~D.}

\subsection{Adjudication and the Gold Rule}\label{sec:lexen:adjudication}

\shortlong{Gold is decided by a single rule, frozen before adjudication and applied mechanically. A reviewed item is \emph{retained} when at least two of three reviewers agree on a sense, which becomes the \lexen{}-v1 label whether or not it matches \maru{}; it is \emph{removed} when at least two return cannot-answer or when all three disagree (no two-of-three majority). The rule never invents a label and only ratifies or discards. Applied to the 363 items it yields the composition in Table~\ref{tab:t1}: \textbf{124} retained unanimously, \textbf{183} by two-of-three (\textbf{307} retained), and \textbf{56} removed (\textbf{27} for cannot-answer, \textbf{29} for no consensus). Released \lexen{}-v1 thus contains \textbf{4{,}861} items: the 307 retained plus \textbf{4{,}554} unreviewed items carried over with their \maru{} label. Of the 307 retained, \textbf{211} received a corrected label.}{The reviewed items are decided by a single rule, frozen before adjudication and applied mechanically to the three reviewers' choices. An item is \emph{retained} when at least two of the three reviewers agree on a sense; that agreed sense becomes the \lexen{}-v1 label, whether or not it matches \maru{}. An item is \emph{removed} when at least two reviewers return cannot-answer, or when all three reviewers disagree on the sense (no two-of-three majority). The rule never invents a label and never lets a single reviewer override the others; it only ratifies or discards. Applied to the 363 items it yields the composition in Table~\ref{tab:t1}: \textbf{124} retained by unanimous three-way agreement, \textbf{183} retained by two-of-three agreement (\textbf{307} retained in all), and \textbf{56} removed, of which \textbf{27} were removed because at least two reviewers could not answer and \textbf{29} because the three reviewers reached no sense consensus. The released \lexen{}-v1 therefore contains \textbf{4{,}861} items: the 307 reviewed-and-retained, plus the \textbf{4{,}554} unreviewed items carried over with their \maru{} label. Of the 307 retained items, \textbf{211} received a corrected label; 56 items were dropped from the source.}

\shortlong{The design point is conservatism: because retention requires two independent professionals to converge, a corrected label is never one annotator's opinion but tracks expert \emph{meaning} judgments, not annotation history. The two boxes below (with further examples in Appendix~\ref{sec:app-examples}) illustrate the error types the rule catches, one per correction type; all are unanimous three-way corrections carrying the reviewers' own rationale. Sense keys follow \citet{maru2022}, written \sk{lemma\%pos:\allowbreak lexfile:\allowbreak lexid::}.}{The design point is conservatism. Because retention requires two independent professionals to converge, a corrected label is never one annotator's opinion: where the reviewers replace \maru{} (on 211 of the 307 retained items, roughly two-thirds), the new sense is backed by at least two lexicographers who chose it without seeing each other's answer or the original label. The corrections track expert \emph{meaning} judgments, not annotation history. The five boxes below illustrate the kinds of error the rule catches, one per correction type; all are unanimous three-way corrections, and each carries the reviewers' own recorded rationale. We use \citet{maru2022} sense keys throughout, written as \sk{lemma\%pos:\allowbreak lexfile:\allowbreak lexid::}.}

\begin{examplebox}[E1a: \emph{week} --- over-specific source label (NOUN)]
\textit{``\ldots{}the same sort of verbal fireworks that have dominated the talks for the past \tw{week}.''} \;(\texttt{semeval2013.d000.s011.t003})\\[3pt]
\maru{}: \sk{week\%1:28:02::} ``a period of seven consecutive days starting on Sunday'' $\rightarrow$
\lexen{}: \sk{week\%1:28:00::} ``any period of seven consecutive days''.\\[3pt]
RF, PW, and PH independently chose the general sense; the three frontier models agreed.
\emph{Verdict:} the source picked a more specific sense than the text supports --- the cleanest failure mode, with no genuine ambiguity.
\end{examplebox}

\begin{examplebox}[E1b: \emph{study} --- process vs.\ product (NOUN)]
\textit{``The \tw{study} focuses on a microbe found on Earth.''} \;(\texttt{semeval2013.d007.s010.t000})\\[3pt]
\emph{Context (next sentence):} \textit{``\ldots{} speculation \ldots{} that \tw{the report} would disclose the discovery of extraterrestrial life.''}\\[3pt]
\maru{}: \sk{study\%1:04:00::} ``a detailed critical inspection'' (the activity) $\rightarrow$
\lexen{}: \sk{study\%1:10:00::} ``a written document describing the findings'' (the report).\\[3pt]
PH: \textit{``\ldots{}this is more about the paper itself and less about the process of investigating.''} All three reviewers and all three models chose the report sense.
\emph{Verdict:} regular polysemy --- the activity confused with the document it produces. The target sentence alone is ambiguous; the surrounding text (``the report'') fixes the document reading, which is why reviewers worked from the full context. A shape shared by \emph{discovery}, \emph{creation}, \emph{negotiation}, and \emph{sale}.
\end{examplebox}

\journalonly{\begin{examplebox}[E1c: \emph{inherent} --- fine adjective shade (ADJ)]
\textit{``\ldots{}the latent talents and the wonderful\ldots{} potentialities \tw{inherent} in the souls of all our children.''} \;(\texttt{senseval2.d002.s011.t019})\\[3pt]
\maru{}: \sk{inherent\%5:00:00:implicit:00} ``in the nature of something though not readily apparent'' $\rightarrow$
\lexen{}: \sk{inherent\%5:00:00:intrinsic:00} ``existing as an essential constituent or characteristic''.\\[3pt]
RF, PW, and PH were unanimous on the \emph{intrinsic} reading. \emph{The models split: GPT-5.5 was correct; Gemini-3.1-Pro and Claude-Opus-4.8 kept the source sense.}
\emph{Verdict:} a genuine satellite-adjective distinction where experts agree but models do not --- evidence that fine granularity is hard for models too.
\end{examplebox}

\begin{examplebox}[E1d: \emph{however} --- ultra-fine function word (ADV)]
\textit{``\tw{However}, in both studies, patients whose cancer was not affecting squamous cells had longer survival times\ldots{}''} \;(\texttt{semeval2015.d000.s036.t000})\\[3pt]
\maru{}: \sk{however\%4:02:00::} ``despite anything to the contrary (usually following a concession)'' $\rightarrow$
\lexen{}: \sk{however\%4:02:04::} ``by contrast; on the other hand''.\\[3pt]
PH: \textit{``Sense 1's `(usually following a concession)'\ldots{} led me to feel that sense 2 was more suited.''} RF, PW, and PH unanimous; GPT-5.5 correct, the other two models kept the source label.
\emph{Verdict:} WordNet splits even \emph{however} into concession and contrast; the inventory's granularity, not the word's difficulty, drives the error.
\end{examplebox}

\begin{examplebox}[E1e: \emph{call} --- verb sense via argument structure (VERB)]
\textit{``\ldots{}the sound of bells cascading from its tower, \tw{calling} the faithful to evensong.''} \;(\texttt{senseval2.d000.s003.t014})\\[3pt]
\maru{}: \sk{call\%2:41:04::} ``call a meeting; invite or command to meet'' $\rightarrow$
\lexen{}: \sk{call\%2:32:05::} ``order, request, or command to come''.\\[3pt]
PH: \textit{``I selected sense 5 because it takes a human object, whereas in sense 7 it's the meeting that is called.''} RF, PW, and PH unanimous; all three models correct.
\emph{Verdict:} the correction turns on argument structure --- a human object versus an event object --- and exposes the lexicographer's reasoning directly.
\end{examplebox}
}

\journalonly{These five span the part-of-speech range and the major correction mechanisms; a further sixteen worked examples, organized by mechanism, appear in Appendix~B.}

\subsection{The \glite{} Coarsening Layer}\label{sec:lexen:glite}

\shortlong{Several corrections above turn on distinctions even three professionals will not always draw the same way, and the agreement study in Section~\ref{sec:results-ceiling} shows the effect is systematic: fine WordNet senses over-specify. \lexen{} therefore ships a second layer, the \emph{\glite{} coarse} layer, a many-to-one map from sense keys to \glite{} concepts. The map is a deterministic lookup applied identically to gold and predictions --- a coarse score is the fine evaluation re-graded, never re-annotated. We release this map in full over the evaluation set --- \textbf{10{,}412} WordNet sense keys over 6{,}505 concepts, enough to coarse-grade every prediction on all 4{,}861 items --- so every coarse number in this paper is reproducible from the released artifacts. \lexen{} in fact ships \emph{two} coarsening layers --- the authored \glite{} map and a bundled public CSI inventory \citep{lacerra2020} --- and \sensebench{} re-scores every run under WordNet-fine, \glite{}, or CSI labels, each selectable on the leaderboard. We state the authored layer's status directly: the grouping is ours rather than a third-party standard, though it was developed independently of, and prior to, this study and not constructed for this evaluation, so the coarse results are reported under that authored inventory; to show they do not rest on our grouping alone, \Cref{sec:results-noise} re-grades the frontier models under three public coarsenings no author controls (CSI \citep{lacerra2020}, WordNet supersenses \citep{ciaramita2003}, and WordNet Domains \citep{magnini2000}), where the qualitative result holds, and the granularity itself is externally validated: of six candidate coarsenings, \glite{} most closely reproduces the sense divisions of five professional dictionaries (\Cref{tab:tdict}). This lets Section~\ref{sec:results-ceiling} report both granularities and show that much fine disagreement dissolves when over-specified distinctions are collapsed.}{Several of the corrections above turn on distinctions that even three professionals will not always draw the same way, and the agreement study in Section~\ref{sec:results-ceiling} shows the effect is systematic: fine WordNet senses over-specify. To support measurement at a granularity where the distinctions are practically meaningful, \lexen{} ships a second labeling layer: the \emph{\glite{} coarse} layer, a many-to-one map from WordNet sense keys to \glite{} concepts. The coarsening is a deterministic lookup applied identically to gold labels and to model predictions, so a coarse score is exactly the fine evaluation re-graded under the \glite{} coarse map; we never re-annotate at the coarse level. We release this map in full over the candidate inventory of the evaluation set --- \textbf{10{,}412} distinct WordNet sense keys mapped to 6{,}505 \glite{} concepts, enough to coarse-grade every prediction on all 4{,}861 items, with the few uncovered gold keys carrying an explicit \texttt{unmapped} marker rather than a guess --- so every coarse-granularity number in this paper is reproducible from the released artifacts. In fact \lexen{} ships \emph{two} coarsening layers: the authored \glite{} map described here and a bundled public CSI inventory \citep{lacerra2020}, and \sensebench{} re-scores every run under WordNet-fine, \glite{}, or CSI labels, each selectable on the public leaderboard, so the coarse view is never tied to a single authored grouping. We state the authored layer's status directly: the grouping is ours rather than a third-party standard --- though it was developed independently of, and prior to, this study, and not constructed for this evaluation --- so coarse-granularity results are reported under that authored inventory. So that the coarse-level finding does not rest on our grouping alone, \Cref{sec:results-noise} re-grades the same frontier predictions under three public coarse inventories no author of this paper controls --- CSI \citep{lacerra2020}, the WordNet supersenses \citep{ciaramita2003}, and WordNet Domains \citep{magnini2000} --- and finds the qualitative result --- coarse accuracy far above fine, hard-item coarse accuracy in the high-70s to low-80s --- unchanged. The granularity of the layer is itself externally validated: of six candidate coarsenings, \glite{} reproduces the sense divisions of five professional dictionaries more closely than any public inventory (\Cref{tab:tdict}). This layer is what lets Section~\ref{sec:results-ceiling} report agreement and accuracy at both granularities, and show that a large share of fine disagreement dissolves when the over-specified distinctions are collapsed.}

\begin{table}[H]
\centering
\scriptsize
\setlength{\tabcolsep}{4pt}
\caption{\textbf{Professional-dictionary validation of coarse inventories.} Each mapping induces a partition of fine \wn{} senses; the score is agreement with the consensus partition induced by five professional dictionaries on 100 polysemous words.}
\label{tab:tdict}
\begin{tabular}{@{}lrr@{}}
\toprule
\textbf{Mapping} & \textbf{Consensus (\%)} & \textbf{Groups/word} \\
\midrule
\glite{} & 76.5 & 2.86 \\
CSI & 72.9 & 3.54 \\
WordNet Domains & 58.5 & 2.47 \\
WordNet supersenses & 58.1 & 2.63 \\
Homonymy proxy & 52.1 & 2.34 \\
Hypernym root & 40.6 & 2.01 \\
\bottomrule
\end{tabular}
\\[2pt]\footnotesize\textit{The dictionaries are Merriam-Webster, Collins, Cambridge, Oxford Learner's, and Longman. \glite{} is author-developed, so this table is an external convergent-validity check rather than a scoring result. \emph{Homonymy proxy} groups senses by coarse homonymy (related senses merged, unrelated homonyms split); \emph{Hypernym root} groups them by their top-level WordNet hypernym. Professional dictionaries average 3.24 groups per word; fine WordNet averages 4.44.}
\end{table}

\subsection{Release and Governance}\label{sec:lexen:release}

\shortlong{\lexen{}-v1 is released as an immutable, content-addressed artifact: a cryptographic hash on the dataset file, and every label change recording the three reviewer choices, the frozen rule, the triage evidence, and the cannot-answer types, so any \lexen{}-v1 label traces to its evidence. The release is frozen, never edited in place --- a revision ships as \lexen{}-v2 with its own hash, leaving v1 citable. To detect leakage into training data, it embeds a contamination canary, a unique improbable string, so a model echoing it can be flagged. The data layer is CC\,BY-NC; the code is Apache\,2.0. With the harness of Section~\ref{sec:sensebench}, these terms make every \lexen{} claim third-party re-derivable.}{\lexen{}-v1 is released as an immutable, content-addressed artifact. The dataset file carries a cryptographic hash; every label change records the three reviewer choices and the frozen rule that produced it; and the triage evidence, the per-reviewer selections, and the cannot-answer types are all retained, so any single \lexen{}-v1 label can be traced back to its evidence. The release is versioned and frozen, never edited in place: a future revision would be published as \lexen{}-v2 with its own hash, leaving v1 citable and reproducible. To guard against the benchmark leaking into model training data, the release embeds a contamination canary --- a unique, improbable string distributed with the data --- so that a model echoing it can be flagged. Licensing follows the standard split for this kind of resource: the data layer is released under CC\,BY-NC, and the accompanying construction and evaluation code under Apache\,2.0. Together with the auditable evaluation harness of Section~\ref{sec:sensebench}, these terms make \lexen{} a resource whose every claim a third party can re-derive from the released artifacts.}

\journalonly{\paragraph{A reusable recipe.} The construction we have described --- rank with a model panel, review the suspicious tail with independent professionals, adjudicate with a frozen majority rule, and release with full lineage --- is not specific to this benchmark. Model-assisted triage concentrates expensive human attention where it pays off --- 211 of the 363 flagged items were corrected --- while independent adjudication and provenance-hidden review ensure that models do not set retained labels. The models still shape which items are reviewed and which candidates are highlighted, so the recipe is efficient and auditable rather than fully independent. We offer it as a general procedure for correcting near-saturated benchmarks, and return to its governance implications in Section~\ref{sec:discussion}.}

\journalonly{\begin{table}[H]
\centering
\small
\caption{Dataset composition of lexEN-v1.}
\label{tab:t1}
\begin{tabular}{lr}
\toprule
\textbf{Quantity} & \textbf{Count} \\
\midrule
Maru2022 source instances & 4917 \\
lexEN-v1 retained & 4861 \\
Unreviewed, kept with Maru2022 label & 4554 \\
Model-triaged (reviewed) & 363 \\
— retained, three-way exact agreement & 124 \\
— retained, two-of-three agreement & 183 \\
— removed, $\geq$2 reviewers cannot-answer & 27 \\
— removed, three-way no-consensus & 29 \\
Gold labels changed (of 307 retained) & 211 \\
\bottomrule
\end{tabular}
\\[2pt]\footnotesize\textit{The 363 reviewed items split into 307 retained (124 three-way + 183 two-of-three agreement) and 56 removed (27 with $\geq$2 reviewers cannot-answer + 29 three-way disagreement); 211 of the 307 retained had their gold corrected. The other 4{,}554 source items were kept with the Maru2022 label, unreviewed.}
\end{table}
}
 \section{\sensebench{}: An Auditable Evaluation Protocol for LLM WSD}\label{sec:sensebench}

\shortlong{With leading systems separated by less than a point, an unaudited harness becomes the dominant source of irreproducibility: a silent change to the prompt, candidate ordering, or reply parsing can move a system by more than the gap to its neighbour. \sensebench{} is the evaluation counterpart to \lexen{}: a public harness for LLM WSD that fixes the task, registers the prompts, retains the raw model responses, and re-derives every published accuracy from them in continuous integration, scoring each run at multiple sense granularities and on the hard reviewed subset as well as the full set. We specify the task (\Cref{sec:sb-task}), the prompt registry (\Cref{sec:sb-prompts}), run artifacts and verification (\Cref{sec:sb-verify}), and statistical methodology (\Cref{sec:sb-stats}).}{A corrected layer is only useful if the numbers measured on it can be reproduced and contested. The
regime that motivates \lexen{} --- leading systems separated by less than a point --- is exactly the
regime in which an unaudited evaluation harness becomes the dominant source of irreproducibility: a
silent change to the prompt, the candidate ordering, or the parsing of a model's reply can move a
system by more than the gap that separates it from its neighbour. \sensebench{} is the evaluation
counterpart to \lexen{}: a public harness for LLM WSD that fixes the task, registers the prompts, retains the raw
model responses, and re-derives every published accuracy from those responses in continuous
integration, scoring each run at multiple sense granularities and on the hard reviewed subset as well
as the full set. This section specifies the task formulation (\Cref{sec:sb-task}), the immutable prompt
registry (\Cref{sec:sb-prompts}), the run artifacts and their verification (\Cref{sec:sb-verify}),
and the statistical methodology behind the leaderboard's confidence intervals and rank ranges
(\Cref{sec:sb-stats}).}

\subsection{Task formulation: constrained-index multiple choice}\label{sec:sb-task}

\shortlong{\sensebench{} scores WSD as \emph{constrained-index multiple choice} --- the inventory-constrained disambiguation step, with the target span identified and the candidate senses supplied (we scope this against end-to-end lexical semantics in \limref) --- not open generation. For each instance the harness presents the marked target word with the full set of WordNet candidate senses for that lemma and part of speech, each rendered with its gloss (and, under the rich prompt, synonyms and examples) and an integer index. The model returns one \texttt{sense\_index}; the harness maps it to a sense key and scores against the \lexen{}-v1 label. The candidate set is exactly the inventory the lexicographers saw, so a model is never penalized for missing a sense not on offer, and every prediction is a well-formed inventory member.}{\sensebench{} scores WSD as \emph{constrained-index multiple choice} --- the inventory-constrained disambiguation step, with the target span identified and the candidate senses supplied (we scope this against end-to-end lexical semantics in \limref) --- not open generation. For each
test instance the harness presents the target word marked in its sentence context together with the
full set of WordNet candidate senses for that lemma and part of speech, each candidate rendered with
its gloss and (under the rich prompt) its synonyms and usage examples and assigned a small integer
index. The model returns a single \texttt{sense\_index}; the harness maps that index back to a
WordNet sense key and scores the key against the \lexen{}-v1 label. The candidate set is exactly
the sense inventory the lexicographers themselves saw, so a model is never penalized for failing to
guess a sense that was not on offer, and the prediction is always a well-formed member of the
inventory.}

\journalonly{This design is a deliberate methodological choice with a known cost. Constraining the output to an
index of a presented candidate removes two confounds that would otherwise contaminate a cross-family
comparison. It removes \emph{surface-form ambiguity}: an open-generation system that emits a gloss or
a synonym string must then be matched back to a sense key by a fuzzy procedure that is itself a
source of disagreement, and that matching can favour models whose phrasing happens to resemble
WordNet's. It also removes \emph{inventory recall} from the measurement: open generation entangles
the model's knowledge of which senses exist with its ability to choose among them, whereas the index
task isolates the choice. The result is a clean, mechanically verifiable target that every system in
the leaderboard answers under identical conditions.}

\shortlong{The cost is that constrained-index selection is \emph{easier} than open generation: the candidate list is handed over, and chance accuracy is bounded by the reciprocal of the candidate count. The accuracies here are thus upper bounds on what the same models would reach producing sense keys unaided, not directly comparable to generation- or retrieval-based figures --- a measurement limitation (\limref) bearing on how the headline accuracies read, not on the internal validity of the cross-system comparison the harness protects.}{The cost of that cleanliness is that constrained-index selection is \emph{easier} than open
generation: the model is handed the candidate list rather than having to recall or construct it, and
chance accuracy is bounded below by the reciprocal of the candidate count rather than by the size of
the whole inventory. The accuracy numbers in this paper are therefore upper bounds on what the same
models would achieve if required to produce sense keys unaided, and they are not directly comparable
to generation-based or retrieval-based WSD figures. We flag this explicitly as a limitation of the
measurement (\limref); it bears on how the headline accuracies should be read, not on
the internal validity of the cross-system comparison, which is what the harness is built to protect.}

Because every system answers every item from the supplied candidate list --- full coverage, no
abstention --- the accuracy we report coincides with the micro-averaged F1 conventionally reported
in the WSD literature; we use the shorter name throughout.

\subsection{Prompts: an immutable registry}\label{sec:sb-prompts}

\shortlong{Every \emph{prompt template} is a registered JSON object with a stable identifier, and is immutable: changing any rendering decision produces a new identifier, never a silent edit. A record fixes the message templates, surrounding context, candidate ordering (its accuracy effect is isolated in \Cref{sec:rn-ordering}), which fields accompany each candidate, and the output format, and is rendered per item into the prompt a model actually sees. Because the identifier travels with every run, two accuracies are comparable iff they share it, and any reader can reconstruct the exact text a model saw.}{Every \emph{prompt template} \sensebench{} can run is a registered JSON object with a stable identifier, and a
registered template is immutable: changing any rendering decision produces a new identifier rather than
a silent edit to an existing one. A prompt record fixes the system and user message templates, the
amount of surrounding context, the candidate ordering (whose accuracy effect we isolate with a randomized-order control in \Cref{sec:rn-ordering}), which fields accompany each candidate (gloss,
synonyms, usage examples, sense key), and the expected output format; the template is rendered per item, with the target word and its candidate senses substituted in, into the prompt a model actually sees. Because the identifier travels
with every run and every leaderboard row, two accuracies are comparable if and only if they share a
prompt identifier, and any reader can reconstruct, character for character, the text a model saw.}

\journalonly{The registry separates two questions that are easy to conflate: how good a model is at the task, and
how much the task framing helps it. To answer the second question without contaminating the first,
the registry pairs prompts that differ along controlled axes.}
\shortlong{The two prompts anchoring the leaderboard sit at opposite ends of the information spectrum. Prompt \texttt{p001} is rich: a ``5+1'' context (five preceding, one following sentence) over detokenized text, each candidate rendered with sense key, definition, synonyms, and examples, frequency-ordered, the answer a JSON object \texttt{\{"sense\_index": k\}}. Prompt \texttt{p002} is minimal: the target sentence alone, candidates as index plus definition and at most one example, no keys or synonyms, a bare-integer answer. Because \texttt{p001} and \texttt{p002} differ only in context and gloss enrichment, the gap between them on the leaderboard measures the value of both directly; intermediate prompts (\texttt{p003} onward) isolate individual factors for the ablation in \Cref{sec:results-noise}. Appendix~\ref{sec:app-prompt} and the registry give a fully rendered prompt.}{The two prompts that anchor the main leaderboard sit at opposite ends of the information spectrum.
Prompt \texttt{p001} is the rich variant: it places the target sentence inside a window of five
preceding and one following sentence (a ``5+1'' context), detokenizes the Penn-Treebank source into
natural English before marking the target, and renders each candidate sense with its WordNet sense
key, definition, up to six synonyms, and up to two usage examples, ordered by WordNet frequency, with
the answer returned as a JSON object \texttt{\{"sense\_index": k\}}. Prompt \texttt{p002} is the
minimal variant: the target sentence alone, no surrounding context, candidates given as index plus
definition and at most one example, no sense keys and no synonyms, and the answer returned as a bare
integer. Because \texttt{p001} and \texttt{p002} differ only in context and gloss enrichment, the gap
between them on the leaderboard measures the value of both directly, and a family of intermediate prompts (\texttt{p003} onward) isolates individual factors
for the controlled ablation reported in \Cref{sec:results-noise}. Box~E6 shows a fully rendered \texttt{p001} prompt.}

\journalonly{\begin{examplebox}[E6: a rendered \texttt{p001} prompt (item \texttt{senseval2.d000.s003.t009}, target \tw{fields})]
\ttfamily\footnotesize\raggedright
\textbf{[system]}\\
You are an English linguist performing fine-grained Word Sense Disambiguation. WordNet glosses are
schematic. Use the sentence context, the examples, and the synonyms to choose the best indexed sense.
Do not hedge. Choose the single best indexed sense.\\[4pt]
\textbf{[user]}\\
Target lemma: field\\
Target surface form: fields\\
Context:\\
\dots{} Dorothy L.\ Sayers, ``The Nine Tailors''. ASLACTON, England --- Of all scenes that evoke
rural England, this is one of the loveliest: An ancient stone church stands amid the
{<}t{>}fields{<}/t{>}, the sound of bells cascading from its tower, calling the faithful to
evensong. The parishioners of St.\ Michael and All Angels stop to chat at the church door
\dots{}\\[3pt]
Candidate senses:\\
1. sense\_key=field\%1:15:00:: | definition=a piece of land cleared of trees and usually
enclosed | examples=he planted a field of wheat\\
2. sense\_key=field\%1:15:04:: | definition=a region where a battle is being (or has been)
fought | synonyms=battlefield, battleground, field of battle, field of honor | examples=they made
a tour of Civil War battlefields\\
\dots{}\\
9. sense\_key=field\%1:17:00:: | definition=extensive tract of level open land | synonyms=plain,
champaign | examples=they emerged from the woods onto a vast open plain; he longed for the fields
of his youth\\
\dots{} \textrm{(17 candidates total, frequency-ordered; elisions ours)}\\[3pt]
Instructions:\\
1. Choose the single best indexed sense for the target word in context.\\
2. WordNet glosses are schematic; use the context, synonyms, and examples together.\\
3. Do not answer with a WordNet sense key.\\
4. Return only a JSON object exactly like \{"sense\_index": 3\}.\\[4pt]
\textbf{[assistant]} \{"sense\_index": 1\}\\[2pt]
\normalfont\footnotesize Index \texttt{1} resolves to \sk{field\%1:15:00::} (enclosed farmland), the
\lexen{}-v1 label; \maru{} had labeled this item \sk{field\%1:17:00::} (open plain), index
\texttt{9}.
\end{examplebox}
}

\subsection{Run artifacts and verification}\label{sec:sb-verify}

\shortlong{Each evaluation produces three artifacts that make an accuracy reconstructible from first principles. A \texttt{run.json} header records the resolved model and provider, prompt identifier, dataset content hash, decoding policy, token usage and cost, harness git commit, and date --- pinning a leaderboard row to a specific model version on a specific day. A \texttt{predictions.jsonl} records per item the parsed \texttt{sense\_index}, its sense key, and the verdict. A compressed \texttt{calls.jsonl.gz} retains the \emph{raw} request and response for every item, before any parsing.}{Each evaluation produces three artifacts that together make a published accuracy reconstructible from
first principles. A \texttt{run.json} header records the resolved model identifier and provider, the
prompt identifier, the dataset content hash, the sampling and decoding policy, token usage and cost,
the git commit of the harness, and the date; this is the provenance that pins a leaderboard row to a
specific model version on a specific day. A \texttt{predictions.jsonl} file records, per item, the
parsed \texttt{sense\_index}, the sense key it resolves to, and the correctness verdict. A
compressed \texttt{calls.jsonl.gz} file retains the \emph{raw} request and response for every item ---
the exact rendered prompt and the exact model reply, before any parsing.}

\shortlong{The raw call log makes the harness auditable, not merely documented. On every change, continuous integration re-parses \texttt{calls.\allowbreak jsonl.\allowbreak gz}, re-resolves indices to sense keys, re-scores against the frozen \lexen{} labels, and checks that the regenerated \texttt{predictions.\allowbreak jsonl} reproduces the published accuracy bit for bit. A scoring change that altered a number could not pass silently: the artifact verdict would no longer match the code, and the build would fail.}{The raw call log is what makes the harness auditable rather than merely documented. On every change,
continuous integration re-parses the responses in \texttt{calls.\allowbreak jsonl.\allowbreak gz},
re-resolves indices to sense keys, and re-scores against the frozen \lexen{} labels, then checks that
the regenerated \texttt{predictions.\allowbreak jsonl} reproduces the published accuracy bit for bit.
A scoring change that
altered a number could not pass review silently: the verdict in the artifact would no longer match
the verdict the code produces, and the build would fail. \journalonly{This inverts the usual trust
relationship for a leaderboard. Rather than asking readers to trust a reported accuracy, the harness
ships the evidence from which that accuracy is derived and a mechanical procedure that any third
party can rerun; the published number is a claim about the raw artifacts, and the artifacts are
public.}}

The canary introduced in \Cref{sec:lexen:release} supports the same contract as a matter of release
governance rather than automated enforcement: the sentinel ships embedded in the released data, so a
model that surfaces it in its output can be flagged and its score treated as contaminated rather than
valid; the harness does not itself scan responses for the sentinel.
\journalonly{The canary does not prevent contamination; it makes contamination \emph{detectable},
which is the property a living, public leaderboard needs to retain credibility as the underlying data
ages into model training corpora.}

\subsection{Statistical methodology}\label{sec:sb-stats}

\shortlong{Differences of a point or less carry this paper's claims --- whether top frontier families separate at all, and at which reasoning tier they are compared --- so the leaderboard reports uncertainty, not point estimates. Each accuracy carries a bootstrap 95\% confidence interval, resampling test items with replacement to propagate the finite-sample uncertainty of a 4{,}861-item evaluation. Read as a ranking, the leaderboard also reports \emph{rank ranges}: the span of positions a system could occupy across resamples, stating directly whether two adjacent systems separate at all. We give CIs and rank ranges for the headline systems in \Cref{sec:results-noise}, where the leader's interval clears the third system on the full set but the three interleave on the hard reviewed items.}{Differences of a point or less carry the central claims of this paper --- whether top frontier
families separate, and at which reasoning tier they are being compared --- so the leaderboard reports
uncertainty rather than point estimates alone. Each accuracy is accompanied by a bootstrap 95\%
confidence interval, resampling test items with replacement to propagate the finite-sample
uncertainty of a 4{,}861-item evaluation into the reported figure. Because the leaderboard is read as
a ranking, we also report \emph{rank ranges}: the span of leaderboard positions a system could occupy
across bootstrap resamples, which states directly whether two adjacent systems are separable at all
given the sample size. We report 95\% CIs and rank ranges for the headline systems in
\Cref{sec:results-noise}, where the leading system's interval clears the third on the full set, while
on the hard reviewed items the three intervals overlap and their rank ranges interleave.}

\shortlong{For close head-to-head comparisons we use a paired test rather than comparing marginal intervals, following the paired-bootstrap protocol of \citet{du2025}: both systems are scored on the same items, so the procedure resamples once and recomputes \emph{both} accuracies per resample, preserving the per-item correlation an unpaired comparison discards. This is the appropriate test for whether $A$ beats $B$ on this dataset, not whether each differs from a constant. A comparison is significant only when the paired difference excludes zero at the 95\% level.}{For close head-to-head comparisons we use a paired test rather than comparing marginal intervals,
following the paired-bootstrap protocol of \citet{du2025}. Two systems are scored on the same items,
so the paired procedure resamples items once and recomputes \emph{both} systems' accuracies on each
resample, preserving the per-item correlation that an unpaired comparison of two separate confidence
intervals discards; this is the statistically appropriate test when the question is whether system
$A$ beats system $B$ on this dataset rather than whether each differs from a fixed constant. We report
a comparison as significant only when the paired difference excludes zero at the 95\% level.}
\journalonly{The distinction matters precisely in the regime this paper studies: two systems whose
marginal confidence intervals overlap substantially can still be reliably ordered by a paired test,
and --- more often in our results --- two systems with non-overlapping marginal intervals can fail to
separate once shared per-item difficulty is accounted for. We therefore treat the top of the
leaderboard as a band of statistically indistinguishable systems rather than a strict order, and say
so wherever the ranking is reported.}

\shortlong{Finally, every comparison is frozen in time. Proprietary endpoints drift, so each row pins its resolved model version and evaluation date in \texttt{run.json}, and headline numbers cite specific dated runs. \Cref{tab:t12} gives the reproducibility map for four representative \texttt{p001} runs --- the top three families plus a Llama-3.1-8B cross-check validating the harness against published WSD figures --- with the run identifier, resolved version, reasoning setting, and date that let any result be re-verified.}{Finally, every comparison is frozen in time. Proprietary model endpoints drift, so each leaderboard
row pins the resolved model version and the evaluation date in its \texttt{run.json}, and the paper's
headline numbers cite specific dated runs. \Cref{tab:t12} gives the reproducibility map for four
representative \texttt{p001} runs --- the leaderboard's top three families plus a Llama-3.1-8B
cross-check used to validate the harness against published WSD figures --- showing the full run
identifier, the resolved model version, the reasoning setting, and the date that together let any
result be located and re-verified.}

\journalonly{\begin{table*}[t]
\centering
\small
\caption{Reproducibility map: a representative p001 run per top-three system (plus Llama-3.1-8B as a harness cross-check). These are the exact runs shipped in the supplementary package, so accuracy re-derives from released artifacts; the GPT-5.5 row is its medium-effort run (the xhigh run is the Table~\ref{tab:t4} champion). Run-ids are shown in full; all four use prompt p001 on the lexEN-v1 dataset (4{,}861 items, content-hash \texttt{sha256:5fd4382b}\ldots) and were executed on 2026-06-14. Each run.json additionally records the git commit, sampling and decoding policy, token usage, and cost.}
\label{tab:t12}
\resizebox{\linewidth}{!}{%
\begin{tabular}{llll}
\toprule
\textbf{Run ID} & \textbf{Resolved model} & \textbf{Reasoning} & \textbf{Date} \\
\midrule
{\scriptsize\ttfamily gpt-5.5-medium-reasoning-p001-lexen-v1-20260614} & {\scriptsize\ttfamily gpt-5.5-2026-04-23} & medium & 2026-06-14 \\
{\scriptsize\ttfamily gemini-3.1-pro-high-reasoning-p001-lexen-v1-20260614} & {\scriptsize\ttfamily gemini-3.1-pro-preview} & high & 2026-06-14 \\
{\scriptsize\ttfamily claude-fable-5-xhigh-reasoning-p001-lexen-v1-20260701-fallback-claude-opus-4-8} & {\scriptsize\ttfamily claude-fable-5} & xhigh & 2026-07-01 \\
{\scriptsize\ttfamily vllm-llama-3.1-8b-bf16-a100-p001-lexen-v1-20260614} & {\scriptsize\ttfamily meta-llama/Llama-3.1-8B-Instruct} & n/a & 2026-06-14 \\
\bottomrule
\end{tabular}
}
\end{table*}
}
 \section{Results I: Label Noise and the Current Frontier}\label{sec:results-noise}

\shortlong{This section tests \lexen{}'s promise that a human-adjudicated label layer changes which systems look good. Holding
predictions fixed and varying only the labels separates systems that improve as the gold improves
from those that do not (\Cref{sec:rn-triple}); we then read the \lexen{}-v1 leaderboard and its
cross-family checks (\Cref{sec:rn-board}), trace the cost/accuracy frontier
(\Cref{sec:rn-pareto}), treat reasoning effort as a first-class axis (\Cref{sec:rn-effort}),
isolate which parts of the prompt drive accuracy (\Cref{sec:rn-ablation}),
show that the frequency ordering of the candidates is a disclosed prior rather than the source of the
result (\Cref{sec:rn-ordering}), and close on what correction reveals about how each system
represents meaning (\Cref{sec:rn-sensitivity}).}{}\journalonly{The promise of \lexen{} is that a human-adjudicated label layer would change which systems look good and
by how much; this section tests that promise directly. We hold a system's predictions fixed and vary
only the labels they are scored against, separating systems that improve as the gold improves from
systems that do not (\Cref{sec:rn-triple}). We then read the \lexen{}-v1 leaderboard, where the
arrangement of the top scores supplies a cross-family check (\Cref{sec:rn-board}),
trace the cost/accuracy frontier across three orders of magnitude in price
(\Cref{sec:rn-pareto}), treat reasoning effort as a first-class axis of the comparison
(\Cref{sec:rn-effort}), isolate which parts of the prompt actually drive accuracy
(\Cref{sec:rn-ablation}), show that the frequency ordering of the candidate senses is a disclosed
prior rather than the source of the result (\Cref{sec:rn-ordering}), and close on what label
correction reveals about how each family of system represents word meaning
(\Cref{sec:rn-sensitivity}).}

\subsection{Label correction lifts the LLMs most}\label{sec:rn-triple}

\shortlong{The cleanest evidence that label quality, not model error, now governs WSD measurement comes from
re-scoring one set of predictions against successively better golds. \Cref{tab:t6}
scores each system's frozen output on the same 4{,}861 items three times: against the original
Raganato \textsc{ALL} labels \citep{raganato2017}, the \maru{} re-annotation \citep{maru2022}, and
\lexen{}. The predictions never change, so every movement is attributable to the labels, and its
shape is diagnostic.}{}\journalonly{The cleanest evidence that label quality, not model error, now governs WSD measurement comes from
re-scoring a single set of predictions against successively better gold standards.
\Cref{tab:t6} takes each system's frozen output on the same 4{,}861 items and scores
it three times: against the original Raganato \textsc{ALL} labels \citep{raganato2017}, against the
\maru{} re-annotation \citep{maru2022}, and against \lexen{}. Because the predictions never change,
every movement across the three columns is attributable to the labels alone, and the shape of that
movement is diagnostic.}

\begin{table*}[t]
\centering
\small
\caption{Label-noise triple-score: each system's fixed predictions re-scored under the original Raganato, Maru2022, and lexEN gold labels (same 4{,}861 fine-grained items). The three $\Delta$ columns are the accuracy change as the labels improve: R$\to$M (Maru$-$Raganato), M$\to$L (lexEN$-$Maru), and R$\to$L (lexEN$-$Raganato).}
\label{tab:t6}
\begin{tabular}{lrrrrrr}
\toprule
\textbf{System} & \textbf{Raganato} & \textbf{Maru2022} & \textbf{lexEN} & \textbf{$\Delta$\,R$\to$M} & \textbf{$\Delta$\,M$\to$L} & \textbf{$\Delta$\,R$\to$L} \\
\midrule
GPT-5.5 & 85.4 & 91.8 & 95.6 & +6.4 & +3.8 & +10.2 \\
Gemini-3.1-Pro & 85.2 & 92.1 & 94.9 & +6.9 & +2.8 & +9.7 \\
Claude-Fable-5 & 85.6 & 91.9 & 95.2 & +6.3 & +3.3 & +9.6 \\
GPT-5-mini & 82.2 & 88.1 & 90.7 & +5.9 & +2.6 & +8.5 \\
gemma-4-26B & 81.8 & 87.2 & 89.4 & +5.4 & +2.2 & +7.6 \\
GPT-4o-mini & 76.2 & 81.0 & 82.7 & +4.8 & +1.7 & +6.5 \\
\midrule
ConSeC & 81.8 & 84.2 & 84.9 & +2.4 & +0.7 & +3.1 \\
ESCHER & 77.7 & 80.8 & 81.4 & +3.1 & +0.6 & +3.7 \\
BEM & 76.6 & 79.3 & 79.7 & +2.7 & +0.4 & +3.1 \\
SANDWiCH & 86.6 & 86.1 & 85.2 & -0.5 & -0.9 & -1.4 \\
MFS & 58.9 & 61.3 & 61.6 & +2.4 & +0.3 & +2.7 \\
\bottomrule
\end{tabular}
\\[2pt]\footnotesize\textit{All columns score the \emph{same} predictions on the \emph{same} 4{,}861 lexEN items; only the gold key changes. As the labels improve almost every system's accuracy rises --- the corrections are improvements nearly all systems agree with, not an artifact favouring one family --- though the gains are far larger for the LLMs (top, $+6.5$ to $+10.2$) than for the supervised systems (bottom, $+2.7$ to $+3.7$, with one exception that declines slightly). The gap in slope, not the level, is the diagnostic signal. ConSeC here uses its paper-best training (SemCor plus WordNet's sense-tagged glosses and examples), whereas ESCHER is trained on SemCor alone; matched on SemCor the two differ by only $\sim$1 F1, so the larger ConSeC--ESCHER gap shown here is mostly a training-data effect, not architecture.}
\end{table*}

\shortlong{For the frontier language models the movement is large, monotonic, and consistent across families.
GPT-5.5 reads 85.4\% against Raganato, 91.8\% against \maru{}, and 95.6\% against \lexen{}, a
gain of 10.2 points with no reversal at either step; Gemini-3.1-Pro and Claude-Fable-5 trace the
same staircase ($+9.7$/$+9.6$), and smaller models climb 6.5--8.5 points along parallel paths. Even on
the existing public \maru{} labels, before any \lexen{} correction, the best model reads \textbf{92.1\%}
(Gemini-3.1-Pro) against \textbf{86.1\%} for the strongest current supervised system (SANDWiCH, trained on
today's corpora; its released predictions re-scored on the same labels): among such systems the frontier LLMs are the state of the art
on fine-grained English WSD on the standard benchmark, not only on our corrected layer ---
\Cref{sec:results-training} shows that repairing the training corpus roughly halves this gap. The
interpretation is direct: much of what looked like LLM error was the labels being wrong. These
systems were already answering the way the lexicographers eventually did; the benchmark was scoring
them against annotations the experts themselves later overturned.}{}\journalonly{For the frontier language models the movement is large, monotonic, and consistent across families.
GPT-5.5 reads 85.4\% against Raganato, 91.8\% against \maru{}, and 95.6\% against \lexen{}, a
gain of 10.2 points from the original labels to the corrected ones, with no reversal at either step.
Gemini-3.1-Pro and Claude-Fable-5 trace the same staircase, rising 9.7 and 9.6 points
respectively, and the smaller GPT-5-mini, Gemma 4 26B, and GPT-4o-mini climb by 8.5, 7.6, and 6.5
points along parallel paths. Even on the existing public \maru{} labels, before any \lexen{}
correction, the best model reads \textbf{92.1\%} (Gemini-3.1-Pro) against \textbf{86.1\%} for the
strongest current supervised system (SANDWiCH, trained on today's corpora; its released
predictions re-scored on the same labels): among such
systems, on the standard benchmark and not only on our corrected layer, the frontier LLMs are the
state of the art on fine-grained English WSD --- though \Cref{sec:results-training} shows that
repairing the training corpus roughly halves the gap. The interpretation is direct: a large fraction of what looked like LLM
error against the original labels was the labels being wrong, and as the gold standard is corrected
the apparent error dissolves. These systems were already answering the items the way the
lexicographers eventually did; the benchmark was scoring them against annotations that the experts
themselves later overturned.}

\shortlong{The supervised systems improve too, just much less. Over the full Raganato-to-\lexen{} span ConSeC
rises 3.1 points, ESCHER 3.7, BEM 3.1, and the most-frequent-sense baseline 2.7 --- real gains, but
most of them come from the larger Raganato-to-\maru{} re-annotation rather than from \lexen{}. On
\lexen{}'s own \maru{}-to-\lexen{} correction the supervised systems add only $+0.7$, $+0.6$, $+0.4$,
and $+0.3$ points, against $+1.7$ to $+3.8$ for the language models. The two
classes move by very different amounts, and they are furthest apart on this final step ---
the LLMs keep tracking the corrections the lexicographers make, the supervised systems much less so.
\emph{On \lexen{}'s own correction the LLMs keep gaining strongly while the supervised systems add
only a fraction of a point.} Because corrections could arise only on items the GPT-5.5 triage panel
contested, the gap's size is identified only over the triaged region; its direction, however,
survives the obvious control: frontier families with no member in the panel gain $+2.8$ to $+3.3$
on the same step --- including Claude-Fable-5, released after the triage ran ($+3.3$ overall;
$+76$ points on the 211 corrected items, $p<10^{-34}$) --- and that panel-independent gap, not the
full-span slope, is the diagnostic signal.}{}\journalonly{The supervised systems improve too, just much less. Re-scoring the same way moves ConSeC by 3.1
points across the full Raganato-to-\lexen{} span, ESCHER by 3.7, BEM by 3.1, and the
most-frequent-sense baseline by 2.7 --- real gains, but most of them come from the larger
Raganato-to-\maru{} re-annotation rather than from \lexen{}. On the step that isolates \lexen{}'s own
contribution, the final \maru{}-to-\lexen{} correction, the supervised systems add only $+0.7$,
$+0.6$, $+0.4$, and $+0.3$ points, against $+1.7$ to $+3.8$ for the language models. The two classes
move by very different amounts, and they are furthest apart on this
last step. The supervised systems' residual errors fall largely on a different set of items from the
ones \lexen{} corrects, so \lexen{}'s correction adds little to their scores; the language models, by
contrast, keep tracking the corrections the lexicographers make. \emph{On \lexen{}'s own correction
the LLMs keep gaining strongly while the supervised systems add only a fraction of a point.} One
caution and one check apply before reading this gap causally. The caution: items entered
lexicographer review only where the GPT-5.5 triage panel disputed \maru{} --- all 211 corrections
sit on items the triage recorded as contested by at least six of the eight panel variants --- so the \emph{size} of the class
asymmetry is identified only over the region the triage examined; a differently anchored triage
could surface different corrections. The check: the direction does not depend on the panel.
Stratifying the 211 corrections by flagging family, the frontier families with no member in the
panel gain $+2.8$ to $+3.3$ points on the same step ($+65$ to $+76$ points on the corrected items
themselves; exact McNemar $p<10^{-24}$), and Claude-Fable-5 --- released after the triage ran, so
structurally incapable of having flagged any of these items --- gains $+3.3$. Even on the 56
corrections that no supervised panel member had flagged, the non-panel models move
with the lexicographers while the supervised systems lose ground. That surviving gap on the final
step --- direction unconditionally, magnitude over the triaged region --- is the quantitative
signature of a benchmark whose remaining errors one class of system still shares with the gold and
the other largely does not.} \journalonly{This is also why the regime is new. While supervised systems sat in the
79--83\% band, their errors swamped the few-percent label-error rate, and a re-annotation pass like
\maru{}'s was a refinement rather than a verdict. Once a system clears 90\%, the residual label
error is the same order of magnitude as the residual model error, and which of the two a benchmark is
actually measuring stops being a rhetorical question.}

\shortlong{The same exercise at coarse granularity sharpens the picture. Re-scored against the \glite{} coarse
\lexen{} labels (\Cref{tab:t6b}), the frontier families reach 98.6--98.7\%, under two points of
headroom at the practically meaningful granularity. The fine-grained 94--95\% headline and the
coarse 98.7\% score describe the same predictions at two granularities: most of the fine-level
shortfall falls on distinctions the inventory draws more finely than its own annotators reproduce.}{}\journalonly{The same exercise at coarse granularity sharpens the picture. \journalonly{The \glite{} coarse
layer collapses the WordNet distinctions that competent readers do not reliably reproduce (the
inter-annotator evidence is in \Cref{sec:results-ceiling}), and scoring against the coarsening answers
a different question, whether a system picks the right \emph{concept} rather than the right
WordNet sense key.} Re-scored against the \glite{} coarse \lexen{} labels (\Cref{tab:t6b}), the frontier
families reach 98.6--98.7\%, leaving well under two points of headroom at the practically meaningful
level of granularity. The fine-grained 94--95\% headline and the coarse 98.7\% score describe the same predictions at two granularities: most of the fine-level
shortfall these systems show falls on distinctions the inventory draws more finely than its own
annotators reproduce.}
\journalonly{\begin{table*}[t]
\centering
\small
\caption{Label-noise triple-score at the \glite{} \emph{coarse}-concept level (companion to Table~\ref{tab:t6}): the same fixed predictions on the same 4{,}861 items, now comparing coarse concepts instead of fine sense keys. The three $\Delta$ columns are as in Table~\ref{tab:t6}.}
\label{tab:t6b}
\begin{tabular}{lrrrrrr}
\toprule
\textbf{System} & \textbf{Raganato} & \textbf{Maru2022} & \textbf{lexEN} & \textbf{$\Delta$\,R$\to$M} & \textbf{$\Delta$\,M$\to$L} & \textbf{$\Delta$\,R$\to$L} \\
\midrule
GPT-5.5 & 94.7 & 97.4 & 98.7 & +2.7 & +1.3 & +4.0 \\
Gemini-3.1-Pro & 94.8 & 97.6 & 98.6 & +2.8 & +1.0 & +3.8 \\
Claude-Fable-5 & 94.8 & 97.6 & 98.7 & +2.8 & +1.1 & +3.9 \\
GPT-5-mini & 93.5 & 96.1 & 97.0 & +2.6 & +0.9 & +3.5 \\
gemma-4-26B & 92.5 & 95.0 & 95.6 & +2.5 & +0.6 & +3.1 \\
GPT-4o-mini & 89.4 & 91.7 & 92.3 & +2.3 & +0.6 & +2.9 \\
\midrule
ConSeC & 93.3 & 93.9 & 93.9 & +0.6 & +0.0 & +0.6 \\
ESCHER & 90.6 & 91.2 & 91.1 & +0.6 & -0.1 & +0.5 \\
BEM & 90.5 & 91.4 & 91.0 & +0.9 & -0.4 & +0.5 \\
SANDWiCH & 94.9 & 93.9 & 93.4 & -1.0 & -0.5 & -1.5 \\
MFS & 75.2 & 77.0 & 76.6 & +1.8 & -0.4 & +1.4 \\
\bottomrule
\end{tabular}
\\[2pt]\footnotesize\textit{Coarsening collapses exactly the fine WordNet senses lexEN corrects, so accuracies rise to 92--98\% and the label-noise gaps shrink: the lexEN$-$Raganato gain falls from up to $+10$ points (fine) to about $+4$ (coarse).}
\end{table*}
}

\shortlong{Because the \glite{} coarsening is ours, we re-grade the same finding under three public coarse
inventories no author controls: CSI \citep{lacerra2020}, the 45 WordNet supersenses, and WordNet
Domains. The direction is inventory-independent (\Cref{tab:tcoarse}): every coarsening lifts hard-item
accuracy far above the 66\% fine floor --- to \textbf{78--82\%} across the public inventories, against
\textbf{87.5\%} under \glite{} --- raises three-way inter-annotator agreement, and preserves the model
ranking (Spearman $\rho$ 0.92--1.00). The public inventories land a few points \emph{below} \glite{}
rather than above, a conservative corroboration of the qualitative result rather than the exact
numbers; whether the reviewer$\leftrightarrow$model agreement envelope also survives the change of
inventory is the sharper test, taken up in \Cref{sec:results-ceiling}.}{}\journalonly{Because the \glite{} coarsening is our own, we re-grade the same finding under three public coarse
inventories that no author of this paper controls: CSI \citep{lacerra2020}, the 45 WordNet supersenses
(lexicographer files, derivable from the sense key), and WordNet Domains. The qualitative finding is
inventory-independent (\Cref{tab:tcoarse}). Every coarse inventory lifts hard-item accuracy far above
the 66\% fine floor --- to roughly \textbf{78\%} under CSI (the conservative uniform composite-partition
grading; ${\sim}82\%$ under CSI's native set-overlap), \textbf{81\%} under supersenses, and
\textbf{81\%} under WordNet Domains, against \textbf{87.5\%} under \glite{} --- raises three-way
inter-annotator agreement, and leaves the model ranking essentially unchanged (Spearman $\rho$
0.92--1.00 against fine). The public inventories land a few points \emph{below} \glite{} rather than
above, a conservative corroboration rather than an inflation: they partition senses differently, and
CSI maps fewer adjective and adverb keys, so they confirm the \emph{direction} of the finding ---
coarse accuracy far exceeds fine, and hard-item coarse accuracy reaches the high-70s to low-80s under
independent public coarsenings --- without reproducing the exact \glite{} numbers. The sharper test,
whether the reviewer$\leftrightarrow$model agreement envelope also survives the change of inventory, is
taken up in \Cref{sec:results-ceiling}.}

\subsection{The \lexen{}-v1 leaderboard and cross-family agreement}\label{sec:rn-board}

\shortlong{\Cref{tab:t4} reports the current \lexen{} leaderboard (one row per family at its best \texttt{p001}
run) with accuracy and per-item cost. The arrangement of its top gives a useful cross-family check on
the corrected tail.}{}\journalonly{\Cref{tab:t4} reports the current \lexen{} leaderboard (one row per model family, each at its best
\texttt{p001} run) with accuracy and per-item cost, and \Cref{tab:t5} places the classic
supervised systems on the same \lexen{}-v1 labels. The arrangement of the top of \Cref{tab:t4}, not any
single number in it, gives a useful cross-family check on the corrected tail.}

\begin{table*}[t]
\centering
\small
\caption{SenseBench lexEN-v1 leaderboard: one row per model family (its best-scoring p001 run). Accuracy includes a marginal 95\% item-bootstrap confidence interval; cost (\$ per million items) is from the same run. \emph{Hosting} is a metered cloud API (with reasoning effort) or a self-hosted GPU.}
\label{tab:t4}
\begin{tabular}{llllr}
\toprule
\textbf{Model} & \textbf{Type} & \textbf{Hosting} & \textbf{lexEN acc [95\% CI]} & \textbf{\$/1M} \\
\midrule
GPT-5.5 & proprietary & cloud (xhigh) & 95.6 [95.0, 96.2] & 10{,}700 \\
Claude Fable 5 (+ Opus 4.8 fallback) & proprietary & cloud (xhigh) & 95.2 [94.6, 95.8] & 14{,}555 \\
Gemini 3.1 Pro & proprietary & cloud (high) & 94.9 [94.3, 95.5] & 7{,}227 \\
Gemma 4 31B & open & self-host, H100 & 93.4 [92.7, 94.0] & 368 \\
GLM-5 & open & cloud (low) & 93.4 [92.7, 94.1] & 3{,}333 \\
Kimi k2.7 Code & open & cloud (xhigh) & 93.3 [92.6, 94.0] & 2{,}565 \\
Grok 4.3 & proprietary & cloud (medium) & 93.1 [92.4, 93.8] & 2{,}583 \\
Qwen3.7-Plus & open & cloud & 92.7 [92.0, 93.4] & 794 \\
DeepSeek V4 Pro & open & cloud (high) & 92.4 [91.7, 93.2] & 1{,}984 \\
MiniMax M3 & open & cloud & 90.6 [89.8, 91.4] & 545 \\
Llama 4 Maverick 17B 128E & open & self-host, B300 & 87.6 [86.6, 88.5] & 1{,}062 \\
c4ai Command A 03 2025 & open & self-host, H200 & 85.2 [84.2, 86.3] & 137 \\
NVIDIA Nemotron 3 Super 120B A12B & open & self-host, H200 & 81.9 [80.7, 83.0] & 43.6 \\
Hunyuan A13B & open & self-host, A100 & 78.7 [77.6, 79.9] & 33.1 \\
Mistral Small 3.2 24B & open & self-host, A100 & 77.2 [76.0, 78.4] & 49.1 \\
Olmo 3.1 32B & open & self-host, H100 & 74.0 [72.7, 75.2] & 24.8 \\
Granite 4.1 8B & open & self-host, H200 & 70.2 [68.9, 71.4] & 29.6 \\
Phi 4 Mini & open & self-host, A100 & 64.0 [62.7, 65.3] & 5.8 \\
\bottomrule
\end{tabular}
\\[2pt]\footnotesize\textit{Cost is estimated from each provider's list price applied to measured token usage (cloud), or from GPU-hours (self-hosted), per million items. WSD cleanly separates model capability from Phi-4-Mini (64.0) to GPT-5.5 (95.6).}
\end{table*}

\shortlong{The triage panel was built around GPT-5.5 (\Cref{sec:lexen}), raising an obvious objection: a
benchmark corrected with help from one model might merely measure agreement with it. The leaderboard
therefore gives a cross-family check. With each model at its best reasoning tier, GPT-5.5 leads at 95.6\%, and two
families that played no part in GPT-centered triage land within half a point: Claude-Fable-5 at
95.2\% --- a model released only after the triage had run\footnote{``+ Opus 4.8 fallback'' in the
run label (\Cref{tab:t4}) is a repair pass, not an ensemble: Claude-Fable-5's provider-side safety
filter declined 297 of the 4{,}861 items (6.1\%; empty responses clustered almost entirely in two
source documents), and exactly those items were re-run once with Claude-Opus-4.8 under the
identical prompt and protocol. All reported figures score the mixed run; on the items
Claude-Fable-5 answered itself, accuracy is 95.3\% against the 95.2\% mixed headline, so the
mixture is conservative for the post-triage-release argument.} --- and Gemini-3.1-Pro at 94.9\%.
Overlapping marginal intervals are the wrong test here, so we run a paired
bootstrap over the shared items \citep{du2025}. No pairwise difference among the three survives a
Bonferroni-corrected paired test (GPT-5.5 over Claude-Fable-5 $\Delta=+0.39$, 95\% CI
$[-0.14,+0.95]$, $p=0.20$; over Gemini-3.1-Pro $\Delta=+0.68$, $[+0.08,+1.28]$, $p=0.03$, not
surviving the three-way correction): the top three families are statistically indistinguishable on
the full set. Convergence does not prove independence; it argues against a
purely GPT-specific correction artifact. Cross-family agreement supports the reading that \lexen{}-v1
captures real residual label errors in the reviewed tail; estimating residual errors outside that tail
requires a random control sample (as a computational stand-in, unanimous consensus among the five
non-panel families disputes only 0.35\% of the unreviewed labels, a four-of-five majority 1.1\%;
\limref{}).}{}\journalonly{The triage panel that selected items for lexicographer review was built around GPT-5.5
(\Cref{sec:lexen}), which raises an obvious objection: a benchmark corrected with help from one model
might simply be measuring agreement with that model, and its top score would then be an artifact of
grading one's own work. The leaderboard therefore gives a cross-family check.
With each model reported at its best reasoning tier, GPT-5.5 leads at 95.6\% (95\% CI [95.0, 96.2]),
but two model families that played no part in GPT-centered triage land within half a point:
Claude-Fable-5 at 95.2\% [94.6, 95.8] --- a model released only after the triage had run, so
structurally incapable of having shaped the corrections\footnote{The ``+ Opus 4.8 fallback'' in
this run's label (\Cref{tab:t4}) is a mechanical repair pass, not an ensemble.
Claude-Fable-5's provider-side safety filter declined 297 of the 4{,}861 items (6.1\%) ---
HTTP-successful responses with content-filter stop reasons and empty output on both attempts,
concentrated almost entirely in two source documents (209 items in one Senseval-2 document, 80 in
one SemEval-2013 document), consistent with the model's documented over-sensitive content
filtering --- and exactly those items were re-evaluated once with Claude-Opus-4.8 under the
identical prompt and protocol. Every Claude-Fable-5 figure in the paper (accuracy, cost,
agreement, and the hard subset, where 24 of the 307 items were fallback-served) scores this mixed
run as released. The mixture is conservative for the post-triage-release argument: on the 4{,}564
items Claude-Fable-5 answered itself, accuracy is 95.3\% against the 95.2\% mixed headline (the
fallback-served items score 93.9\%), and excluding the 18 fallback-served items among the 211
corrections leaves the correction-step result essentially unchanged ($+74.6$ points on the
remaining 193, against $+75.8$ on all 211).} --- and Gemini-3.1-Pro at 94.9\% [94.3, 95.5].
Marginal confidence intervals are the wrong test for so close a comparison, so we run a paired
bootstrap over the shared items \citep{du2025}. No pairwise difference among the three survives:
GPT-5.5 over Claude-Fable-5 is $\Delta=+0.39$ points (95\% CI $[-0.14,+0.95]$, McNemar $p=0.20$);
GPT-5.5 over Gemini-3.1-Pro is $\Delta=+0.68$ ($[+0.08,+1.28]$, $p=0.03$), which does not survive a
Bonferroni correction across the three pairwise tests; and Claude-Fable-5 over Gemini-3.1-Pro is
$\Delta=+0.29$ ($[-0.31,+0.89]$, $p=0.38$). The honest reading is that the top three families are
statistically indistinguishable on the full set. On the hard reviewed items where the labels were
genuinely contested the three are likewise not
statistically separable (\Cref{sec:results-ceiling}), though that subset is far smaller --- 307 items
against 4{,}861 --- so its wide, overlapping intervals are in part a loss of statistical power: a
true pairwise gap of up to roughly seven points at fine granularity (four at coarse) would have gone
undetected there at 80\% power. The argument against a purely GPT-specific
correction artifact therefore does not rest on a null result alone. It rests on the direction of the
point estimates: on the hard items the triage family GPT-5.5 is not even on top --- Claude-Fable-5
leads at both fine and coarse granularity (\Cref{tab:t7}) --- and on the same items Claude and Gemini
agree with the human reviewers at least as well as GPT-5.5 does (reviewer$\leftrightarrow$model coarse
$\kappa$ 0.816 and 0.814 against 0.807; \Cref{sec:results-ceiling}), neither of which one would expect
if the corrections were GPT-biased. Together these support the reading that \lexen{}-v1 captures real
residual label errors in the reviewed tail. It is not a substitute for a random unflagged control, and it does
not eliminate shared training-data, WordNet, frequency-prior, or semantic biases across frontier
families. (As a computational stand-in for that control, unanimous consensus among the five
non-panel families disputes only 0.35\% of the unreviewed labels, and a four-of-five majority
1.1\% --- 16 and 51 of 4{,}554 items --- a model-estimated bound discussed in the Limitations.) Each score on the public leaderboard carries a bootstrap 95\% CI, close ranks come with
bootstrap rank ranges, and close pairwise comparisons are assessed with this paired bootstrap.}

\journalonly{\begin{table}[H]
\centering
\small
\caption{Classic supervised baselines: prior reported score vs.\ re-scored on lexEN-v1.}
\label{tab:t5}
\begin{tabular}{llr}
\toprule
\textbf{System} & \textbf{Prior reported} & \textbf{lexEN-v1} \\
\midrule
SANDWiCH & 89.0 (Raganato ALL, BabelNet) & 85.2 \\
ConSeC & 83.2 (Raganato ALL, +WNGE) & 84.9 \\
ESCHER & 80.7 (Raganato ALL) & 81.4 \\
BEM & 79.0 (Raganato ALL) & 79.7 \\
MFS & 65.2 (Raganato ALL) & 61.6 \\
\bottomrule
\end{tabular}
\\[2pt]\footnotesize\textit{The lexEN-v1 column reports the available re-scored outputs under the lexEN gold on the same 4{,}861 fine-grained items. For ESCHER and ConSeC the re-scored predictions come from our reproduced checkpoints (79.6 and 82.9 on Raganato \textsc{all}, within the standard reproduction band; \limref{}). The prior-reported column preserves each original evaluation setting and marks inventory differences where applicable, so prior scores are context rather than cross-row evidence.}
\end{table}
}

\shortlong{A second, mechanical check guards against the harness inflating scores: Llama-3.1-8B zero-shot reads
0.588 on \lexen{} against the 0.559 \citet{basile2025} report on comparable XL-WSD English data ---
within three points under non-identical conditions, the agreement expected when the harness measures
the model rather than biasing it. The supervised baselines then read at face value: SANDWiCH 85.2\%,
ConSeC 84.9\%, ESCHER 81.4\%, BEM 79.7\%, and the most-frequent-sense baseline 61.6\%, the strongest
classic systems \emph{as published} trailing the frontier LLMs by roughly ten points on \lexen{}-v1
--- a gap that is itself partly training-label noise: retrained on relabeled SemCor, the same systems
close about half of it (\Cref{sec:results-training}).}{}\journalonly{A second, more mechanical check guards against the harness silently inflating scores. We run
Llama-3.1-8B zero-shot under the same multiple-choice protocol and read 0.588 on \lexen{}, against
the 0.559 that \citet{basile2025} report for the same model on the comparable XL-WSD English data.
The two figures sit within three points of each other under non-identical conditions, which is the
agreement one expects when the harness is measuring the model rather than introducing a bias of its
own; a harness that quietly advantaged every system would not reproduce an external baseline this
closely. With the harness validated, the supervised baselines in \Cref{tab:t5} can be read at face
value: on \lexen{}-v1, SANDWiCH scores 85.2\%, ConSeC 84.9\%, ESCHER 81.4\%, BEM 79.7\%, and the
most-frequent-sense baseline 61.6\% --- the strongest classic systems \emph{as published} trailing
the frontier LLMs by roughly ten points on \lexen{}-v1, a gap that is itself partly training-label
noise: retrained on relabeled SemCor, the same systems close about half of it
(\Cref{sec:results-training}). \journalonly{The full public leaderboard holds 57 models
across 192 runs and 18 families; \Cref{tab:t4} shows one representative row per family, and the complete table with confidence
intervals and rank ranges is part of the living \sensebench{} release.}}

\subsection{The cost/accuracy frontier spans three orders of magnitude}\label{sec:rn-pareto}

\begin{figure*}[t]
\centering
\includegraphics[width=\linewidth]{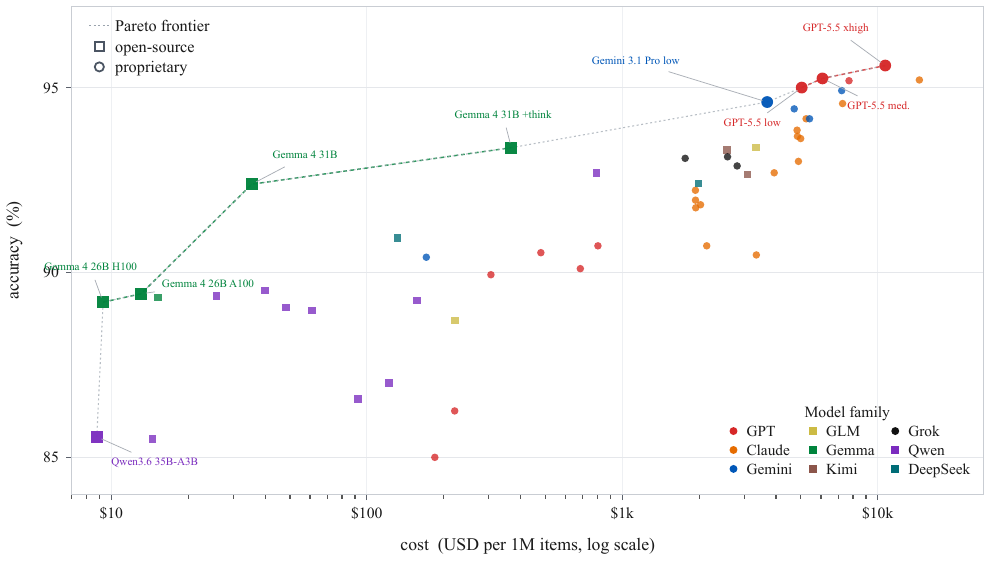}
\caption{Cost/accuracy Pareto frontier for the high-accuracy leaderboard slice. Per-item cost
(log scale, \$ per million items) is plotted against \lexen{} accuracy. Squares mark open-source
models, circles proprietary models, and color denotes model family. All non-dominated frontier
points are labelled; representative context runs are shown for comparison. The plotted slice spans
more than $1{,}200\times$ in cost. Gemma 4 31B reaches 92.4\% at \$35.4 per million, and the last few
points to the top cost steeply more per item.}
\label{fig:f3}
\end{figure*}

\shortlong{At the frontier, accuracy is no longer the most consequential axis. \Cref{fig:f3} plots accuracy
against per-item cost for the high-accuracy slice of the leaderboard, keeping all non-dominated
frontier points and representative context runs. Even within this slice, the price axis spans more
than $1{,}200\times$: from \$8.8 per million items (Qwen3.6 35B-A3B, 85.6\%) to \$10{,}700 per
million at the accuracy frontier (GPT-5.5 at \texttt{xhigh} reasoning, 95.6\%). The same task,
answered to within a few points of the same accuracy, can cost orders of magnitude more or less
depending on the system.}{}\journalonly{Accuracy is no longer the only axis on which these systems differ, and at the frontier it is no
longer the most consequential one. \Cref{fig:f3} plots accuracy against per-item cost for the
high-accuracy slice of the leaderboard, keeping all non-dominated frontier points and representative
context runs rather than the low-accuracy tail. Even within this slice, the price axis spans more than
$1{,}200\times$: from \$8.8 per million items (Qwen3.6 35B-A3B, 85.6\%) to \$10{,}700 per million at
the accuracy frontier (GPT-5.5 at the \texttt{xhigh} reasoning setting, 95.6\%). Reasoning effort is
itself part of this span: the same GPT-5.5 reads 95.0\% at \texttt{low} for \$5{,}040 and 95.6\% at
\texttt{xhigh} for \$10{,}700, so the final half-point of the leader costs roughly twice as much per
item. The same task, answered to within a few points of the same accuracy, can cost orders of
magnitude more or less depending on the system.}

\shortlong{The frontier's shape is what matters at scale: the curve is steep then flat, so most achievable
accuracy is available well before the most expensive systems. An open model, Gemma 4 31B, reaches
92.4\% at \$35.4 per million, within about three points of the 95.6\% frontier; closing most of that
gap with GPT-5.5 at medium effort costs \$6{,}077 per million, roughly $172\times$ more per item. At
the billions-of-words scale that motivates large-scale WSD, that multiplier inverts the calculation
and a near-frontier open model becomes the rational default. The practical frontier of WSD has moved
from accuracy to cost, a theme we return to in \Cref{sec:results-ceiling} and the discussion.}{}\journalonly{The shape of the frontier is what matters for anyone running WSD at scale. The curve is steep and
then flat: accuracy rises quickly through the cheap and mid-priced models and then bends over, so
that most of the achievable accuracy is available well before the most expensive systems. An open
model, Gemma 4 31B, reaches 92.4\% at \$35.4 per million items, within about three points of the 95.6\%
frontier; closing most of that gap with GPT-5.5 at medium effort costs \$6{,}077 per million, roughly
$172\times$ more per item. \journalonly{For a one-off academic evaluation the absolute prices
are trivial and the frontier model is the obvious choice. For the applications that motivate
large-scale WSD --- dictionary construction, corpus annotation, the lexical layer of a learning
system, all of which run to billions of words --- that multiplier on the last three points
inverts the calculation, and a near-frontier open model becomes the rational default.} The practical
frontier of WSD, in other words, has moved from accuracy to cost, a theme we return to in
\Cref{sec:results-ceiling} and the discussion.}

\subsection{Reasoning effort is a first-class axis}\label{sec:rn-effort}

\shortlong{Reasoning effort is not uniform across vendors: GPT-5.5 and Claude expose low/medium/high/xhigh,
Gemini-3.1-Pro caps at high, and most models offer a single setting --- so \Cref{tab:t4} necessarily
reports each model at its \emph{own} best tier, and cross-model gaps there mix tiers. \Cref{tab:teffort}
holds the comparison fixed. Effort buys little at the top (GPT-5.5 moves
95.0$\to$95.6 from low to xhigh) and can even cost accuracy (Grok 4.3 declines with effort), while it
helps a weak model sharply (GPT-5 Nano 77.0$\to$84.3). At the matched \texttt{high} tier --- the
common ceiling of the top three --- GPT-5.5 (95.2), Gemini-3.1-Pro (94.9), and Claude-Opus-4.8 (94.2)
span exactly one point, the same order as the best-tier leaderboard: the ranking is not an artifact of
unequal effort.}{}\journalonly{A leaderboard that reports each model at its strongest configuration must be read with one caveat:
reasoning effort is not uniform across vendors. GPT-5.5 and Claude-Opus-4.8 expose four tiers
(low/medium/high/xhigh), Gemini-3.1-Pro caps at high, and most models offer a single setting, so the
headline \Cref{tab:t4} puts each model at its own best tier and a cross-model gap there can conflate
capability with effort. \Cref{tab:teffort} makes effort an explicit axis for
the seven cloud models with a tier ladder (the two open Gemma thinking-toggle models are reported
in the table note). Two regularities stand out. First, effort has sharply
diminishing returns with capability: it is worth only $+0.6$ points for GPT-5.5 (95.0 at low to 95.6
at xhigh) and actually \emph{costs} accuracy for Grok 4.3 (93.1 at low to 92.9 at high), while it
lifts the weak GPT-5 Nano by more than seven points (77.0 to 84.3). Second, and more important for the
leaderboard, comparing the top three at the matched \texttt{high} tier --- the highest tier all three
share --- preserves the ranking: GPT-5.5 95.2, Gemini-3.1-Pro 94.9, Claude-Opus-4.8 94.2, a one-point
spread in the same order as the best-tier table. The leader's advantage is therefore not an artifact of
its having one more reasoning tier available than Gemini. We report effort as a measured axis rather
than fold it silently into a single number, and the paired comparisons in \Cref{sec:rn-board} hold
each model at its own best tier.}

\begin{table}[H]
\centering
\small
\caption{\textbf{Reasoning effort is a first-class axis.} Best fine-grained \lexen{} accuracy (\%) under prompt \texttt{p001}, per reasoning tier, for the cloud models that expose more than one tier. Comparisons are only valid \emph{within} a column (same effort across models); a dash means the vendor does not offer that tier.}
\label{tab:teffort}
\begin{tabular}{lrrrr}
\toprule
\textbf{Model} & \textbf{Low} & \textbf{Medium} & \textbf{High} & \textbf{xHigh} \\
\midrule
GPT-5.5 & 95.0 & 95.2 & 95.2 & 95.6 \\
Gemini 3.1 Pro & 94.6 & 94.4 & 94.9 & -- \\
Claude Opus 4.8 & 93.7 & 93.6 & 94.2 & 94.6 \\
Grok 4.3 & 93.1 & 93.1 & 92.9 & -- \\
Claude Sonnet 5 & 92.2 & 92.0 & 91.8 & 91.8 \\
GPT-5 Mini & 89.9 & 90.5 & 90.7 & -- \\
GPT-5 Nano & 77.0 & 83.6 & 84.3 & -- \\
\bottomrule
\end{tabular}
\\[2pt]\footnotesize\textit{The frontier's headline run is its best available tier, so cross-model gaps in Table~\ref{tab:t4} mix tiers; at the matched \emph{high} tier the top three (GPT-5.5 95.2, Gemini 3.1 Pro 94.9, Claude Opus 4.8 94.2) span 1.0 point. Gemma~4 (open) instead exposes a binary reasoning toggle, reported separately.}
\end{table}

\subsection{What in the prompt actually helps}\label{sec:rn-ablation}

\shortlong{The rich prompt (\texttt{p001}) beats the minimal one (\texttt{p002}) almost everywhere, but the two
differ along several axes at once (\Cref{tab:t9}), so the leaderboard-wide gap is suggestive rather
than diagnostic. To attribute the gain we ran a controlled single-variable ablation (the registered \texttt{p003}-onward prompts) adding each
component back to \texttt{p002} in isolation, on the two open Gemma models we fully configure, Gemma 4 26B and Gemma 4 31B.}{}\journalonly{The leaderboard's rich prompt (\texttt{p001}) beats the minimal prompt (\texttt{p002}) almost
everywhere, but the two prompts differ along several axes at once (context window, synonyms, sense
keys, system message, output format), and the leaderboard-wide comparison mixes runs at different
reasoning tiers (\Cref{tab:t9} therefore holds the tier fixed within each row). The
leaderboard-wide gap is suggestive rather than
diagnostic, and to attribute the gain to any one factor we ran a controlled single-variable
ablation --- the registered \texttt{p003}-onward prompts (\Cref{sec:sb-prompts}) --- that adds each component back to \texttt{p002} in isolation, on the two open Gemma models, Gemma 4 26B and Gemma 4 31B, where
we control the full configuration.}

\begin{table*}[t]
\centering
\small
\caption{Prompt/context ablation with reasoning effort held fixed within each row: minimal (p002) vs.\ 5+1-context (p001), accuracy (\%) and cost (\$ per million items). $\Delta$\$ is p001's relative cost premium over p002.}
\label{tab:t9}
\begin{tabular}{llrrrrrr}
\toprule
\textbf{Model} & \textbf{Reasoning} & \textbf{p002 acc} & \textbf{p001 acc} & \textbf{$\Delta$acc} & \textbf{p002 \$/M} & \textbf{p001 \$/M} & \textbf{$\Delta$\$ (\%)} \\
\midrule
GPT-5.5 & high & 94.2 & 95.2 & +1.0 & 5{,}625 & 7{,}718 & +37 \\
Claude Opus 4.8 & high & 91.8 & 94.2 & +2.4 & 2{,}121 & 5{,}243 & +147 \\
Gemma 4 31B & on & 92.5 & 93.4 & +0.9 & 176 & 368 & +109 \\
Gemma 4 31B & off & 91.2 & 92.4 & +1.2 & 11.5 & 35.4 & +208 \\
GLM-5 & low & 92.2 & 93.4 & +1.2 & 3{,}246 & 3{,}333 & +3 \\
Qwen3.7-Plus & none & 91.5 & 92.7 & +1.2 & 659 & 794 & +20 \\
Kimi K2.5 & none & 92.4 & 92.7 & +0.3 & 2{,}615 & 3{,}091 & +18 \\
DeepSeek V4 Pro & high & 91.5 & 92.4 & +0.9 & 1{,}338 & 1{,}984 & +48 \\
DeepSeek V4 Flash & high & 89.7 & 90.9 & +1.2 & 88.6 & 132 & +49 \\
Claude Sonnet 4.6 & low & 88.9 & 90.7 & +1.8 & 809 & 2{,}141 & +165 \\
GPT-5 Mini & medium & 90.0 & 90.5 & +0.5 & 363 & 480 & +32 \\
\bottomrule
\end{tabular}
\\[2pt]\footnotesize\textit{Relative to p001, p002 also drops the surrounding context (5+1$\rightarrow$single sentence), WordNet sense keys, synonyms and the system prompt, and switches JSON output to a plain integer (definitions are kept). $\Delta$\$ (\%) $=$ (p001$-$p002)/p002. \textbf{Reasoning} is the reasoning-effort tier (\emph{high}/\emph{medium}/\emph{low}; \emph{none} for non-reasoning models; \emph{on}/\emph{off} for the open Gemma~4 31B thinking toggle) held fixed across \emph{both} prompts in that row: for each model we use the highest tier for which both p001 and p002 were run, so the accuracy and cost deltas isolate the prompt change rather than conflating it with reasoning effort. This can make p001 here lower than the leaderboard headline, which takes each model's best tier even where p002 was not run there (GPT-5.5 and Claude-Opus-4.8 reach 95.6 and 94.6 only at \emph{xhigh}, which we ran for p001 alone). Gemma~4 31B is listed at both toggle settings.}
\end{table*}

\shortlong{The ablation localizes the effect. The \emph{5+1 context window is the single largest driver}, worth
$+0.91$ points on Gemma 4 26B and $+0.78$ on Gemma 4 31B (more than any other component, on its own
roughly half the gap). The sense glosses' \emph{synonyms} are the best value per unit cost ($+0.72$/$+0.54$
at almost no cost), while forcing \emph{JSON output} actively hurts ($-1.14$/$-0.82$). No single
factor accounts for the whole gap, consistent with the leaderboard-wide pattern: the rich prompt
beats the minimal fairly uniformly (GPT-5.5 by 1.0, Claude-Opus-4.8 by 2.4, DeepSeek V4 Flash by 1.2,
Kimi by 0.3; \Cref{tab:t9}). The per-factor decomposition is measured on the two Gemma models alone,
a limitation we state in \limref.}{}\journalonly{The single-variable ablation localizes the effect. Starting from the minimal prompt and adding one
factor at a time, the \emph{5+1 context window is the single largest driver}, worth $+0.91$ points on
Gemma 4 26B and $+0.78$ on Gemma 4 31B (more than any other component) and, on its own, roughly
half of the full p001-over-p002 gap (about 46\% of the gap on the 26B model, 68\% on the 31B). The
sense glosses' \emph{synonyms} are the best value per unit cost, adding $+0.72$ and $+0.54$ points
while costing almost nothing to include, where richer additions such as the WordNet sense keys buy
little accuracy ($+0.01$/$+0.31$) at a steep price (a 45\% cost increase on the 31B model). Two
factors actively hurt: forcing JSON output costs $-1.14$ and $-0.82$ points, and the more elaborate
structured user prompt costs $-0.44$ and $-0.27$. No single factor accounts for the whole gap, which
is consistent with the leaderboard-wide pattern: across the main systems the rich prompt beats the
minimal one fairly uniformly (GPT-5.5 by 1.0 point, Claude-Opus-4.8 by 2.4,
DeepSeek V4 Flash (a distinct, cheaper family) by 1.2, Kimi by 0.3; \Cref{tab:t9}),
corroborating that context and gloss enrichment help in general, even
though the per-factor decomposition is measured on the two Gemma models alone. That two-model scope is
a real limitation of the controlled ablation, and we state it as such in \limref: the
single-variable attributions hold for the systems we could fully configure, and the leaderboard-wide
deltas indicate, but do not prove, that they generalize.}

\shortlong{A small combination study on the same two models confirms it. The two dominant factors, context and
synonyms, together recover most of the gap on their own: adding both to \texttt{p002} closes $75\%$
of the gap on Gemma 4 26B ($+1.54$ points) and $94\%$ on 31B ($+1.18$), at a fraction of
\texttt{p001}'s cost; a second example lifts 26B to $96\%$. Complementary trims confirm it from the
other side: dropping the WordNet sense keys and JSON output from \texttt{p001} saves roughly a third
of the cost while staying flat or positive on 31B. The gap is thus driven by a small, identifiable
bundle rather than by the full richness of \texttt{p001}.}{}\journalonly{Because no single factor accounts for the full gap, we ran a small combination study on the same two
Gemma models, building the cheap prompt up from \texttt{p002} and trimming the rich prompt down from
\texttt{p001}. The two factors that dominate the single-variable analysis, context and synonyms,
together recover most of the gap on their own: adding both to \texttt{p002} closes $75\%$ of the
p001-over-p002 gap on Gemma 4 26B ($+1.54$ points) and $94\%$ on Gemma 4 31B ($+1.18$), at a fraction
of \texttt{p001}'s cost. Target metadata adds essentially nothing on top ($+1.58$/$+1.18$,
statistically tied with context-plus-synonyms). A second example is the only further ingredient that
helps, and only on the smaller model: it lifts the 26B combination to $+1.96$ points ($96\%$ of the
gap, nearly matching \texttt{p001}) while moving 31B only to $+1.27$ ($101\%$). The complementary
trims confirm it from the other side: removing the WordNet sense keys from \texttt{p001} cuts cost by
roughly a fifth to a quarter for negligible accuracy change ($-0.21$ on 26B, $+0.05$ on 31B), and
additionally dropping JSON output saves about a third of the cost while staying flat or positive on
31B ($+0.11$ points at $-33\%$ cost). The gap is thus driven by a small, identifiable bundle, context
and synonyms plus a second example on smaller models, rather than by the full richness of
\texttt{p001}. \journalonly{The combination study shares the single-variable ablation's two-Gemma
scope, and the caveat in \limref{} applies to it in the same way.}}

\subsection{Candidate ordering is a disclosed prior, not the result}\label{sec:rn-ordering}

\shortlong{\sensebench{}'s leaderboard prompt lists each item's WordNet candidates in frequency order, a
deliberate and openly registered choice (\Cref{sec:sb-prompts}) that encodes a real prior: a blind
``pick the first candidate'' rule already scores \textbf{61.6\%} on \lexen{}. A skeptic could
reasonably ask whether the frontier accuracies ride that prior rather than the meaning, so we built
the matching control. Prompt \texttt{p004} is identical to \texttt{p001}---same instruction, context,
candidate definitions, examples, synonyms, sense keys, output format, and parser---except that the
candidates are placed in a fixed random order, which changes the order on 87.6\% of items while
leaving length and content untouched. Re-scoring five frontier configurations (\Cref{tab:tord}),
accuracy is preserved: the mean change is $-0.23$ pp, four of the five paired bootstrap intervals
include zero, and only Claude-Opus-4.8 shows a clear drop ($-0.76$ pp, McNemar $p=0.006$). The
mechanism is the diagnostic part. Under randomization the first candidate is correct only 25.2\% of
the time, yet the models do not keep choosing it: their first-position choice rate falls from about
58\% to about 23\% and redistributes toward later indices, they predict the same WordNet sense on
roughly 96\% of items, and when the originally chosen sense moves to a new index they still select it
on 93.8--96.0\% of cases. Candidate order is thus a small, bounded prior---genuinely helpful, most of
all for Claude---not the source of frontier accuracy: \texttt{p001} scores reflect sense
understanding, not a position shortcut.}{}\journalonly{\sensebench{}'s leaderboard prompt lists each item's WordNet candidates in frequency order, most
frequent first. This is a deliberate and openly registered rendering choice (\Cref{sec:sb-prompts}),
and it encodes a genuine prior: WordNet's frequency ranking is informative, so a blind ``always pick
the first candidate'' rule already scores \textbf{61.6\%} on \lexen{}. A careful reader could
therefore ask whether the frontier accuracies are partly an artifact of frequency-ordered answer
choices---models exploiting position rather than resolving meaning. The question deserves a direct
answer rather than a wave of the hand, so we built the matching ablation. Prompt \texttt{p004} is
identical to \texttt{p001} in every respect---the same linguist instruction, the same 5+1 sentences
of context, the same candidate definitions, examples, synonyms, and sense keys, the same JSON output
and parser---except that the candidate senses are presented in a fixed random order rather than by
frequency. The permutation changes the candidate order on 87.6\% of items while leaving prompt length
and information content untouched, so any movement is attributable to position alone.}

\begin{table*}[t]
\centering
\small
\caption{Candidate-order ablation on five frontier configs (lexEN-v1, 4{,}861 items): WordNet-frequency-ordered candidate senses (p001) vs.\ a fixed random order (p004), otherwise identical prompts. \emph{Same key} is the share of items on which the model predicts the same WordNet sense under both orders.}
\label{tab:tord}
\begin{tabular}{lrrrcr}
\toprule
\textbf{Model / config} & \textbf{p001} & \textbf{p004} & \textbf{$\Delta$ (pp)} & \textbf{95\% CI (pp)} & \textbf{Same key (\%)} \\
\midrule
GPT-5.5 (medium) & 95.25 & 95.15 & -0.10 & [-0.58, +0.37] & 95.9 \\
Gemini 3.1 Pro (high) & 94.92 & 94.53 & -0.39 & [-0.86, +0.06] & 96.5 \\
Claude Opus 4.8 (xhigh) & 94.57 & 93.81 & -0.76 & [-1.30, -0.23] & 94.7 \\
Gemini 3.5 Flash & 94.16 & 94.20 & +0.04 & [-0.43, +0.53] & 95.8 \\
Gemma 4 31B (thinking) & 93.38 & 93.44 & +0.06 & [-0.45, +0.58] & 95.1 \\
\bottomrule
\end{tabular}
\\[2pt]\footnotesize\textit{p004 removes WordNet's frequency prior (the order changes on 87.6\% of items). A pick-candidate-\#1 rule would score 61.6\% under p001 but only 25.2\% under p004, yet measured accuracy is preserved (mean $-0.23$ pp; only Claude-Opus-4.8's interval excludes zero, McNemar $p=0.006$). Even when the p001-chosen sense moves to a new index, models keep it on 93.8--96.0\% of cases, so p001 accuracy reflects sense understanding, not a frequency/position shortcut.}
\end{table*}

\journalonly{Re-running five frontier configurations under \texttt{p004} (\Cref{tab:tord}), the leaderboard barely
moves. The mean accuracy change is $-0.23$ pp and the median $-0.10$ pp; four of the five paired
bootstrap intervals include zero, and the lone clear effect is Claude-Opus-4.8, which loses 0.76 pp
(paired interval $[-1.30,-0.23]$, McNemar $p=0.006$). Every tested system stays inside the frontier
cluster, between 93.4\% and 95.2\%, and two models improve fractionally. If the frontier numbers were
riding the frequency prior, removing it should have been close to catastrophic, because the first
candidate's correctness collapses from 61.6\% under \texttt{p001} to 25.2\% under \texttt{p004};
instead almost nothing happens.}

\journalonly{The per-item behavior explains why. The models do not keep choosing the first option once it stops
being the frequent sense: their first-position choice rate falls from roughly 58\% under frequency
order to roughly 23\% under randomization and redistributes toward later indices, exactly as it
should if a system is tracking a particular sense as that sense moves down the list. On average the
models predict the \emph{same} WordNet sense key under both orders on about 96\% of items, and---most
diagnostic of all---when the sense a model chose under \texttt{p001} is relocated to a different index
under \texttt{p004}, the model still chooses that same sense on 93.8--96.0\% of those moved cases.
Candidate ordering is therefore a small, bounded prior: a real and disclosed aid, largest for
Claude-Opus-4.8, but not the mechanism behind frontier-level accuracy. We keep frequency ordering as
the default precisely because it is a mild, honest help, and we register \texttt{p004} as the control
showing that \texttt{p001} accuracy reflects sense understanding rather than a frequency-or-position
shortcut.}

\subsection{The gains are item-level corrections}\label{sec:rn-sensitivity}

\shortlong{Does cleaning the labels fix specific answers, or just nudge an aggregate percentage? We grade each
system twice on the same 4{,}861 items --- against \maru{}, then against corrected \lexen{} --- and
count, item by item, which way each grade flips: \emph{vindications} (wrong before, right now) versus
\emph{new penalties} (right before, wrong now). A paired test (McNemar) is sharply one-sided for the
LLMs: GPT-5.5 gets \textbf{190} vindications against only \textbf{5} new penalties ($\chi^2=173.6$,
$p\approx9\times10^{-50}$), Claude-Fable-5 the same (177 vs.\ 17). The supervised systems barely move, their
flips cancelling (ConSeC 113 vs.\ 79, $p\approx0.02$; BEM 95 vs.\ 79, not significant). So the
\maru{}$\to$\lexen{} jump of \Cref{sec:rn-triple} is real item-level repair: the LLMs were already
giving the sense the lexicographers later endorsed, with almost nothing lost in return, while the
supervised systems stay put.}{}\journalonly{To check that the \maru{}$\to$\lexen{} gains of \Cref{sec:rn-triple} are real fixes to specific answers
rather than a percentage drifting upward for aggregate reasons, picture grading the same 4{,}861 items
twice --- under the old \maru{} key, then the corrected \lexen{} key --- and counting, for each item,
which way the grade flips: a \emph{vindication} (wrong under the old key, right under the new; $c$
items) or a \emph{new penalty} (right then wrong; $b$ items). A paired test (McNemar) compares the two
counts, and for the LLMs they are heavily lopsided: correcting the key gives GPT-5.5 \textbf{190}
vindications against only \textbf{5} new penalties ($\chi^2=173.6$, exact two-sided $p\approx9\times10^{-50}$),
and Claude-Fable-5 the same shape ($b=17$, $c=177$). On roughly 190 items GPT-5.5 was already giving
the sense the lexicographers later adjudicated correct; the old key simply marked it wrong, with almost
nothing it had right lost in the trade. The supervised systems barely move, their flips cancelling and
dominated by items wrong under both keys (ConSeC $b=79$, $c=113$, $+0.7\%$, $p\approx0.02$; BEM $b=79$,
$c=95$, $p=0.26$, not significant). The net movement $(c-b)/N$ reproduces the marginal
\maru{}$\to$\lexen{} deltas of \Cref{tab:t6} exactly, so the headline gain is fully accounted for by
these item-level flips --- the correction lifts the LLMs through many per-item fixes with almost no
regressions, while leaving the supervised systems essentially where they were.}
 \section{Results II: Repairing the Training Labels}\label{sec:results-training}

\shortlong{Section~\ref{sec:results-noise} located the noise in the \emph{test} labels and showed it now governs
the ranking. It also left a loose end: under label correction the frontier models gained sharply while
the classic supervised systems barely moved (\Cref{sec:rn-triple}). If the test labels were the only
problem, those systems should have gained too. They did not --- because for them the binding noise is
upstream, in the corpus they are \emph{trained} on. This section closes the loop. We relabel the
SemCor training corpus with a frontier model, retrain three standard supervised systems with no other
change, and measure the effect on held-out test sets. The same architectures that defined the
79--83\% plateau rise by several points. The noise was on both ends of the pipeline, and it is
repairable.}{Section~\ref{sec:results-noise} located the noise in the \emph{test} labels and showed that, once
systems clear 90\%, it governs the ranking. It also left a loose end. Under successive label
correction the frontier language models gained ten points while the classic supervised systems
gained barely one on the step that isolates \lexen{}'s own correction (\Cref{sec:rn-triple}). We read
that gap as diagnostic of where each class of system's residual errors fall. But it raises an obvious
question the test-side analysis cannot answer: if correcting the labels barely helps the supervised
systems, what \emph{was} holding them at 79--83\% for a decade --- architecture, or the data they were
trained on? This section answers it, for both axes. We relabel the SemCor training corpus with a frontier model,
retrain three standard supervised systems changing nothing but the training labels, and measure the
effect on test sets the relabeling never touched. The same architectures that defined the plateau
rise by several points. The label noise was on both ends of the pipeline --- the test set we score
against and the corpus we learn from --- and, unlike the inventory granularity we turn to in
\Cref{sec:results-ceiling}, it is repairable.}

\subsection{Relabeling SemCor}\label{sec:rt-relabel}

\shortlong{We relabel all \textbf{226{,}036} instances of SemCor --- the corpus on which BEM, ESCHER, and ConSeC
are trained --- with the same immutable \lexen{} \texttt{p003} prompt used for evaluation
(\Cref{sec:sensebench}): the target in a 5+1-sentence window, the full WordNet candidate set with
glosses, synonyms, and examples, and a single sense index returned. Each instance keeps its original
SemCor gold key alongside the new label, so the result is a drop-in training corpus aligned $1{:}1$
with the original. To our knowledge SemCor has not previously been relabeled wholesale; prior
responses to the sense-annotation bottleneck \citep{pasini2020} generate new data rather than
repairing the existing corpus. We run two relabelers --- GPT-5.5 and the open Gemma 4 31B, at a one-time cost of
\$866 and \$155 respectively --- and release both
corpora, \textbf{SemCor-GPT5.5} and \textbf{SemCor-Gemma}, both publicly available (link in the de-anonymized version).}{We relabel all \textbf{226{,}036} sense-annotated instances of SemCor --- the corpus on which BEM,
ESCHER, and ConSeC are all trained --- with the same immutable \lexen{} \texttt{p003} prompt used for
evaluation (\Cref{sec:sensebench}): each target word is presented in a 5+1-sentence context
window with the full set of WordNet candidate senses for its lemma and part of speech, each with its
gloss, synonyms, and examples, and the model returns a single sense index. We run two relabelers
independently, producing two released corpora --- \textbf{SemCor-GPT5.5} (GPT-5.5; a one-time
relabeling cost of \$866) and
\textbf{SemCor-Gemma} (the open Gemma 4 31B; \$155). Every instance retains its original SemCor gold
key beside the new one, so each relabeled corpus is a drop-in replacement aligned $1{:}1$ with the
original SemCor, suitable for a controlled A/B comparison in which the training labels are the only
thing that changes. To our knowledge SemCor has not previously been relabeled wholesale; earlier
responses to the sense-annotation bottleneck \citep{pasini2020} generate new annotations rather than
repairing the existing corpus. Both corpora are publicly released at \url{https://github.com/GliteTech/research-semcor-relabeling}.}

\begin{figure}[t]
\centering
\includegraphics[width=\shortlong{\linewidth}{0.8\linewidth}]{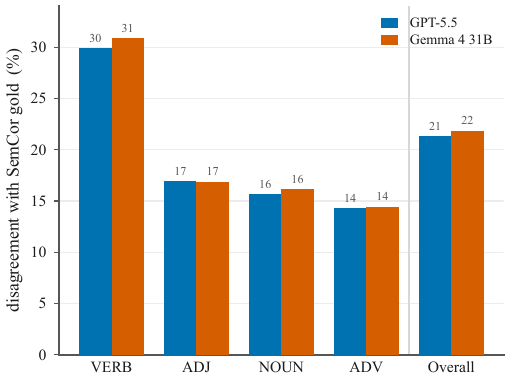}
\caption{Fraction of the 226{,}036 SemCor training instances whose relabel disagrees with the
original gold, by part of speech and overall, for two independent relabelers. Verbs dominate
($\sim$30\%) and the two models trace a near-identical profile --- the disagreement is a property of
the corpus, not of one model.}
\label{fig:fdisagree}
\end{figure}

\shortlong{The relabeling disagrees with the original SemCor gold on \textbf{21.30\%} of instances under GPT-5.5
and \textbf{21.85\%} under Gemma, and the two models disagree with the corpus in almost exactly the
same places (\Cref{fig:fdisagree}): the rate is highest on verbs (\textbf{29.90\%} / \textbf{30.86\%})
and lower and similar on nouns, adjectives, and adverbs. That both an OpenAI and an open Google model --- different developers, one proprietary and one open --- disagree with SemCor most on verbs indicates the disagreement is a property of the corpus, not an idiosyncrasy of
one relabeler. We report this as model--corpus \emph{disagreement}, not a verified error rate: as in
\Cref{sec:lexen}, we make no claim that every changed label is a corpus error. The evidence that the
changes are, on balance, improvements is downstream, in the retrained models.}{The relabeling disagrees with the original SemCor gold on \textbf{21.30\%} of instances under GPT-5.5
and \textbf{21.85\%} under Gemma, and --- the more telling fact --- the two models disagree with the
corpus in almost exactly the same places (\Cref{fig:fdisagree}). Both rank the parts of speech
identically: the disagreement rate is highest on verbs (\textbf{29.90\%} for GPT-5.5, \textbf{30.86\%}
for Gemma) and substantially lower, and close to one another, on nouns, adjectives, and adverbs. That
this profile reproduces across an OpenAI model and an open Google model --- different developers, one proprietary and one open --- is evidence that the disagreement
reflects a property of the corpus rather than the taste of any one relabeler. We are careful about
what the 21\% means: as with the reviewed subset in \Cref{sec:lexen}, this is model--corpus
\emph{disagreement}, not a verified corpus error rate, and we make no claim that every changed label
is wrong in SemCor. The evidence that the changes are, in aggregate, corrections rather than noise is
not in the disagreement count; it is downstream, in what happens when a model is retrained on them.}

\shortlong{The disagreements concentrate where \Cref{sec:results-ceiling} finds the inventory hardest. The two
boxes below illustrate the kind of change the relabeling makes; unlike the corrections in
\Cref{sec:lexen}, these are author-selected illustrations adjudicated by us, not by the three
lexicographers, and we mark them as such.}{The disagreements concentrate, by part of speech, exactly where \Cref{sec:results-ceiling} will show
the WordNet inventory is hardest even for professional lexicographers --- the verbs. The two boxes
below give a feel for the changes. We stress that, unlike the \lexen{} corrections in
\Cref{sec:lexen}, these are not three-reviewer adjudications: they are author-selected illustrations,
judged by us, and we present them as illustrative of the disagreement signal rather than as
independently certified corrections.}

\begin{examplebox}[E-SC1: \emph{restrict} --- a fine verb split (VERB; author-judged)]
\textit{``\ldots{}Can you consider \tw{restricting} any additional employee benefits to those paid for
by profit-sharing money\ldots{}''} \;(\texttt{semcor.d000.s013.t001})\\[3pt]
SemCor: \sk{restrict\%2:30:02::} ``place restrictions on; curtail'' $\rightarrow$
relabel: \sk{restrict\%2:30:00::} ``place limits on (extent or access)''.\\[3pt]
The two senses are near-paraphrases; here the target limits \emph{which} benefits qualify, which reads
more naturally as placing a limit on extent than as curtailing an activity. \emph{Verdict (authors):}
a hair-fine verb distinction of the kind that drives the 29.9\% verb disagreement rate --- exactly
where \Cref{sec:results-ceiling} finds the inventory over-specifies.
\end{examplebox}

\begin{examplebox}[E-SC2: \emph{improved} --- a plausible correction (ADJ; author-judged)]
\textit{``\ldots{}one that has the goal of \tw{improved} employee morale and, consequently, increased
productivity\ldots{}''} \;(\texttt{semcor.d000.s001.t007})\\[3pt]
SemCor: \sk{improved\%3:00:00::} ``made more desirable or valuable or profitable'' $\rightarrow$
relabel: \sk{improved\%5:00:00:better:00} ``become or made better in quality''.\\[3pt]
``Improved morale'' is morale made \emph{better in quality}, not made more marketable; the relabel
reads as the more natural sense. \emph{Verdict (authors):} a candidate correction, not merely a tie ---
the kind the aggregate retraining result suggests is, on balance, an improvement.
\end{examplebox}

\subsection{Retraining lifts the classic systems}\label{sec:rt-retrain}

\shortlong{We retrain BEM, ESCHER, and ConSeC on each relabeled corpus --- GPT-5.5 and the open Gemma 4 31B ---
holding architecture, hyperparameters, and recipe fixed; the only change is the training gold.
\Cref{tab:tretrain} reports accuracy against three test sets for both relabelers; the figures below are
for the GPT-5.5-relabeled corpus. \textbf{Every system improves, on test sets the relabeling never saw.} On the original
Raganato \textsc{all} labels each gains \textbf{+2.4 to +3.4} points (ESCHER 79.57$\to$82.39, ConSeC
81.62$\to$84.06, BEM 77.66$\to$81.06); on the corrected Maru \textsc{all\_new} the gains roughly
double (ESCHER \textbf{+5.48}, ConSeC \textbf{+5.32}). Because nothing but the labels changed, the lift
is causal: the training data was holding these systems well below their own architectures' reach ---
and \Cref{sec:rt-encoder} shows the architecture axis is worth a comparable amount on top.}{We retrain BEM, ESCHER, and ConSeC on each relabeled corpus --- the GPT-5.5-relabeled and the open
Gemma 4 31B-relabeled SemCor --- holding the architecture, hyperparameters, optimizer, and training
recipe of each system fixed at its published configuration; the only variable that changes is the
training gold. \Cref{tab:tretrain} reports the accuracy of the retrained systems against three
evaluation sets for both relabelers; the figures we discuss are for the GPT-5.5-relabeled corpus, with
the Gemma 4 31B-relabeled rows providing a second independent relabeler. \textbf{Every system improves,
and it improves on test sets the relabeling never saw.} On the original Raganato \textsc{all} labels --- the standard
benchmark --- each of the three gains between \textbf{+2.4 and +3.4} points (ESCHER 79.57$\to$82.39,
ConSeC 81.62$\to$84.06, BEM 77.66$\to$81.06). On the lexicographer-corrected Maru \textsc{all\_new}
labels the gains roughly double (ESCHER \textbf{+5.48}, ConSeC \textbf{+5.32}). Because the architecture
and recipe are held fixed, this is a controlled, single-variable result: the lift is attributable to
the training labels alone. It is the causal complement to the observational test-side analysis of
\Cref{sec:results-noise}, and it answers the question that section could not: what held the classic
supervised systems at 79--83\% was, to a first approximation, the quality of the corpus they were
trained on --- though not it alone. \Cref{sec:rt-encoder} measures the other axis and finds that the
age of the architectures' configurations contributes a comparable, separable amount; the two compose,
and neither was a hard ceiling.}

\begin{table*}[t]
\centering
\small
\caption{\textbf{Repairing the training labels.} Classic supervised systems retrained on relabeled SemCor --- two independent relabelers, GPT-5.5 and the open Gemma 4 31B --- versus the original SemCor labels, with no change to architecture or training recipe --- the only variable is the training gold. Accuracy (\%) on Raganato \textsc{all}, the corrected Maru \textsc{all\_new}, and \lexen{}-v1, on the same items used throughout.}
\label{tab:tretrain}
\begin{tabular}{llrrr}
\toprule
\textbf{System} & \textbf{Training labels} & \textbf{Raganato \textsc{all}} & \textbf{Maru \textsc{all\_new}} & \textbf{\lexen{}-v1} \\
\midrule
BEM & traditional SemCor & 77.66 & 77.16 & 78.87 \\
BEM & GPT-5.5-relabeled & 81.06 & 83.34 & 86.75 \\
BEM & Gemma 4 31B-relabeled & 79.83 & 82.88 & 85.17 \\
\midrule
ESCHER & traditional SemCor & 79.57 & 79.34 & 81.30 \\
ESCHER & GPT-5.5-relabeled & 82.39 & 84.82 & 88.12 \\
ESCHER & Gemma 4 31B-relabeled & 81.61 & 83.49 & 86.88 \\
\midrule
ConSeC & traditional SemCor & 81.62 & 82.25 & 83.30 \\
ConSeC & GPT-5.5-relabeled & 84.06 & 87.57 & 90.72 \\
ConSeC & Gemma 4 31B-relabeled & 84.10 & 87.49 & 90.39 \\
\bottomrule
\end{tabular}
\\[2pt]\footnotesize\textit{The relabeled corpus is produced with the same \lexen{} \texttt{p003} prompt used for evaluation. Retraining lifts every system, and the lift widens as the test labels themselves get cleaner (Raganato $<$ Maru $<$ \lexen{}) --- the training-side counterpart of Table~\ref{tab:t6}. We read the result from the Raganato and Maru columns, which the relabeling never saw; the \lexen{} column, sharing a labeling function with the relabeling, is confirmatory.}
\end{table*}

\subsection{The gains are not model self-agreement}\label{sec:rt-circularity}

\shortlong{A frontier model relabeled the data, so the natural worry is circularity: are the systems just learning
to imitate the relabeler? Three facts say no. First, the gains are measured on \textbf{Raganato \textsc{all}
and Maru \textsc{all\_new}} --- independently human-annotated test sets that predate the frontier models and that the relabeler never touched --- and they appear in
\textbf{non-LLM} systems. Second, the keystone: a model trained on labels that disagree with SemCor 21\%
of the time scores \emph{higher} on Raganato's own original human labels, which it can only do if those
labels are, on balance, better. Third, two independent relabelers produce the near-identical disagreement
signal of \Cref{fig:fdisagree}. We therefore read the result from the Raganato and Maru columns; the
\lexen{} column in \Cref{tab:tretrain}, which shares a model with the relabeling via triage, we treat as
confirmatory only.}{A frontier model produced the training labels, so the central worry is the same circularity that
shadows the test-side results: are the retrained systems simply learning to imitate the relabeler,
in which case the ``improvement'' would be agreement with a model dressed up as accuracy? Three
features of the design rule this out. First, the gains we report are measured on \textbf{Raganato
\textsc{all} and Maru \textsc{all\_new}} --- independently human-annotated test sets that predate the frontier models, that the relabeling process never saw, and that
no model in this paper had a hand in constructing --- and they appear in \textbf{classic, non-LLM}
supervised systems whose only contact with a language model is the corpus they were trained on.
Second, and most directly, the keystone observation: a system trained on labels that disagree with
the original SemCor gold on 21\% of instances nonetheless scores \emph{higher} on Raganato's own
original human-annotated test labels. A model that had merely absorbed a relabeler's idiosyncrasies
would do worse against the original annotation, not better; doing better is what we expect only if the
relabeled training signal is, on balance, closer to the truth. Third, the effect is not specific to
one model: two independent relabelers, one proprietary and one open, produce the near-identical
disagreement profile of \Cref{fig:fdisagree}. For these reasons we read the retraining result from the
Raganato and Maru columns of \Cref{tab:tretrain}; the \lexen{} column, which shares a model
with the relabeling via triage, we treat as confirmatory rather than load-bearing.}

\shortlong{A fourth check targets the rival that the relabeler merely aligned SemCor with
first-sense convention. The opposite holds: the relabels move \emph{away} from
the WordNet first sense (73.7\%$\to$68.1\% first-sense share; net changed-label
flow $-26\%$), sense entropy rises, and when test items are split by whether the
gold is the lemma's first sense, the whole gain sits on \textbf{non-first-sense}
items (+11.7 to +17.6 GPT-5.5, +11.2 to +17.9 Gemma; all McNemar
$p<10^{-26}$), with first-sense items down 0.5--3.3. That is the signature of
repairing noise that over-defaults to sense~1, not of convention alignment.}{A fourth check addresses a subtler rival: that the relabeler did not so much fix
errors as align SemCor with annotation convention --- more first-sense labels,
flattened rare-sense tails --- with the retrained systems then harvesting that
alignment on convention-leaning test sets. The distributions say otherwise. The
relabels move \emph{away} from the WordNet first sense (SemCor's first-sense
share falls from 73.7\% to 68.1\%; among changed labels, half leave the first
sense and under a quarter move to it), per-lemma sense entropy rises slightly
(+0.06 bits; the relabeled corpus attests \emph{more} senses than the original),
and mean hypernym depth is unchanged. Decisively, when every test item is split
by whether its gold label is the WordNet first sense of its lemma, the entire
retraining gain concentrates on the \textbf{non-first-sense} items: +11.7 to
+17.6 points across the three systems and both unseen surfaces for the GPT-5.5
retrains (exact McNemar $p \le 3.8\times10^{-33}$ in every cell), against a
0.5--3.3-point loss on first-sense items, with the Gemma retrains replicating
the pattern (+11.2 to +17.9). Convention alignment predicts the opposite
concentration. This is instead the signature of repairing a corpus whose noise
over-defaults to the first sense: the retrained systems stop over-predicting
sense~1 and recover minority senses --- on Maru's all-non-first-sense 42D subset,
the hardest slice, BEM gains +23.0.}

\begin{figure}[t]
\centering
\includegraphics[width=\shortlong{\linewidth}{0.8\linewidth}]{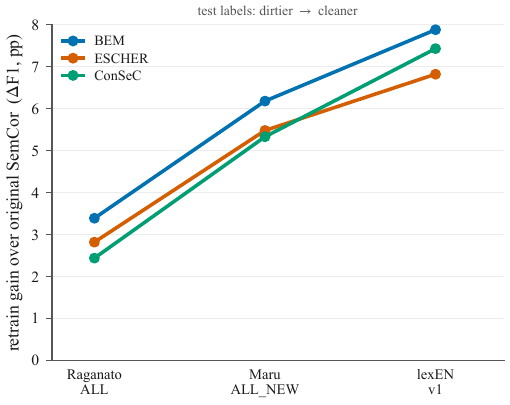}
\caption{Retrain gain of each supervised system (accuracy on GPT-5.5-relabeled training minus original
SemCor training) against three test sets ordered by label cleanliness. The gain grows as the test
labels get cleaner --- the training-side counterpart of the test-side staircase in \Cref{tab:t6}; dirty
test labels under-count the improvement.}
\label{fig:fretrain}
\end{figure}

\shortlong{The gains also have a revealing shape (\Cref{fig:fretrain}): for every system the improvement grows as
the \emph{test} labels themselves get cleaner --- smallest on the original Raganato gold, larger on the
corrected Maru labels, larger still on \lexen{}. This is the training-side mirror of the staircase in
\Cref{tab:t6}. A model trained on better labels is partly penalized when scored against a dirty test
set, for being right where the test gold is wrong; the Raganato gain therefore \emph{under}-counts the
true improvement, and the cleaner the yardstick, the more of it shows.}{The gains also have a shape that reinforces the reading (\Cref{fig:fretrain}). For every retrained
system the improvement grows monotonically as the \emph{test} labels themselves get cleaner: it is
smallest measured against the original Raganato gold, larger against the corrected Maru labels, and
larger still against \lexen{}. This is precisely the training-side mirror of the test-side staircase
in \Cref{tab:t6}, and it has the same explanation. A model trained on cleaner labels is partly
penalized when it is scored against a noisy test set --- it loses credit on exactly those items where
the test gold is itself wrong --- so the Raganato number \emph{under}-counts the true gain, and the
cleaner the measuring stick, the more of the improvement becomes visible. The same label noise that
\Cref{sec:results-noise} showed reshuffles the leaderboard also suppresses the measured benefit of
fixing the training data.}

\subsection{A modern bi-encoder on the repaired labels}\label{sec:rt-encoder}

\begin{table*}[t]
\centering
\small
\caption{\textbf{Labels versus architecture.} A 2$\times$2 over training labels (original vs.\ GPT-5.5-relabeled SemCor) and architecture within the same bi-encoder family: BEM at its published 2020 configuration versus \lens{}, a modernized 298M dual encoder (ModernBERT-base backbone, wider context, current training recipe). Accuracy (\%) on Raganato \textsc{all}, Maru \textsc{all\_new}, and \lexen{}-v1. On Raganato \textsc{all} the two axes contribute comparably and compose.}
\label{tab:tbienc}
\begin{tabular}{llrrr}
\toprule
\textbf{System} & \textbf{Training labels} & \textbf{Raganato \textsc{all}} & \textbf{Maru \textsc{all\_new}} & \textbf{\lexen{}-v1} \\
\midrule
BEM & traditional SemCor & 77.66 & 77.16 & 78.87 \\
BEM & GPT-5.5-relabeled & 81.06 & 83.34 & 86.75 \\
\midrule
\lens{} (ours) & traditional SemCor & 80.84 & 82.14 & 83.52 \\
\lens{} (ours) & GPT-5.5-relabeled & 83.64 & 87.37 & 90.48 \\
\bottomrule
\end{tabular}
\\[2pt]\footnotesize\textit{BEM rows repeat Table~\ref{tab:tretrain}. The \lens{} relabeled row is the mean of three training seeds (largest per-surface range 0.6); the original-SemCor \lens{} row is a single run. As there, we read the result from the Raganato and Maru columns, which the relabeling never saw; the \lexen{} column is confirmatory. On repaired labels \lens{} exceeds every published SemCor-only supervised system (BEM 79.0, ESCHER 80.7, ConSeC 82.0) and is level with ConSeC+WNGE (83.2), which trains on roughly $3\times$ the labelled data.}
\end{table*}

\shortlong{\Cref{sec:rt-retrain} froze the architectures to isolate the labels; here we measure the converse
axis --- what does modernizing the architecture buy at fixed, repaired labels? \lens{} is a
BEM-style bi-encoder rebuilt from today's standard parts: two ModernBERT-base towers
\citep{warner2024modernbert} (298M parameters in total), a 5+1-sentence context window, candidate
glosses enriched with synonyms and examples --- the same presentation the \texttt{p003} prompt gives
the LLMs (\Cref{sec:rn-ablation}) --- and a current contrastive training recipe. Trained on
SemCor-GPT5.5 it reaches 83.6 on Raganato \textsc{all} and 87.4 on Maru
\textsc{all\_new} (\Cref{tab:tbienc}) --- to our knowledge the strongest bi-encoder result reported,
\textbf{+4.6} over published BEM, above every published SemCor-only supervised system, and level with
ConSeC+WNGE (83.2) on a third of its labelled data; the \lexen{} column (90.5) is
confirmatory, per \Cref{sec:rt-circularity}. The $2\times2$ in \Cref{tab:tbienc} shows the label and
architecture axes contribute comparably on Raganato ($+3.4$ labels on BEM; about $+2.6$ architecture
at fixed relabeled labels) and compose to about $+6$ points; the component attribution
(\Cref{sec:app-biencoder}) lands on the backbone, the recipe, and the context window. The design is
also structurally cheap: the gloss side is context-independent, so the sense inventory is embedded
once into a precomputed gallery and inference reduces to one context encoding plus dot products ---
\$0.126 per million items at 90.5\% \lexen{} accuracy (\Cref{sec:discussion-cost}).}{\Cref{sec:rt-retrain} deliberately held the architectures at their published 2020--21 configurations,
because the question there demanded it: only a frozen system isolates the training labels as the
single variable. But the freeze leaves the converse question open: with the labels repaired, how much
does merely \emph{modernizing} the same architecture family buy? To measure that axis we build
\lens{}, a bi-encoder in the mould of BEM \citep{blevins2020}, built from today's standard parts: two
ModernBERT-base towers \citep{warner2024modernbert} of 149M parameters each, one encoding the target
word in a five-preceding-plus-one-following sentence window, the other encoding each candidate
sense's gloss enriched with synonyms and example sentences --- the same context-and-candidate
presentation the \texttt{p003} prompt gives the LLMs (\Cref{sec:rn-ablation}) --- scored by a dot
product and trained with a current contrastive recipe. The simplicity is deliberate and
load-bearing: with every ingredient standard current practice, whatever \lens{} gains over BEM is
attributable to modernization itself rather than to any single invention --- exactly the axis this
subsection sets out to measure.

Trained on SemCor-GPT5.5, \lens{} reaches 83.6 on Raganato \textsc{all} and 87.4
on Maru \textsc{all\_new} (\Cref{tab:tbienc}) --- to our knowledge the strongest bi-encoder result
reported on the standard benchmark, \textbf{+4.6} over published BEM (79.0), above every published
SemCor-only supervised system of any architecture (ESCHER 80.7, ConSeC 82.0), and level with
ConSeC+WNGE (83.2), which trains on roughly $3\times$ the labelled data. As everywhere in this
section, we read the claim from the Raganato and Maru columns, which the relabeling never saw. On
\lexen{}-v1 the model scores 90.5, within striking distance of the frontier band of
\Cref{tab:t4} --- but \lens{} is trained on GPT-5.5 labels, so per \Cref{sec:rt-circularity} we treat
that column as confirmatory only.

\Cref{tab:tbienc} completes a $2\times2$ over the two axes, and on Raganato \textsc{all} they
contribute comparably: repairing the labels moves BEM by $+3.4$; modernizing the architecture at
fixed repaired labels moves it by about $+2.6$ (about $+3.2$ at the original labels); together they
are worth about $+6$ points over the published-configuration BEM. The two cheap ingredients compose,
and neither alone explains the plateau. The component-level attribution (\Cref{sec:app-biencoder})
locates essentially the whole architecture axis in three places --- the backbone, the training
recipe, and the wider context window --- while the structured gloss and the in-batch negatives
contribute little at the margin. One component sits outside the axis altogether: the target-span
pooling that \lens{} inherits from BEM. Replacing it with sentence-level CLS pooling collapses
accuracy by nine points on the BERT-base anchor and by almost nine on the full ModernBERT recipe
(\Cref{sec:app-biencoder}) --- a foundation both stacks require, not a modernization gain. The
components form a co-adapted stack, not a sum of independent parts.

The reason to care about a bi-encoder in 2026, though, is not its score but its cost structure.
Because the gloss tower is context-independent, the entire sense inventory is embedded \emph{once}
into a precomputed gallery, and inference reduces to a single context encoding plus dot products ---
a property that cross-encoders, which must re-read every gloss in context, and LLMs, which must
generate, cannot share. Measured end to end this puts \lens{} at \textbf{\$0.126 per million
disambiguations} at 90.5\% \lexen{} accuracy; we place that number in the cost frontier of
\Cref{sec:discussion-cost}.}

\shortlong{The practical consequence is a deployment recipe. A frontier model is run \emph{once}, offline, to
relabel a training corpus; the corrected corpus then trains a cheap supervised system that runs at a
fraction of the per-item cost (\Cref{sec:rn-pareto}). The frontier model's competence is distilled into
the labels and, through them, into a model that need never be queried at inference time --- the point we
take up in \Cref{sec:discussion-cost}. None of this makes the task solved: the retraining repairs the
\emph{noise}, but the residual disagreement it leaves behind sits on the fine WordNet distinctions that
\Cref{sec:results-ceiling} shows neither models nor experts reliably reproduce.}{The practical consequence is a deployment recipe rather than only a diagnosis. A frontier model is run
\emph{once}, offline, to relabel a training corpus; the relabeled corpus then trains a cheap supervised
system that runs at a small fraction of the frontier's per-item inference cost (\Cref{sec:rn-pareto}).
The frontier model's competence is distilled into the labels, and through them into a model that need
never be queried at inference time --- a route around the cost frontier we develop in
\Cref{sec:discussion-cost}. We are careful not to over-read the result. Retraining on relabeled data
repairs the \emph{noise} in the training signal; it does not, and cannot, dissolve the residual
disagreement that remains, which falls on the fine WordNet distinctions that \Cref{sec:results-ceiling}
shows neither frontier models nor professional lexicographers reproduce consistently. Removing the
fixable problem is what brings the inherent one into view.}
 \section{Results III: Human Agreement and the Granularity Ceiling}\label{sec:results-ceiling}

\shortlong{With the leading systems within a point of one another at 94--95\% (\Cref{sec:results-noise}), the next question is whether the remaining few points are model error, residual label error in the unreviewed items, or an artifact of the granularity at which the benchmark scores. Using the reviewed items --- three independent expert judgments each, which control for residual label error --- we show the residual disagreement is shared between models and lexicographers, concentrates at a granularity the WordNet inventory imposes rather than the task requires, and that on the hard items a top model rates within the reviewers' agreement band.}{The numbers in \Cref{sec:results-noise} put the leading systems within a point of one another at 94--95\%, and the obvious next question is whether the remaining few points are model error, residual label error in the unreviewed part of the test set, or an artifact of the granularity at which the benchmark scores. This section uses the reviewed items --- the only part of the benchmark where we hold three independent expert judgments per item, which removes residual label error as a confound --- to show that the residual disagreement is shared between models and lexicographers, that it concentrates at a granularity the WordNet inventory imposes rather than one the task requires, and that on the hard items a top model rates within the reviewers' agreement band.}

\shortlong{Two populations run through this section. Agreement statistics (\Cref{sec:results-ceiling:iaa}) are over all \textbf{363} reviewed items, from the lexicographers' raw choices (Fleiss and pairwise Cohen $\kappa$); model-versus-human comparisons (\Cref{sec:results-ceiling:envelope}) are over the \textbf{307} retained hard items that survived adjudication and carry a defensible label (mean pairwise Cohen $\kappa$). We label which is in play each time.}{A note on populations, because two distinct ones run through this section and we keep them strictly apart. The agreement statistics in \Cref{sec:results-ceiling:iaa} are computed over all \textbf{363} reviewed items, using the three lexicographers' raw choices (Fleiss and pairwise Cohen $\kappa$). The model-versus-human comparisons in \Cref{sec:results-ceiling:envelope} are computed over the \textbf{307} retained hard items, those that survived adjudication and therefore carry a defensible label, as mean pairwise Cohen $\kappa$. The two populations and the two $\kappa$ estimators are not interchangeable, and we label which is in play every time.}

\subsection{Inter-Annotator Agreement}\label{sec:results-ceiling:iaa}

\shortlong{Fine WordNet WSD is partly ill-posed even for professionals. Over the 363 reviewed items the three lexicographers reach exact three-way agreement on the fine sense only \textbf{35.5\%} of the time, Fleiss $\kappa = \textbf{0.537}$ (\Cref{tab:t3}, \Cref{fig:f5}) --- moderate agreement \citep{landiskoch1977,murray2004}, among experts who chose independently from the same candidate list with written rationales, not crowd workers (no pair exceeds 59\% exact fine agreement). A benchmark scored at that granularity measures, in part, a distinction its own annotators cannot reliably reproduce.}{Fine-grained WordNet sense disambiguation is partly ill-posed even for professionals. Over the 363 reviewed items, the three lexicographers reach exact three-way agreement on the fine WordNet sense only \textbf{35.5\%} of the time, a Fleiss $\kappa$ of \textbf{0.537} (\Cref{tab:t3}, \Cref{fig:f5}). That is moderate agreement by any conventional reading of $\kappa$ \citep{landiskoch1977,murray2004}, and it is measured among experts who chose independently, from the same candidate list, with written rationales --- not among crowd workers. The pairwise picture is the same: no two reviewers exceed 59\% exact fine agreement. When trained lexicographers disagree this often about which WordNet sense a token carries, a benchmark scored at that granularity is measuring, in part, a distinction its own annotators cannot reliably reproduce.}

\shortlong{Coarsening the inventory largely dissolves the disagreement. Re-grading the identical reviewer choices under the \glite{} many-to-one sense map (Section~\ref{sec:lexen}) raises three-way agreement to \textbf{63.1\%} and Fleiss $\kappa$ to \textbf{0.740} (\Cref{tab:t3}), and \textbf{42.7\%} of the non-unanimous fine items become unanimous once over-specified distinctions collapse (26--38\% under the other coarse inventories tested, \Cref{tab:tcoarse}: the effect is not specific to \glite{}). This is no averaging artifact: same experts, items, and answers, re-scored under a coarser equivalence on senses. The gap between 35\% fine reliability and 63\% coarse reliability is the granularity ceiling made quantitative --- a large share of fine WordNet WSD is a distinction experts make differently, not a fact about the word.}{Coarsening the inventory largely dissolves the disagreement. Re-grading the identical reviewer choices under the \glite{} many-to-one sense map (Section~\ref{sec:lexen}) raises three-way agreement to \textbf{63.1\%} and Fleiss $\kappa$ to \textbf{0.740} (\Cref{tab:t3}), and \textbf{42.7\%} of the fine items on which the reviewers were not unanimous become unanimous once the over-specified distinctions are collapsed --- 26--38\% under the other coarse inventories tested (\Cref{tab:tcoarse}), so the effect is not specific to \glite{}. The lift is not an averaging artifact: it is the same experts, the same items, the same answers, re-scored under a coarser equivalence on senses. What looked like 35\% reliability at the fine level is 63\% reliability at the level where the distinctions are practically meaningful. The gap between those two numbers is the granularity ceiling made quantitative --- a large share of fine WordNet WSD is a distinction experts make differently, not a fact about the word.}

\shortlong{\emph{Chance model.} Each item offers its own candidate set (2--49 senses), so we compute
$\kappa$ in the standard pooled-marginal form \citep{cohen1960,fleiss1971,artstein2008} over the
global space of sense selections; with lemma-specific categories the chance term is small
($p_e\le0.017$) and $\kappa$ tracks raw agreement, while coarsening mechanically raises per-item
chance (mean $1/K_i$: $0.21\to0.37$). Neither finding is an artifact of this choice: under a
free-marginal $\kappa$ \citep{brennan1981} charging each item its own candidate space, the
fine$\to$coarse rise remains large ($0.43\to0.60$) and the coarse reviewer--model gap stays
small ($+0.011$, point estimate well inside the $0.048$ margin; 95\% CI $[-0.038,+0.058]$).}{\emph{Chance model.} A caveat on the $\kappa$s in this section: every item draws its answer from
its own candidate set (2--49 senses of its lemma), so there is no category space shared across
items, and we compute Fleiss and Cohen $\kappa$ in their standard pooled-marginal forms
\citep{cohen1960,fleiss1971,artstein2008} over the global space of sense selections (each
distinct selection one category). With lemma-specific categories the pooled chance term is small
($p_e \le 0.017$ for the fine and \glite{}-coarse reviewer $\kappa$s here; the public-inventory
$\kappa$s of \Cref{tab:tcoarse}, whose categories span lemmas, carry larger chance terms), so these
$\kappa$s track raw agreement closely --- and, because coarsening shrinks each item's candidate
space (a mean of $7.7$ fine candidates collapses to $4.2$ coarse classes), part of the raw
fine$\to$coarse rise is agreement that two raters would reach by chance (mean per-item chance
$0.21$ fine vs $0.37$ coarse). Neither finding depends on this choice. Recomputing everything
under a chance model that charges each item its own candidate space --- a free-marginal $\kappa$
\citep{brennan1981} with item-specific $K_i$ --- the fine$\to$coarse rise remains large
($0.43 \to 0.60$, $+0.17$ against the $+0.20$ published), and the coarse reviewer--model gap
moves from $+0.007$ to $+0.011$ (95\% bootstrap CI $[-0.038, +0.058]$), the point estimate
remaining well inside the pre-specified $0.048$ equivalence margin (\Cref{sec:discussion-solved})
though its wider free-marginal interval is no longer fully contained by it. The granularity
ceiling and the envelope result are conclusions about agreement, not about the chance
correction.}

\journalonly{\begin{figure*}[t]
\centering
\includegraphics[width=\shortlong{\linewidth}{0.85\linewidth}]{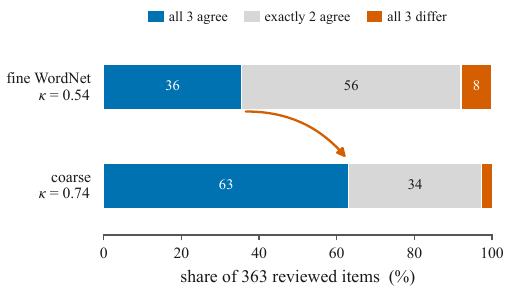}
\caption{Inter-annotator agreement over the 363 reviewed items, fine WordNet versus \glite{} coarse. Three-way exact agreement rises from 35.5\% to 63.1\% and Fleiss $\kappa$ from 0.537 to 0.740 under coarsening; 42.7\% of non-unanimous fine items become unanimous. The underlying pairwise and Fleiss figures are in \Cref{tab:t3}.}
\label{fig:f5}
\end{figure*}}

\begin{table*}[t]
\centering
\small
\caption{Inter-annotator agreement (RF/PW/PH), fine WordNet vs.\ \glite{} coarse.}
\label{tab:t3}
\begin{tabular}{llrr}
\toprule
\textbf{Granularity} & \textbf{Pair} & \textbf{Agreement (\%)} & \textbf{$\kappa$} \\
\midrule
fine & RF+PW & 57.3 & 0.567 \\
fine & RF+PH & 46.8 & 0.461 \\
fine & PW+PH & 59.0 & 0.585 \\
\glite{} & RF+PW & 75.5 & 0.751 \\
\glite{} & RF+PH & 68.9 & 0.683 \\
\glite{} & PW+PH & 79.1 & 0.788 \\
fine & Fleiss (3 raters) & 35.5 & 0.537 \\
\glite{} & Fleiss (3 raters) & 63.1 & 0.740 \\
\bottomrule
\end{tabular}
\\[2pt]\footnotesize\textit{Pairwise rows give raw exact agreement and Cohen's $\kappa$; Fleiss rows give the all-three-agree rate and the 3-rater Fleiss $\kappa$.}
\end{table*}

\journalonly{The boxes below make the mechanism concrete. Each is an item where the reviewers chose \emph{different fine WordNet senses that map to the same \glite{} concept} --- a recorded fine disagreement that the coarsening removes. We give the reviewers' own rationales verbatim and write sense keys as \sk{lemma\%pos:\allowbreak lexfile:\allowbreak lexid::}.}

\journalonly{\begin{examplebox}[E3a: \emph{evidence} --- legal vs.\ everyday ``grounds'' (NOUN)]
\textit{``\ldots{}saying that the news media were `carefully controlled' and that there was no \tw{evidence} the jury was driven by passion.''} \;(\texttt{semeval2013.d005.s009.t007})\\[3pt]
\sk{evidence\%1:09:00::} ``your basis for belief or disbelief; knowledge on which to base belief'' (chosen by RF, PW) \;vs.\;
\sk{evidence\%1:10:00::} ``(law) all the means by which an alleged matter of fact is established or disproved at judicial trial'' (chosen by PH).\\[3pt]
RF: \textit{``this use is not the legal one\ldots{} not evidence presented as part of a trial.''} PH: \textit{``the context is a court ruling\ldots{} so I've opted for the more specific (legal) sense.''}
\emph{Fine: exactly two agree $\rightarrow$ Coarse: all three agree} (all three models also chose the everyday sense). Two defended expert readings of practically the same meaning; the coarse concept removes the split.
\end{examplebox}

\begin{examplebox}[E3b: \emph{fundamental} --- fine-unscoreable, coarse-unanimous (ADJ)]
\textit{``\ldots{}studying a tragic but uncommon tumor made possible some \tw{fundamental} insights about the most basic workings of cancer\ldots{}''} \;(\texttt{senseval2.d001.s026.t007})\\[3pt]
\sk{fundamental\%5:00:00:basic:00} ``being or involving basic facts or principles'' (PH; included by RF) \;vs.\;
\sk{fundamental\%5:00:00:significant:00} ``far-reaching and thoroughgoing in effect'' (PW; included by RF).\\[3pt]
RF selected both fine senses, PW the ``far-reaching'' sense, PH the ``basic principles'' sense, so there is no single fine consensus. RF: \textit{``Both senses fit perfectly, and the context doesn't clarify which.''}
\emph{Fine: all differ $\rightarrow$ removed from fine scoring. Coarse: all three agree.} The strongest case --- \lexen{} drops the item at the fine level, yet all three reviewers agree once coarsened.
\end{examplebox}}

\journalonly{\begin{examplebox}[E3c: \emph{say} --- spoken vs.\ written reporting verb (VERB)]
\textit{``It was the confirming evidence we all needed\ldots{},'' \tw{says} Ray White at Howard Hughes Medical Institute\ldots{}} \;(\texttt{senseval2.d001.s067.t009})\\[3pt]
\sk{say\%2:32:00::} ``express in words'' and \sk{say\%2:32:15::} ``utter aloud'' (PW, PH) \;vs.\;
\sk{say\%2:32:13::} ``state as one's opinion or judgement; declare'' (RF).\\[3pt]
PW: \textit{``Unclear if reported speech or quote from something written.''} PH: \textit{``no proof that he uttered this aloud; it could have been a written statement.''}
\emph{Fine: split $\rightarrow$ Coarse: all three agree} (all three models chose \emph{express in words}). The classic reporting-verb ambiguity --- spoken or written --- is irrecoverable from the text and irrelevant once coarsened.
\end{examplebox}}

\shortlong{Worked instances appear above (\emph{evidence}, \emph{fundamental}, \emph{say}); further reviewed corrections spanning the part-of-speech range are in Appendix~B.}{Further reviewed corrections spanning the part-of-speech range are in Appendix~B.}

\subsection{Easy versus Hard}\label{sec:results-ceiling:easyhard}

\shortlong{The 94--95\% headline is a full-set number; the retained items are by construction the hard core, selected because strong models disagreed with the source label (Section~\ref{sec:lexen}). \Cref{tab:t7} reports, for each top model, accuracy on the full set, on the hard items at fine granularity, and on the hard items at \glite{} coarse granularity.}{The 94--95\% headline is a full-set number, and it is worth seeing what it averages over. The reviewed-and-retained items are, by construction, the hard core: the triage panel selected them precisely because strong models disagreed with the source label (Section~\ref{sec:lexen}). Restricting evaluation to those 307 items isolates the difficulty the full-set average dilutes. \Cref{tab:t7} and \Cref{fig:f7} report, for each top model, accuracy on the full set, on the hard items at fine granularity, and on the hard items at \glite{} coarse granularity.}

\shortlong{On the hard items fine accuracy collapses to roughly two-thirds: GPT-5.5 from 95.6 to \textbf{66.1}, Gemini-3.1-Pro from 94.9 to \textbf{64.8}, Claude-Fable-5 from 95.2 to \textbf{68.1}. In isolation this looks like a model that cannot do WSD. But the same items at coarse granularity --- where experts themselves agree 63\% of the time --- recover most of the loss: \textbf{87.0}, \textbf{86.6}, \textbf{87.3}, and the lift survives the change of inventory (78--81\% under three public coarsenings; \Cref{tab:tcoarse}). Fine hard-item accuracy is low for the same reason fine inter-annotator agreement is low: both read the over-specified tail of the inventory. The 95\% and the 66\% are not in tension --- one averages over an easy benchmark, the other scores its hardest fraction at fine granularity --- and the $\sim$88\% coarse score reconciles them.}{On the hard items the fine accuracy collapses to roughly two-thirds: GPT-5.5 falls from 95.6 to \textbf{66.1}, Gemini-3.1-Pro from 94.9 to \textbf{64.8}, Claude-Fable-5 from 95.2 to \textbf{68.1}. Read in isolation, those numbers look like a model that cannot do WSD. But the same items at the coarse granularity --- the one at which experts themselves agree 63\% of the time --- recover most of the loss: \textbf{87.0}, \textbf{86.6}, and \textbf{87.3} respectively --- and the lift survives the change of inventory, reaching 78--81\% under three public coarsenings (\Cref{tab:tcoarse}). The fine hard-item accuracy is low for the same reason the fine inter-annotator agreement is low; both are reading the over-specified tail of the inventory. The coarse hard-item accuracy is the honest measure of how often a model gets the practically meaningful sense right on the genuinely difficult items, and it sits near 88\%. The 95\% and the 66\% are not in tension: one is the average over an easy benchmark, the other is the fine-grained score on its hardest fraction, and the coarse score reconciles them.}

\shortlong{One mechanical caveat: because predictions and golds pass through the same many-to-one
map, coarsening can never reduce exact-match accuracy, so some lift is guaranteed. A
matched-granularity control --- 1{,}000 random sense maps preserving, for every lemma,
exactly the group-size multiset (hence coverage) that \glite{} induces --- shows random
merges of identical granularity yield only +9.2--9.8 points on the hard subset and +0.09
Fleiss $\kappa$, against the observed \textbf{+19.2 to +21.8} and $\mathbf{+0.20}$: every
observed rise exceeds the most extreme of the 1{,}000 draws ($>$99.9th percentile; the CSI
replication behaves identically). The coarsening merges the distinctions experts and
models actually confuse, not arbitrary granularity.}{One mechanical caveat must be dealt with before reading these rises as evidence. Because
predictions and gold labels pass through the same many-to-one map, coarsening can never
reduce exact-match accuracy, so some lift is guaranteed by construction. We therefore ran
a matched-granularity random-coarsening control: for each lemma's candidate senses we drew
1{,}000 random many-to-one maps preserving exactly the number and sizes of the sense
groups (and hence the coverage) that \glite{} induces, and re-scored the accuracy and agreement
rises reported in this section under each map. Random coarsenings of identical granularity do buy a sizable
mechanical lift --- \textbf{+9.2 to +9.8} points on the hard subset and \textbf{+0.093}
Fleiss $\kappa$ on average --- but the observed rises are roughly twice the null mean and
exceed the most extreme of the 1{,}000 draws on every quantity: \textbf{+19.2 to +21.8}
points on the hard subset against a null 97.5th percentile of +11.7 to +12.7, and a Fleiss
rise of \textbf{+0.203} against 0.127 ($>$99.9th percentile throughout; the CSI
replication behaves identically). What the coarsening merges, in other words, is not
arbitrary granularity but the specific sense distinctions that experts and models actually
confuse.}

\journalonly{\begin{figure}[t]
\centering
\includegraphics[width=\shortlong{\linewidth}{0.85\linewidth}]{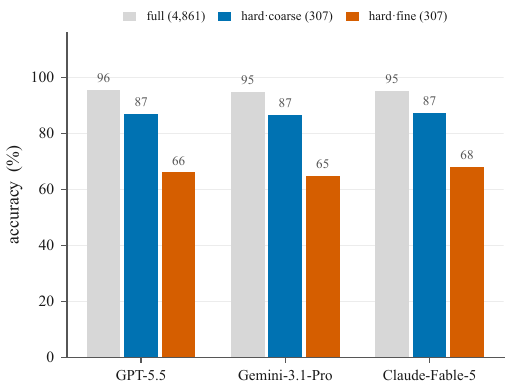}
\caption{Full-set versus hard-subset accuracy for the three frontier families. The hard subset is the 307 reviewer-adjudicated items. Fine accuracy on the hard items falls to $\sim$65\%, but coarse (\glite{}) accuracy on the \emph{same} items recovers to 87--89\% --- the fine drop tracks inventory granularity, not model failure. Numbers in \Cref{tab:t7}.}
\label{fig:f7}
\end{figure}}

\journalonly{\begin{table*}[t]
\centering
\small
\caption{Full-set vs.\ hard-subset accuracy (\%) with 95\% bootstrap confidence intervals.}
\label{tab:t7}
\begin{tabular}{lccc}
\toprule
\textbf{Model} & \textbf{Full set} & \textbf{Hard, fine} & \textbf{Hard, coarse} \\
\midrule
GPT-5.5 & 95.6 [95.0, 96.2] & 66.1 [60.9, 71.3] & 87.0 [83.1, 90.6] \\
Gemini-3.1-Pro & 94.9 [94.3, 95.5] & 64.8 [59.6, 70.0] & 86.6 [82.7, 90.2] \\
Claude-Fable-5 & 95.2 [94.6, 95.8] & 68.1 [62.9, 73.3] & 87.3 [83.4, 90.9] \\
\bottomrule
\end{tabular}
\\[2pt]\footnotesize\textit{The hard subset is the 307 reviewer-adjudicated items. Coarse accuracy uses \glite{} concepts and is always $\geq$ the fine accuracy on the same items.}
\end{table*}
}

\subsection{The Human Envelope}\label{sec:results-ceiling:envelope}

\begin{figure*}[t]
\centering
\includegraphics[width=\shortlong{\linewidth}{0.85\linewidth}]{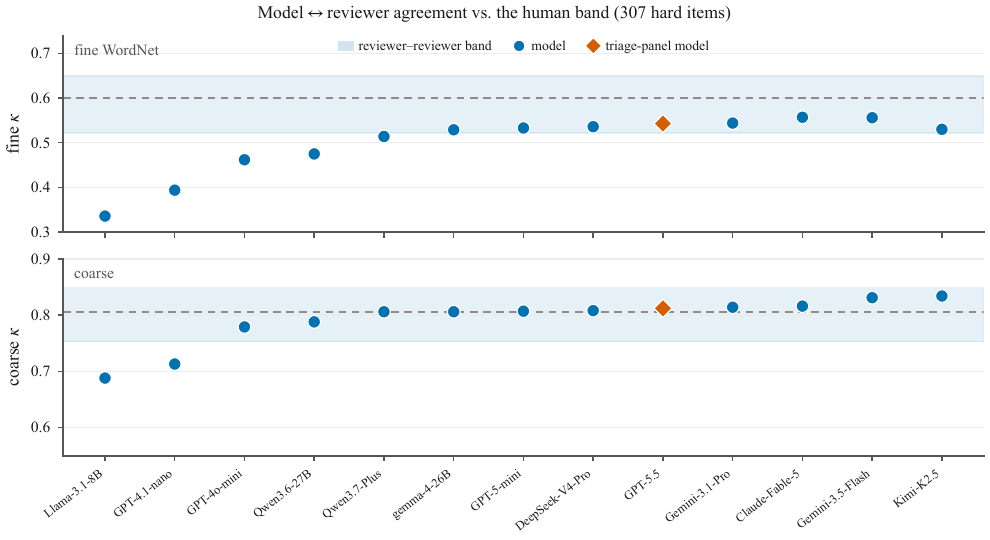}
\caption{Agreement on the 307 hard items, as mean pairwise Cohen $\kappa$, fine WordNet versus the \glite{} coarse map. At the coarse granularity, reviewer$\leftrightarrow$model $\kappa$ (0.812) sits in the same band as reviewer$\leftrightarrow$reviewer $\kappa$ (0.805), not distinguishable at this $n$; per top model the coarse $\kappa$ is 0.807/0.814/0.816. Gemini (0.814) and Claude Fable 5 (0.816) show that the pattern is not limited to the triage-family model (GPT-5.5). At fine granularity reviewer$\leftrightarrow$model $\kappa$ (0.549) is below reviewer$\leftrightarrow$reviewer (0.600). Full panel breakdown in \Cref{tab:t8}.}
\label{fig:f6}
\end{figure*}

\shortlong{A more demanding probe asks whether a model's coarse judgments fall \emph{inside} the range of disagreement the experts exhibit among themselves. This is the strongest reading of the coarse result, and we report it as coarsening-specific rather than as the headline. We measure it on the 307 reviewed-and-retained hard items as mean pairwise Cohen $\kappa$ over three pair types --- reviewer$\leftrightarrow$reviewer, reviewer$\leftrightarrow$model, model$\leftrightarrow$model --- under our inventory and under public coarse inventories (\Cref{tab:t8}, \Cref{tab:tcoarse}, \Cref{fig:f6}); it is a coarse-granularity phenomenon, absent at fine.}{If both models and experts struggle at fine granularity and recover at coarse granularity, a more demanding probe asks whether a model's coarse judgments fall \emph{inside} the range of disagreement the experts already exhibit among themselves. This is the strongest reading of the coarse result, and we treat it as coarsening-specific rather than as the headline finding. We measure it on the 307 reviewed-and-retained hard items as mean pairwise Cohen $\kappa$, comparing reviewer against reviewer, reviewer against model, and model against model, under our inventory and three public ones (\Cref{tab:t8}, \Cref{tab:tcoarse}, \Cref{fig:f6}). It holds at coarse granularity and not at fine, and we report both directions.}

\shortlong{State the harder, fine result first. At fine WordNet granularity the top frontier models do \emph{not} reach the lexicographers' band: reviewer$\leftrightarrow$model fine $\kappa$ is \textbf{0.551} (95\% bootstrap CI $[0.513,0.587]$), below the reviewer$\leftrightarrow$reviewer fine $\kappa$ of \textbf{0.600} ($[0.563,0.637]$); the paired difference is $-0.049$ ($[-0.092,-0.007]$, 10{,}000 item-resamples), excluding zero, so the model is \emph{significantly} below the lexicographers' band at fine. The envelope claim is therefore about coarse granularity only --- models enter the human envelope when, and only when, the inventory is coarsened to the level at which experts themselves agree.}{The fine result is the one to state first, because it is the harder one. At the fine WordNet granularity the top frontier models do \emph{not} reach the lexicographers' band: reviewer-versus-model fine $\kappa$ is \textbf{0.551} (95\% bootstrap CI $[0.513,0.587]$ over the 307 items), below the reviewer-versus-reviewer fine $\kappa$ of \textbf{0.600} ($[0.563,0.637]$). The paired difference, recomputed on each of 10{,}000 item-resamples, is $-0.049$ with 95\% CI $[-0.092,-0.007]$, which excludes zero: at fine granularity the model is \emph{significantly} below the lexicographers' band, not merely numerically below it. The envelope claim is therefore a claim about coarse granularity only. Models enter the human envelope when, and only when, the inventory is coarsened to the level at which the experts themselves agree.}

\shortlong{At the coarse level a top model agrees with the lexicographers about as well as they agree with one another: reviewer$\leftrightarrow$reviewer coarse $\kappa$ is \textbf{0.805} (95\% CI $[0.770,0.837]$), reviewer$\leftrightarrow$model coarse $\kappa$ is \textbf{0.812} ($[0.778,0.845]$). The paired difference is $+0.007$ ($[-0.024,+0.037]$), which includes zero: at coarse granularity the model and the reviewers are not statistically distinguishable (the interval also admits the model being slightly worse, so we read them as the same band, not the model ahead). Human-adjudicated labels and a model land in the same agreement band, a result that holds only after coarsening, since at fine granularity the model sits significantly below the reviewer band.}{At the coarse level a top model agrees with the lexicographers about as well as the lexicographers agree with one another. Reviewer-versus-reviewer coarse $\kappa$ is \textbf{0.805} (95\% bootstrap CI $[0.770,0.837]$) and reviewer-versus-model coarse $\kappa$ is \textbf{0.812} ($[0.778,0.845]$). The paired difference, recomputed on each of 10{,}000 item-resamples, is $+0.007$ with 95\% CI $[-0.024,+0.037]$, which includes zero, so at coarse granularity the model and the reviewers are not statistically distinguishable. The interval also admits the model being a couple of hundredths of a $\kappa$ point worse, so we read reviewer-versus-model and reviewer-versus-reviewer agreement as the same band rather than claiming the model is ahead. The substantive point is that human-adjudicated labels and a model land in the same agreement band, a result that holds only after coarsening, since at fine granularity the model sits significantly below the reviewer band (0.549 against 0.600, above).}

\shortlong{This is not specific to \glite{}. Re-running the same paired comparison under the three public inventories, every coarse $\Delta\kappa$ CI also includes zero --- CSI $-0.010$ ($[-0.049,+0.027]$), supersenses $-0.013$ ($[-0.049,+0.021]$), WordNet Domains $-0.044$ ($[-0.101,+0.011]$) (\Cref{tab:tcoarse}). So at coarse granularity the reviewer$\leftrightarrow$model and reviewer$\leftrightarrow$reviewer $\kappa$ are not significantly different under \emph{every} inventory we test, whereas at fine the model is significantly below. The point estimate places the model marginally inside the band under \glite{} and marginally below under the public inventories, but no coarse difference is significant: the strict ``inside the envelope'' is a \glite{}-specific reading of a result whose direction and (non-)significance are inventory-independent. \glite{} is favorable for a reason we can name --- of the coarsenings tested it is the one whose granularity most closely matches professional lexicographers (\Cref{tab:tdict}).}{This robustness is the answer to the obvious objection, that our own coarsening is what places the models inside the band. Re-running the identical paired comparison under each public inventory, every coarse $\Delta\kappa$ confidence interval also includes zero: CSI $-0.010$ ($[-0.049,+0.027]$), WordNet supersenses $-0.013$ ($[-0.049,+0.021]$), and WordNet Domains $-0.044$ ($[-0.101,+0.011]$), against \glite{}'s $+0.007$ ($[-0.024,+0.037]$) and fine WordNet's $-0.052$ ($[-0.095,-0.011]$, the only interval excluding zero; \Cref{tab:tcoarse}). The conclusion that survives the change of inventory is the statistical one: at coarse granularity the reviewer-versus-model difference is \emph{not significant} under any of the four coarsenings, whereas at fine it is significantly below. The point estimate is most favorable under \glite{}, the only inventory where it is positive, and marginally negative under the public ones; but none of the coarse differences is significant, so the strict ``models inside the human envelope'' is a \glite{}-specific reading of a result whose direction and (non-)significance hold across inventories. \glite{} is favorable for a reason we can name: of all the coarsenings tested, its granularity most closely matches that of professional lexicographers (\Cref{sec:lexen}, \Cref{tab:tdict}).}

\begin{table*}[t]
\centering
\small
\caption{Coarse-granularity robustness across sense inventories. The same fixed model predictions and the same lexicographer judgements are re-graded under our inventory and three public coarse inventories no author of this paper controls. \emph{Hard acc.} is the mean of GPT-5.5, Gemini-3.1-Pro, and Claude-Fable-5 on the 307 hard items (fine baseline 66.3\%); \emph{Fleiss $\kappa$} is reviewers-only over the 363 reviewed items. $\Delta\kappa$ is the paired reviewer$\leftrightarrow$model minus reviewer$\leftrightarrow$reviewer Cohen $\kappa$ on the 307 items (positive $=$ model \emph{inside} the human band), with a 10{,}000-resample item-bootstrap 95\% CI. Under \glite{} and CSI, every observed coarse rise exceeds the entire support of 1{,}000 matched-granularity random coarsenings (\Cref{sec:results-ceiling:easyhard}).}
\label{tab:tcoarse}
\resizebox{\ifdim\width>\linewidth \linewidth\else \width\fi}{!}{%
\begin{tabular}{lrrrrl}
\toprule
\textbf{Inventory} & \textbf{Cls/lem} & \textbf{Key cov.\%} & \textbf{Hard acc.\%} & \textbf{Fleiss $\kappa$} & \textbf{$\Delta\kappa$ (R$\leftrightarrow$M$-$R$\leftrightarrow$R) [95\% CI]} \\
\midrule
Fine WordNet (ref.) & 5.50 & 100 & 66.3 & 0.537 & $-0.052$\, [$-0.095,-0.011$]$^{\dagger}$ \\
\midrule
\glite{} (ours) & 3.34 & 95 & 87.0 & 0.740 & $+0.007$\, [$-0.024,+0.037$] \\
CSI \citep{lacerra2020} & 4.40 & 79 & 78.3 & 0.645 & $-0.010$\, [$-0.049,+0.027$] \\
WordNet supersenses & 2.86 & 100 & 80.7 & 0.683 & $-0.013$\, [$-0.049,+0.021$] \\
WordNet Domains & 2.74 & 100 & 81.2 & 0.575 & $-0.044$\, [$-0.101,+0.011$] \\
\bottomrule
\end{tabular}%
}
\\[2pt]\footnotesize\textit{Every coarse inventory lifts hard accuracy far above the 66\% fine floor and raises Fleiss $\kappa$, and the model ranking is preserved (Spearman $\rho$ 0.92--1.00 vs.\ fine): the coarsening effect is not inventory-specific. At the coarse level \emph{every} $\Delta\kappa$ CI includes zero --- the reviewer$\leftrightarrow$model and reviewer$\leftrightarrow$reviewer difference is not significant under each inventory (a cannot-reject, not a formal equivalence) --- whereas at fine the model is significantly below ($^{\dagger}$ the only CI excluding zero). The point estimate is marginally inside the band under \glite{} and marginally below under the public inventories, but no coarse difference is significant. CSI hard accuracy is the conservative uniform composite-partition grading; CSI's native set-overlap grading gives ${\sim}83\%$. \glite{} Fleiss is the released item-level map (a global one-key-one-label map gives 0.733).}
\end{table*}

\shortlong{The coarse reviewer-agreement pattern is not limited to the triage-family model. Per system, the reviewer-agreement coarse $\kappa$ is \textbf{0.807} (GPT-5.5), \textbf{0.814} (Gemini-3.1-Pro), and \textbf{0.816} (Claude-Fable-5) (\Cref{tab:t8}). GPT-5.5 selected the items, but Gemini and Claude did not --- Fable 5 was released only after the review --- and their reviewer-agreement figures sit at or above the reviewer$\leftrightarrow$reviewer 0.805. The convergence is useful corroboration on the reviewed items, not an estimate of residual errors outside the flagged tail.}{The coarse reviewer-agreement pattern is not limited to the triage-family model. Broken out by system, the per-model reviewer-agreement coarse $\kappa$ is \textbf{0.807} for GPT-5.5, \textbf{0.814} for Gemini-3.1-Pro, and \textbf{0.816} for Claude-Fable-5 (\Cref{tab:t8}). GPT-5.5 was used to select the items, but Gemini and Claude were not --- Fable 5 was released only after the review had concluded --- and their reviewer-agreement figures, Claude at 0.816 and Gemini at 0.814, both sit at or above the reviewer-versus-reviewer 0.805. The convergence is useful corroboration on the reviewed items, not an estimate of residual errors outside the flagged tail.}

\shortlong{The panel-augmentation view says the same from the other direction: adding a model to the three-reviewer panel should barely move its internal agreement if the model rates like the reviewers, and it barely does. Fleiss coarse $\kappa$ over \{RF, PW, PH\} is 0.805; adding GPT-5.5 gives 0.806, Gemini-3.1-Pro 0.809, and both 0.820 (\Cref{tab:t8}). A model joins the panel without disturbing it --- the signature of an in-distribution rater, not an outlier. At the coarse granularity that matters, a top model rates within the panel's agreement band.}{The panel-augmentation view says the same thing from the other direction. If a model rates like the reviewers, then dropping it into the three-reviewer panel should barely move the panel's internal agreement. It barely does. Fleiss coarse $\kappa$ over \{RF, PW, PH\} is 0.805; adding GPT-5.5 moves it to 0.806, adding Gemini-3.1-Pro to 0.809, and adding both to 0.820 (\Cref{tab:t8}). A model joins the panel without disturbing it --- the signature of an additional in-distribution rater rather than an outlier. At the coarse granularity that matters, a top model rates within the panel's agreement band.}

\begin{table*}[t]
\centering
\small
\caption{Agreement on the 307 hard items under fine WordNet and the \glite{} coarse map: pair-type means, per top model, and Fleiss panels. ``Model'' denotes the three frontier families (GPT-5.5, Gemini-3.1-Pro, Claude-Fable-5); reviewer$\leftrightarrow$model and model$\leftrightarrow$model are means over them. The same comparison under three public coarsenings is in \Cref{tab:tcoarse}. At the coarse level reviewer$\leftrightarrow$model $\approx$ reviewer$\leftrightarrow$reviewer.}
\label{tab:t8}
\begin{tabular}{lrr}
\toprule
\textbf{Comparison} & \textbf{fine $\kappa$} & \textbf{coarse $\kappa$} \\
\midrule
reviewer$\leftrightarrow$reviewer & 0.600 & 0.805 \\
reviewer$\leftrightarrow$model & 0.549 & 0.812 \\
model$\leftrightarrow$model & 0.788 & 0.932 \\
\midrule
GPT-5.5 & 0.544 & 0.807 \\
Gemini-3.1-Pro & 0.544 & 0.814 \\
Claude-Fable-5 & 0.557 & 0.816 \\
\midrule
Fleiss: RF,PW,PH & 0.600 & 0.805 \\
Fleiss: RF,PW,PH + gpt-5.5 & 0.572 & 0.806 \\
Fleiss: RF,PW,PH + gemini-3.1-pro & 0.572 & 0.809 \\
Fleiss: RF,PW,PH + gpt-5.5 + gemini-3.1-pro & 0.583 & 0.820 \\
\bottomrule
\end{tabular}
\end{table*}

\shortlong{One caveat travels with this result. The model$\leftrightarrow$model coarse $\kappa$ is \textbf{0.932} (\Cref{tab:t8}), far above any human-involving pair --- \emph{not} evidence the models are right but that they share biases. Frontier models trained on overlapping web-scale corpora make correlated choices, including correlated errors, so their mutual agreement reflects a common prior as much as a common truth. We therefore do not read model$\leftrightarrow$model agreement as corroborating correctness, and anchor every ``inside the envelope'' claim on the reviewer$\leftrightarrow$model comparison against human-adjudicated labels. Cross-family agreement is corroborating evidence on the reviewed items; estimating residual errors outside the flagged tail requires a random control sample.}{One caveat must travel with this result, and we state it plainly. The model-versus-model coarse $\kappa$ is \textbf{0.932} (\Cref{tab:t8}), far above any human-involving pair. That high number is \emph{not} evidence that the models are right; it is evidence that they share biases. Frontier models trained on overlapping web-scale corpora make correlated choices, including correlated errors, so their mutual agreement reflects a common prior as much as a common truth. We therefore do not read model$\leftrightarrow$model agreement as corroborating correctness, and we anchor every ``inside the envelope'' claim on the reviewer$\leftrightarrow$model comparison against human-adjudicated labels. Cross-family agreement is corroborating evidence on the reviewed items; estimating residual errors outside the flagged tail requires a random control sample.}\journalonly{ The boxes below give the texture of what the $\kappa$ values average over.}

\journalonly{\begin{examplebox}[E4a: \emph{field} --- models track the expert correction (convergence)]
\textit{``American companies walked away with stakes in just two of the 10 auctioned \tw{fields}.''} \;(\texttt{semeval2013.d003.s003.t002})\\[3pt]
\maru{}: \sk{field\%1:15:00::} ``a piece of land cleared of trees and usually enclosed'' $\rightarrow$
\lexen{}: \sk{field\%1:15:05::} ``a geographic region under which something valuable is found'' (oil field).\\[3pt]
RF, PW, PH unanimous on the correction; GPT-5.5, Gemini, and Claude all also chose the corrected sense. Here the \maru{}/\lexen{} difference does not even disappear under coarsening --- the two senses map to different \glite{} concepts --- yet all three models move \emph{with} the lexicographers and away from the wrong gold. The models are not anchored to the old label.
\end{examplebox}}

\journalonly{\begin{examplebox}[E4b: \emph{negotiation} --- a shared blind spot (all models wrong together)]
\textit{``\ldots{}Michael Zammit Cutajar\ldots{} boiled down a 180-page \tw{negotiation} text to focus on\ldots{} `the big picture.'\,''} \;(\texttt{semeval2013.d000.s007.t001})\\[3pt]
\maru{} \emph{and all three models}: \sk{negotiation\%1:10:00::} ``a discussion intended to produce an agreement'' (talks) \;vs.\;
\lexen{} (RF, PW, PH unanimous): \sk{negotiation\%1:04:00::} ``the activity or business of negotiating''.\\[3pt]
PH: \textit{``A `negotiation text' is a document produced as part of the process\ldots{} the product of the negotiation.''} All three models agree \emph{with each other} and are all wrong, sharing a plausible reading the experts rejected --- the honest counter-case to the convergence story. (At the coarse level both senses map to one \glite{} concept, so the unanimous-wrong model result is not an error there.)
\end{examplebox}}

\journalonly{\begin{examplebox}[E4c: \emph{best} --- models split where humans split (calibrated uncertainty)]
\textit{``\ldots{}the \tw{best} thing we can do is to realize that our input is being converted into MathML\ldots{}''} \;(\texttt{semeval2015.d001.s009.t003})\\[3pt]
\sk{best\%5:00:00:advisable:00} ``wiser or more advantageous and hence advisable'' --- RF, PW \emph{and GPT-5.5} (= \lexen{}-v1 label, 2-of-3) \;vs.\;
\sk{best\%3:00:00::} ``(superlative of `good') having the most positive qualities'' --- PH \emph{and Gemini, Claude}.\\[3pt]
On an item where the lexicographers themselves split 2--1, the three models split too --- and along the same fault line. Model disagreement concentrates exactly where human disagreement does. (Coarsening removes the split entirely: both senses map to one \glite{} concept.) This is the ``inside the envelope'' picture in a single item.
\end{examplebox}}

\shortlong{Worked instances of all three patterns --- model$\leftrightarrow$expert convergence (\emph{field}), a shared blind spot where all three models agree and are jointly wrong (\emph{negotiation}), and split-mirroring (\emph{best}) --- together with further such items are in Appendix~B.}{Further items --- a convergence case (\emph{country}), a shared model blind spot (\emph{underlying}), and a split the panel mirrors (\emph{local}) --- are in Appendix~B.}

\subsection{Unanswerability Typology}\label{sec:results-ceiling:typology}

\shortlong{Not every hard item has a right answer to disagree about: a non-trivial share are genuinely unanswerable, and three of the four reasons are properties of the inventory or the source text. Reviewers could decline an item with a typed reason (\Cref{tab:t10}, in \Cref{sec:app-floats}). By at least one flag, \textbf{57} items were \emph{no sense applies} (the right sense is missing from WordNet), \textbf{34} \emph{inventory inadequate} (senses do not cover the usage, often a fixed expression), and \textbf{8} \emph{input defective} (broken source sentence); \textbf{77} drew at least one cannot-answer flag. The \textbf{27} flagged by two or more reviewers are removed under the adjudication rule, alongside \textbf{29} for fine three-way disagreement, giving the \textbf{56} removed items of Section~\ref{sec:lexen}. Naming these four shapes is part of the honesty of the correction: \lexen{} removes items whose difficulty is a property of the inventory or the source.}{Not every hard item has a right answer to disagree about. A non-trivial share of the reviewed items are genuinely unanswerable, and the reasons divide cleanly. The division matters because three of the four reasons are properties of the inventory or the source text. The reviewer interface let an annotator decline an item with a typed reason; \Cref{tab:t10} reports the population. By at least one reviewer's flag, \textbf{57} items were marked \emph{no sense applies} (the right sense is missing from WordNet), \textbf{34} \emph{inventory inadequate} (the lemma's senses do not cover the usage, often a fixed expression), and \textbf{8} \emph{input defective} (the source sentence is broken). In total \textbf{77} items drew at least one cannot-answer flag; the \textbf{27} flagged by two or more reviewers are removed under the adjudication rule, alongside \textbf{29} removed for fine three-way disagreement, for the \textbf{56} removed items reported in Section~\ref{sec:lexen}.

These flags fall into four recurring shapes, and naming them is part of the honesty of the correction: \lexen{} removes items whose difficulty is a property of the inventory or the source.}\journalonly{ The four boxes below give one clean instance of each, with the reviewers' rationales verbatim.}

\journalonly{\begin{examplebox}[E2a: \emph{mute} --- missing common sense (\emph{no sense applies}, ADJ)]
\textit{``\ldots{}it would be helpful if our political leaders were \tw{mute}, rather than eloquently `concerned.'\,''} \;(\texttt{senseval2.d002.s043.t004})\\[3pt]
Only WordNet sense on offer: \sk{mute\%5:00:01:inarticulate:00} ``unable to speak because of hereditary deafness''.\\[3pt]
All three reviewers cannot-answer. RF: \textit{``Figurative sense\ldots{} `silent, not saying anything', though out of choice rather than inability.''} PW: \textit{``(of a person) choosing to be silent.''} The everyday ``deliberately silent'' sense is simply absent from WordNet; only the medical sense is listed.
\end{examplebox}

\begin{examplebox}[E2b: \emph{even} --- fixed expression (\emph{inventory inadequate}, ADV)]
\textit{``New York is\ldots{} trying to disengage itself from a 20-year-old commitment\ldots{}, \tw{even} as Chicago and other cities are moving to institute it.''} \;(\texttt{senseval2.d002.s058.t005})\\[3pt]
Only listed sense: \sk{even\%4:02:03::} ``in spite of; notwithstanding''.\\[3pt]
All three cannot-answer. RF/PH: \textit{```even as'\ldots{} means `at the very same time as'. None of the senses come close.''} PW: \textit{```even as' is a fixed phrase.''} WordNet does not list the multiword expression, so the target cannot be tagged in isolation.
\end{examplebox}}

\journalonly{\begin{examplebox}[E2c: \emph{input} --- missing technical sense (\emph{no sense applies}, NOUN)]
\textit{``Below these tabs you will find an \tw{input} field to type your functions or do your calculations.''} \;(\texttt{semeval2015.d001.s008.t002})\\[3pt]
\maru{}: \sk{input\%1:06:00::} ``a component of production''; other listed sense \sk{input\%1:10:00::} ``signal going into an electronic system''.\\[3pt]
RF, PW cannot-answer (\textit{``entering of information into a computer''}); PH chose the signal sense; no two-reviewer consensus, so the item is removed. The computing ``text-entry field'' sense is missing --- a domain-shift inventory gap.
\end{examplebox}}

\journalonly{\begin{examplebox}[E2d: \emph{points} --- defective source text (\emph{input defective}, NOUN)]
\textit{``Nicolas Sarkozy wished\ldots{} to look for missing \tw{points} of growth `with the teeth.'\,''} \;(\texttt{semeval2013.d011.s023.t002})\\[3pt]
\maru{}: \sk{point\%1:10:03::} ``a distinct item in a list''.\\[3pt]
RF: \textit{``a very non-literal translation\ldots{} The word `points' was added in translation and doesn't correspond to anything in the original French.''} PW: \textit{``intended meaning is ambiguous.''} Two cannot-answer, so removed. Not an inventory gap but a translation artifact in the source corpus --- a distinct category, kept separate as a matter of honesty.
\end{examplebox}}

\shortlong{Worked instances of each flag type --- missing common sense (\emph{mute}), inventory inadequacy (\emph{even}), missing technical sense (\emph{input}), and defective source text (\emph{points}) --- with eight further unanswerable items organized by flag type, are in Appendix~B.}{Eight further unanswerable items --- additional missing-sense cases (\emph{receipt}, \emph{smile}), inventory-inadequacy cases (\emph{cost}, \emph{have}, \emph{continental}, \emph{time}), and source-defect cases (\emph{spirit}, \emph{cycle}) --- are in Appendix~B.}

\journalonly{\begin{table}[H]
\centering
\small
\caption{Unanswerable and inventory-gap counts among the 363 reviewed items: reason flags, reviewer agreement, and removal outcome.}
\label{tab:t10}
\begin{tabular}{lr}
\toprule
\textbf{Category} & \textbf{Items} \\
\midrule
Cannot-answer reason: no sense applies & 57 \\
Cannot-answer reason: inventory inadequate & 34 \\
Cannot-answer reason: input defective & 8 \\
\midrule
Flagged cannot-answer by exactly 1 reviewer & 50 \\
Flagged cannot-answer by exactly 2 reviewers & 22 \\
Flagged cannot-answer by all 3 reviewers & 5 \\
\midrule
Removed: $\geq$2 reviewers cannot-answer & 27 \\
Removed: fine three-way no-consensus & 29 \\
\bottomrule
\end{tabular}
\\[2pt]\footnotesize\textit{Reason flags (top) count items for which at least one reviewer cited that reason; an item may carry more than one, so they do not sum to the item totals. 77 items received $\geq$1 cannot-answer flag (middle); the 27 with $\geq$2 are removed, together with 29 removed for fine three-way disagreement (56 removed in all).}
\end{table}
}

\shortlong{Taken together, the four subsections place the residual disagreement at the coarse granularity under the coarsening (\limref). Fine WordNet WSD is a distinction experts reproduce only about a third of the time on these hard items --- one on which the models agree with the experts even less than the experts agree among themselves; coarsening recovers most of the agreement and accuracy; at the coarse level the reviewer$\leftrightarrow$model and reviewer$\leftrightarrow$reviewer agreement are not significantly different under the coarsening; and the genuinely unanswerable items are those where the inventory or the source text leaves no defensible answer. The reading we carry into \Cref{sec:discussion} is two-part: at the coarse, practically meaningful granularity coarse-graining brings these LLMs close to the expert-agreement band across every inventory tested, while at fine WordNet granularity neither models nor experts agree and the question is partly ill-posed. The coarse half is reported under the specific \glite{} coarsening (\limref); it grounds the granularity-split answer to whether English WSD is solved.}{Taken together, the four subsections place the residual disagreement at the coarse granularity under the coarsening (\limref). Fine WordNet WSD is a distinction experts themselves reproduce only about a third of the time on these hard items --- one on which the models agree with the experts even less than the experts agree among themselves (reviewer$\leftrightarrow$model fine $\kappa$ below reviewer$\leftrightarrow$reviewer); coarsening recovers most of the agreement and most of the accuracy; at the coarse level the reviewer$\leftrightarrow$model and reviewer$\leftrightarrow$reviewer agreement are not significantly different under the coarsening; and the items that remain genuinely unanswerable are unanswerable because the inventory or the source text leaves no defensible answer. What the few remaining points reflect, in other words, is mostly a sense inventory finer than its own annotators can apply. The reading we carry into \Cref{sec:discussion} is deliberately two-part: at the coarse, practically meaningful granularity frontier, coarse-graining brings these LLMs close to the expert-agreement band across every inventory tested, while at fine WordNet granularity neither models nor experts agree and the question is partly ill-posed. The coarse half of that statement is reported under the specific \glite{} coarsening (\limref); it grounds the careful, granularity-split answer to whether English WSD is solved.}
 \section{Discussion}\label{sec:discussion}

\journalonly{The measurement results invite a single question, and answering it carefully is the point of this
paper. We take the question head-on, then trace its two consequences: where the real frontier has
moved, and what it takes to keep a benchmark trustworthy once the models being evaluated can read it.}

\subsection{What ``Solved'' Would Mean}\label{sec:discussion-solved}

\shortlong{Two different bottlenecks have been at work, and separating them is what makes the picture coherent.
One is \emph{label noise}: gold annotations that are simply wrong, now documented on both ends of the
pipeline---the test set (\Cref{sec:results-noise}) and the SemCor training corpus
(\Cref{sec:results-training}). It is contingent and fixable; correcting it lifts the frontier models
and, in the training data, the classic supervised systems too. The other is \emph{inventory
granularity}: WordNet draws distinctions finer than competent readers reproduce, which no relabeling
removes. \lexen{}-v1 removes a substantial slice of the fixable label noise surfaced by
model-assisted triage; the remaining hard-item disagreement is the inventory's, and it is shared by
experts and models alike.}{Two different bottlenecks have been at work in this paper, and the most useful thing we can do in
summary is separate them. The first is \emph{label noise}: gold annotations that are simply wrong,
which we have now documented on both ends of the pipeline---in the test set (\Cref{sec:results-noise})
and in the SemCor training corpus (\Cref{sec:results-training}). Label noise is contingent and
repairable; correcting it raises the measured accuracy of the frontier models and, corrected in the
training data, raises the classic supervised systems too. The second is \emph{inventory granularity}:
the WordNet inventory draws distinctions finer than competent readers, expert or model, reliably
reproduce---not a defect to be fixed by better labels, but a property of the question being asked.
\lexen{}-v1 removes a substantial slice of the fixable label noise surfaced by model-assisted triage;
the remaining hard-item disagreement is the inventory's, and it is shared by experts and models alike. The rest of this section reads the two halves of the result in
that light.}

\shortlong{Whether English WSD is solved depends on what you mean by a sense, and that dependence is the
finding. Coarse-graining substantially narrows the gap between frontier models and the experts,
robustly across four sense inventories: at the coarse, practically meaningful granularity the
difference between a top model and the expert reviewers is no longer statistically detectable. Fine-grained WordNet WSD is not at that ceiling, and the coarse
result is reported under the \glite{} coarsening (\Cref{sec:lexen}).}{Whether English WSD is solved depends entirely on what you mean by a sense, and the dependence is the
finding, so we answer in two parts rather than one. Coarse-graining substantially narrows the gap between frontier
models and the experts, robustly across four sense inventories: at the coarse, practically meaningful
granularity the difference between a top model and the expert reviewers is no longer statistically
detectable under our coarsening or the public ones. Fine-grained WordNet WSD
is not at that ceiling, and the coarse result is reported under the \glite{} coarsening
(\Cref{sec:lexen}). The two halves are not a hedge; they are the whole result, and the
granularity split is load-bearing. That ``solved'' is granularity-relative was already argued for BERT
by \citet{loureiro2021}; what is new here is that the claim now holds for \emph{frontier} models on
\emph{corrected} all-words labels, against a human ceiling we measure with professional
inter-annotator agreement.}

\shortlong{On the hard reviewed items, under the \glite{} coarsening, a top model agrees with the experts at
$\kappa\approx$0.82 --- about as well as the experts agree with one another (reviewer--reviewer
$\kappa\approx$0.805) --- Gemini and Claude match or exceed the triage-family model, and adding a
model to the panel barely moves the Fleiss coefficient (\Cref{sec:results-ceiling}). When two judges
agree as often as two experts do, the remaining disagreements are not errors awaiting a better judge;
they mark the granularity at which the question stops having a single defensible answer --- here fixed
by an inventory we author, which is why we describe the coarse result as granularity-relative rather
than solved outright. The same statistical conclusion holds under
three public coarse inventories (\Cref{sec:results-ceiling}).}{Take the coarse half first. On the hard items the lexicographers actually reviewed, under the \glite{}
coarsening, a top model agrees with the experts at
$\kappa\approx$0.82, about as well as the experts agree with one another (reviewer--reviewer
$\kappa\approx$0.805); Gemini and Claude match or exceed the model that drove triage;
and adding a model to the three-reviewer panel barely moves the Fleiss coefficient
(\Cref{sec:results-ceiling}). When two competent judges agree as often as two human experts do, the
disagreements that remain are not errors waiting for a better judge. They mark the granularity at
which the question stops having a single defensible answer --- a granularity that, here, is fixed by
an inventory we author, which is why we describe the coarse result as granularity-relative rather than
solved outright. The same
conclusion --- no statistically detectable reviewer$\leftrightarrow$model difference at coarse granularity
--- holds under three public coarse inventories as well (\Cref{sec:results-ceiling}).}

\shortlong{At fine WordNet granularity the same data tells the opposite story: the binding limit there is the
granularity of the inventory and the agreement it admits. Over the 363 reviewed items, expert fine agreement is only Fleiss $\kappa$=0.537, and roughly 43\% of non-unanimous fine
disagreements dissolve once senses are coarsened (\Cref{sec:results-ceiling}). A model in the
mid-sixties on hard fine items is not failing to understand the word; it is graded against
distinctions three professional lexicographers also decline to make consistently --- a precision the
inventory does not license.}{At fine WordNet granularity the same data tells the opposite story, and tells it about the
granularity of the inventory and the agreement it admits. Over the 363 reviewed items, expert fine agreement is only Fleiss $\kappa$=0.537, and roughly 43\% of the
non-unanimous fine disagreements dissolve the moment senses are coarsened (\Cref{sec:results-ceiling}).
A model that scores in the mid-sixties on hard fine items is not failing to understand the word; it
is being graded against distinctions that three professional lexicographers, reading the same context
with the same inventory, also decline to make consistently. The benchmark is demanding a precision
the sense inventory does not actually license. \journalonly{This is why fine and coarse accuracy on
the hard set diverge by more than twenty points for every frontier family: the gap largely marks the
slice of WordNet on which ``the right answer'' is set by the annotation guidelines as much as by the
language.}}

\shortlong{Stated carefully: where the inventory draws distinctions experts can reproduce, frontier
models reproduce them too; where it does not, the disagreement is shared by models and experts
alike. The answer to ``solved'' is therefore granularity-relative. At the coarse, practically
meaningful granularity a top model's agreement with the expert reviewers is statistically
\emph{equivalent} to the experts' agreement with one another --- any difference is smaller than a
pre-specified margin of $\delta$=0.048 in $\kappa$, half the spread that separates one expert pair
from another (two one-sided tests, $\alpha$=0.05) --- under our coarsening and under CSI and
WordNet supersenses; under WordNet Domains, the coarsest and noisiest of the four, the data support
no verdict at that margin, and at fine granularity the model sits significantly below the expert
band, where the remaining headroom coincides with distinctions the inventory leaves
under-determined. We read these as claims about English all-words WSD scored against the WordNet
inventory, and do not extend them to other languages or sense inventories.}{Stated carefully: where the inventory draws distinctions experts can reproduce, frontier
models reproduce them too; where it draws distinctions experts cannot, the disagreement is shared by
models and experts alike. The answer to ``solved'' is therefore granularity-relative. At the coarse,
practically meaningful granularity a top model's agreement with the expert reviewers is
statistically \emph{equivalent} to the experts' agreement with one another --- any difference is
smaller than a pre-specified margin of $\delta$=0.048 in $\kappa$, half the spread that separates
one expert pair from another (two one-sided tests, $\alpha$=0.05; verdicts unchanged at
$\delta$=0.05) --- under our coarsening and under two public inventories we do not control, CSI and
WordNet supersenses. Under WordNet Domains, the coarsest and noisiest of the four, the data support
no verdict at that margin (the smallest margin it passes is 0.09, and a difference of 0.07 would
have been needed for 80\% power), and at fine WordNet granularity the model sits significantly
below the expert band, where the remaining headroom coincides with distinctions the inventory
itself leaves under-determined. We read these as claims about English all-words WSD scored against
the WordNet inventory, and do not extend them to other languages or sense inventories.}

\subsection{The next frontier is cost}\label{sec:discussion-cost}

\shortlong{Accuracy stopped being the binding constraint, and price took its place. The cost--accuracy frontier
itself is measured in \Cref{sec:rn-pareto}; the point here is what it means in use. For ranking a fixed
test set the premium that buys the last few accuracy points is a small edge; for the applications that
make large-scale WSD worth doing, it is decisive, because at scale the price gap becomes a budget line.}{Accuracy stopped being the binding constraint, and price took its place. The cost--accuracy frontier
itself is measured in \Cref{sec:rn-pareto}; the point here is what that frontier means once you try to
use it. For ranking a fixed academic test set, the premium that buys the last few accuracy points is a
real, if small, edge. For the applications that make large-scale WSD worth doing, it is decisive ---
because at the scale practical WSD needs, the price gap becomes a budget line, not a footnote.}

\shortlong{Those applications operate at billions of words (\Cref{sec:rn-pareto}) --- building a dictionary,
sense-tagging a corpus, estimating which senses a learner already knows, rating the sense-level
difficulty of authentic media --- so a per-item premium
negligible on a test set becomes the difference between a feasible pipeline and an impossible one. The
question is no longer whether a top system can match human agreement on this inventory-constrained,
coarse task --- at coarse granularity several come within sampling error of the lexicographers' band --- but
whether it can do so at a price that lets you disambiguate everything; reporting cost as
a first-class leaderboard axis keeps both frontiers visible. This price frontier is also
distributional: the groups least able to pay for frontier inference --- under-resourced languages,
lower-budget institutions, public-sector deployments --- are precisely those whose lexical resources
most need the help, and the very best annotation stays rationed by compute (a claim about
English-/WordNet-scale evaluation, worse for languages without a WordNet-grade inventory).}{Those applications operate at billions of words (\Cref{sec:rn-pareto}): building or maintaining a
dictionary or sense-tagging a corpus for a lexical resource are both billions-of-words problems, and at that scale
a per-item premium that is negligible on a test set turns into the difference between a feasible
pipeline and an impossible one. The two sense-level uses that motivated this work are of exactly
this kind: estimating which senses of a word a language learner already knows, and rating the
sense-level difficulty of authentic media --- both of which mean disambiguating billions of subtitle
and corpus tokens, and both of which are economically trivial at \lens{}'s ${\sim}\$0.13$ per million
items (\Cref{sec:rt-encoder}). \journalonly{The same arithmetic governs research infrastructure:
re-scoring a benchmark under a new prompt, or sweeping an ablation across a panel of models, is cheap
with a budget model and prohibitive with a frontier one, which is itself a reason the cost axis belongs
on the leaderboard.} The practical question is no longer whether a top system can match human agreement on this
inventory-constrained, coarse task --- at coarse granularity several come within sampling error of the
lexicographers' band --- but whether it can do so at a price that lets you
disambiguate everything. Reporting cost alongside accuracy, as a first-class leaderboard axis rather
than a footnote, is how the field keeps both frontiers visible at once. There is an equity cost here
that the Pareto frame should not obscure: when the best disambiguation is compute-gated, the price
frontier is also a distributional one, and the groups least able to pay for frontier inference ---
under-resourced languages, lower-budget institutions, public-sector deployments --- are precisely those
whose lexical resources most need the help. The cheaper Pareto points partly close that gap, but the
very best annotation remains rationed by compute, and this is a claim about English-/WordNet-scale
evaluation; the equity picture for languages without a WordNet-grade inventory is worse, not better.}

\shortlong{There is a way around that price, and \Cref{sec:results-training} is its priced proof of concept. The
frontier model need not sit in the inference loop at all: run it \emph{once}, offline, to relabel a
training corpus, and its competence is distilled into the labels and, through them, into a cheap
supervised model. Relabeling all of SemCor cost \$155--\$866 \emph{once} (\Cref{sec:rt-relabel});
training \lens{} on the result cost under \$2 of GPU time; the trained model then disambiguates at
\textbf{\$0.126 per million items} at 90.5\% \lexen{} accuracy\footnote{Measured on a single
RTX~5090 at bf16 with the gloss-embedding gallery precomputed, steady-state throughput excluding model
load, GPU price at market hourly rates. The cloud rows of \Cref{tab:t4} are metered API list prices, so
the comparison is indicative rather than like-for-like.} --- roughly $85{,}000\times$ below the
frontier's \$10{,}700 per million (\Cref{tab:t4}) for the last ${\sim}5$ points of accuracy. The cost frontier is in this sense
\emph{amortizable}---the best labels can be bought once and reused---which also softens the equity
problem, since a one-time relabeling budget is far more reachable than perpetual frontier inference.}{There is, however, a way around that price, and the training-side result of \Cref{sec:results-training}
is its priced proof of concept. The frontier model need not sit in the inference loop at all. Run it
\emph{once}, offline, over a training corpus, and its competence is distilled into the corrected labels
and, through ordinary supervised training, into a model that runs for cents. The whole pipeline is now
measured end to end: relabeling the whole of SemCor cost between \$155 (open Gemma) and \$866 (GPT-5.5)
a single time (\Cref{sec:rt-relabel}); training \lens{} on the relabeled corpus cost under \$2 of GPU
time; and the trained bi-encoder disambiguates at \textbf{\$0.126 per million items} at
90.5\% \lexen{} accuracy\footnote{Measured on a single RTX~5090 at bf16 with the
gloss-embedding gallery precomputed, steady-state throughput excluding model load, and the GPU priced
at market hourly rates. The cloud rows of \Cref{tab:t4} are metered API list prices, so the comparison
is indicative rather than like-for-like.} --- roughly $85{,}000\times$ below the \$10{,}700 per million
of the frontier leader (\Cref{tab:t4}), in exchange for the last ${\sim}5$ points of accuracy. The
cheapness is architectural, not an optimization trick: the bi-encoder's gloss side is
context-independent, so the sense inventory is embedded once and inference is a single context encoding
plus dot products (\Cref{sec:rt-encoder}). The cost frontier is, in this sense, \emph{amortizable}: the
very best labels can be paid for once and reused indefinitely, and the one-time budget is no longer
hypothetical but measured. This also softens the distributional concern raised above, since a one-time
relabeling budget is far more reachable for an under-resourced group than a standing bill for frontier
inference on every item---though it does not erase the gap, as producing the relabels still requires
frontier access once.}

\subsection{Benchmark governance and contamination}\label{sec:discussion-governance}

\shortlong{Pervasive mislabeling destabilizes the very rankings a benchmark exists to produce
\citep{northcutt2021}, and a WSD benchmark in this regime faces two problems its predecessors could
ignore. First, label error: when leading systems sit within a point of one another, the few percent of
wrong gold annotations decide the order. Second, contamination: WordNet glosses, the Raganato data,
and \maru{}'s corrections are all public text a frontier model may have seen in training, and a few
memorized labels are exactly the margin that decides a one-point race. We have one empirical handle on
the most worrying form of this, memorization of the published gold: if the frontier scores came from
echoing the public Raganato and \maru{} labels, then \emph{correcting} those labels would \emph{lower}
accuracy on the changed items, as a model kept emitting the now-wrong memorized answer. Instead
accuracy rises --- on the 211 changed items GPT-5.5 gains 190 and loses 5 (\Cref{sec:rn-sensitivity}),
and the models move toward \lexen{}-v1 labels that did not exist when they were trained and so cannot have
been memorized. That is positive evidence that the agreement reflects competence convergent with the
lexicographers rather than regurgitation of the public gold; it bounds gold-memorization, though not
all memorization. Residual inherited-label noise outside the flagged tail remains a separate sampling
question. We still treat contamination as governance, not modeling: each release ships
immutable, hashed prompts and a contamination canary, and the living leaderboard pins the resolved
model version and evaluation date for every row (\Cref{sec:sensebench}). The canary catches only
verbatim echoing, so for proprietary endpoints these measures make contamination \emph{detectable} and
a run re-locatable rather than contamination-proof --- which is the most a public benchmark can
honestly promise.}{Pervasive mislabeling destabilizes the very rankings a benchmark exists to produce
\citep{northcutt2021}, and a WSD benchmark in this regime faces two such problems its predecessors
could ignore. The first is label error: when leading systems sit within a point of one another, the
few percent of gold annotations that are wrong stop being noise and start deciding the order --- the
governance case for correcting the labels before scoring against them. The second is contamination:
the systems it evaluates have
plausibly read it. WordNet glosses, the Raganato data, and \maru{}'s corrections are all public text
that a frontier model may have seen in training, and a few memorized labels are exactly the margin
that decides a one-point race. There is, however, one empirical handle on the most worrying form,
memorization of the published gold. If the frontier scores came from echoing the public Raganato and
\maru{} labels, then \emph{correcting} those labels ought to \emph{lower} accuracy on the items whose
label changed, since a memorizing model would keep emitting the now-wrong answer it had stored. The
opposite happens: accuracy rises monotonically, and on the 211 changed items the movement is
overwhelmingly toward the corrected label (GPT-5.5 gains 190 and loses 5; \Cref{sec:rn-sensitivity}),
toward \lexen{}-v1 labels that did not exist when these models were trained and so cannot have been in
their training data. The agreement therefore reflects competence convergent with the lexicographers
rather than regurgitation of the public gold. This is evidence, not proof, and it bounds memorization
\emph{of the published labels}, not memorization of the underlying correct usage. Residual inherited-label
noise outside the flagged tail remains a separate sampling question; we therefore still treat
contamination as a governance problem rather than a modeling one. Each release ships immutable, hashed
prompts and a contamination canary, so that a benchmark version is a fixed object whose integrity can
be checked rather than a moving target; the living leaderboard records the resolved model version and
evaluation date for every row, so a result is always attached to the conditions that produced it
(\Cref{sec:sensebench}). The canary detects only verbatim echoing, so for proprietary endpoints
immutability and provenance make contamination \emph{detectable} and a run re-locatable rather than
contamination-proof --- which is the most a public benchmark can honestly promise.}

\shortlong{The training-side result (\Cref{sec:results-training}) checks the same worry from an independent
direction. The relabeling that helps is scored on Raganato and Maru---test sets no model in this paper
constructed---and the systems it helps are classic, non-LLM ones; a model trained on labels that
disagree with SemCor on 21\% of instances nonetheless scores \emph{higher} on Raganato's own original
human labels, which echoing a relabeler's preferences could not achieve, and two independent relabelers
yield the same correction signal. On both the test and training ends, convergent competence, not shared
memorization, is the simplest reading.}{The training-side result (\Cref{sec:results-training}) supplies an independent check on the same worry
from the opposite end of the pipeline. There the concern is not that a model memorized the test gold
but that the retrained systems merely imitate the model that produced their training labels. The design
rules this out: the gains are measured on Raganato and Maru, test sets no model in this paper had any
part in constructing; the beneficiaries are classic non-LLM supervised systems; a system trained on
labels that disagree with the original SemCor gold on 21\% of instances nonetheless scores \emph{higher}
against Raganato's own original human annotation, which mere imitation of a relabeler could not produce;
and two independent relabelers, one proprietary and one open, yield a near-identical correction signal.
On both ends, then, the simplest explanation is convergent competence with the lexicographers rather
than shared memorization.}

\shortlong{The construction of \lexen{} suggests a reusable targeted-audit recipe for the correction step. Rather than review
4{,}917 items by hand, a panel of models triages the data, ranks items by how strongly automatic
predictions contest the source label, and routes only the suspicious minority to human adjudication
(\Cref{sec:lexen}); the triage corrected a majority of the flagged items (211 of 363),
concentrating human effort where it was needed. The division of labor keeps the
result defensible --- models propose, independent experts decide under a frozen rule, and the
\lexen{}-v1 label tracks the experts' meaning judgment, never the panel's vote. The pattern of
model-assisted triage, independent human adjudication, and a frozen two-of-three rule transfers to any
benchmark whose labels are public, whose systems are strong enough to contest them, and whose
exhaustive re-annotation is too costly to justify; a random unflagged control and highlight-free
re-review are natural extensions for estimating residual noise and anchoring.}{The construction of \lexen{} suggests a reusable targeted-audit recipe for the correction step itself. Reviewing
4{,}917 items by hand is expensive and mostly wasted on uncontroversial cases; instead a panel of
models triages the data, ranks items by how strongly automatic predictions contest the source label,
and routes only the suspicious minority to human adjudication (\Cref{sec:lexen}). The triage
corrected a majority of the flagged items (211 of 363): evidence that the model panel concentrated human
effort where it was needed rather than diluting it across the corpus. The division of labor is what keeps the result
defensible: models propose, independent experts decide under a frozen rule, and the \lexen{}-v1
label tracks the experts' meaning judgment, never the panel's vote. \journalonly{The model's role
ends before adjudication: models influence triage and candidate highlighting, but the retained reviewed
labels are set by humans.} The same
pattern --- model-assisted triage, independent human adjudication, frozen two-of-three rule --- transfers
to any benchmark whose labels are public, whose systems are strong enough to contest them, and whose
exhaustive re-annotation is too costly to justify; a random unflagged control and highlight-free
re-review are natural extensions for estimating residual noise and anchoring.}
 \section{Conclusion}\label{sec:conclusion}

\shortlong{For most of a decade English all-words WSD was a contest between models scored against labels taken
for granted; frontier systems closed that gap, and the labels became the bottleneck. \lexen{} answers
with a conservative, auditable, human-adjudicated correction layer over \maru{} and \sensebench{} with an
auditable, cost-aware leaderboard, so the field can measure systems that already agree to within a
point without measuring annotation noise instead. The noise is not the test set's alone: relabeling the SemCor training corpus and retraining the classic supervised systems unchanged lifts them by several F1 points, and the same repaired corpus trains \lens{}, a modernized 298M bi-encoder --- to our knowledge the strongest reported --- serving at ${\sim}\$0.13$ per million items (\Cref{sec:results-training}). We release both relabeled corpora, \textbf{SemCor-GPT5.5} and \textbf{SemCor-Gemma}, and the bi-encoder, so the same correction repairs the very models the field deploys. At the coarse granularity where the task is
well-posed, the best models' agreement with the experts is statistically equivalent to the experts'
agreement with one another --- any difference is smaller than the spread among the expert pairs
themselves --- under the coarsening we author and under two public inventories we do not (TOST,
margin 0.048 $\kappa$); a ceiling set by the inventory's granularity and the agreement it admits. We
draw this for English all-words WSD against WordNet, and whether it transfers elsewhere is open. The open problem has
moved: on this inventory-constrained task several systems come within sampling error of human agreement at the coarse granularity, and what separates them, across
a more than $1{,}200\times$ span in the high-accuracy slice, is price --- so the work ahead is to disambiguate at that level
everywhere it would be useful, at a cost that makes ``everywhere'' possible.}{For most of a decade, English all-words WSD was a contest between models scored against labels taken
for granted; frontier systems closed that gap, and the labels became the bottleneck.
\lexen{} answers with a conservative, auditable, human-adjudicated correction layer over \maru{}, and
\sensebench{} answers with an auditable harness and a living, cost-aware leaderboard, so that the
field can measure systems that already agree to within a point without measuring annotation noise
instead. The noise is not the test set's alone: relabeling the SemCor training corpus and retraining
the classic supervised systems unchanged lifts them by several F1 points on test sets the relabeling
never touched, and the same repaired corpus trains \lens{}, a modernized 298M bi-encoder that is, to
our knowledge, the strongest reported, serving at roughly \$0.13 per million items
(\Cref{sec:results-training}). We release both relabeled corpora, \textbf{SemCor-GPT5.5} and
\textbf{SemCor-Gemma}, and the bi-encoder, so the same correction repairs the very models the field
deploys. The picture that emerges is not a celebration of saturation. At the coarse granularity where
the task is well-posed, the best models' agreement with the experts is statistically equivalent to
the experts' agreement with one another --- any difference is smaller than the spread among the
expert pairs themselves --- under the coarsening we author and under two public inventories we do
not (two one-sided tests at a margin of 0.048 in $\kappa$); a ceiling set by the inventory's
granularity and the agreement it admits. We draw these conclusions for English all-words WSD scored against the
WordNet inventory; whether they transfer to other languages or sense inventories is an open empirical
question, not an entailment of our results. Within that scope the open problem has moved: on this inventory-constrained task several systems come within sampling error of human agreement at the coarse granularity, and
what separates them, across a more than $1{,}200\times$ span in the high-accuracy slice, is price --- a frontier that is also a
distributional one. The work ahead is to disambiguate at that level everywhere it would be useful ---
and to do so at a cost that makes ``everywhere'' possible.}
 \section*{Limitations}\label{sec:limitations}

Several constraints bound the claims above, and we state them so a reader can weigh them.

\paragraph{The reviewed subset is model-selected, not random.}
Every item a lexicographer saw was flagged by the triage panel for disagreeing with the \maru{}
label (\Cref{sec:lexen}). The reviewed set is therefore enriched for panel--\maru{} disagreement and
is not a random sample of the corpus. Two consequences follow, and we hold to both. First, we make
\textbf{no claim about a corpus-wide \maru{} error rate}: the 211 changed labels are a property of
the contested items we examined, not an extrapolation to the 4{,}554 items we did not. Second, the
unreviewed items retain their \maru{} labels unchanged; \lexen{} corrects where it looked, and
inherits \maru{}'s judgments everywhere else. Because selection is anchored to panel--\maru{}
disagreement, the reviewed set reflects that criterion, and a different selection rule could surface a different set of items.
Cross-family agreement from Gemini and Claude is useful corroboration on the reviewed items, but
frontier families can share training data, WordNet exposure, frequency priors, and semantic biases; it
is not an estimate of residual error among the unreviewed items. The nearest bound we can offer
without a second review campaign is computational: a unanimous consensus of five non-panel frontier
families disputes the inherited label on 16 of the 4{,}554 unreviewed items (0.35\%), and a
four-of-five majority on 51 (1.1\%) --- a model-estimated bound on residual disagreement, not a
human-verified error rate; only a random-sample lexicographer audit would provide that.

\paragraph{Fine WordNet granularity bounds what any of these numbers can mean.}
The agreement and accuracy figures are conditioned on the WordNet sense inventory, whose fine
distinctions experts reproduce only moderately ($\kappa$=0.537 among the three reviewers,
\Cref{sec:results-ceiling}). Where we report coarse results, they depend on the \glite{} many-to-one sense map (\Cref{sec:lexen}),
which we developed; we treat this as a competing interest and address it directly. The map is
released in full over the candidate inventory of the evaluation set (\textbf{10{,}412 sense keys} over
6{,}505 concepts --- 95\% of candidate keys and 98\% of gold keys, the few uncovered keys carrying an
explicit \texttt{unmapped} marker rather than a guess), so every coarse number is reproducible from
the released artifact. Three results turn on the coarsening --- the accuracy lift, the rise in
inter-annotator agreement, and the reviewer$\leftrightarrow$model envelope --- and we
re-derive all three under three public coarse inventories no author of this paper controls (CSI
\citep{lacerra2020}, WordNet supersenses, WordNet Domains), plus a twelve-scheme granularity sweep
released with the paper (\Cref{tab:tcoarse}, \Cref{sec:results-ceiling}). The accuracy lift and the
agreement rise hold in \emph{direction} under every inventory; the envelope conclusion is the same
\emph{statistical} one under every inventory --- at coarse granularity the reviewer-versus-model
difference is not significant under any of the four coarsenings, and the stronger equivalence
reading (two one-sided tests at the pre-specified 0.048 margin, \Cref{sec:discussion-solved}) additionally
passes under three of them, WordNet Domains remaining inconclusive rather than adverse --- and the
difference is significant only at fine. What is
\glite{}-specific is the \emph{point estimate}: reviewer$\leftrightarrow$model $\kappa$ edges above
reviewer$\leftrightarrow$reviewer $\kappa$ under \glite{} and falls between one and four hundredths below it under
the public inventories, none of these differences being significant. \glite{} is the most favorable
of the coarsenings tested because its granularity is the closest to professional lexicographers' --- it
reproduces the consensus grouping of five professional dictionaries more closely than any public
inventory (\Cref{tab:tdict}) --- not because it is fitted to the distinctions models get wrong. We
could not evaluate OntoNotes \citep{palmer2007}: it is LDC-licensed and has no public WordNet-3.0
sense-grouping map. We report the coarse claims as inventory-specific: they are claims under the
stated coarsening, with public-inventory robustness checks alongside the authored \glite{} view.

\paragraph{Reviewer heterogeneity.}
The three lexicographers do not behave identically. In particular one reviewer flags
\emph{no\_sense\_applies} at a higher rate than the others, so the unanswerability typology
(\Cref{sec:results-ceiling}) reflects a blend of genuine inventory gaps and individual strictness
about when a sense ``fits.'' The frozen two-of-three adjudication rule absorbs some of this, but the
raw per-reviewer rates are not interchangeable, and small panels cannot average reviewer
idiosyncrasy away the way a large crowd would.

\paragraph{The review interface highlighted candidate senses.}
To keep long sense lists manageable, the review tool highlighted the candidate senses
--- those an earlier annotation and the automated triage systems had judged plausible, two or more when they disagreed --- without revealing
which source proposed which, and with explicit instructions that a highlight is not a hint: the
correct sense may be unhighlighted, or the right verdict may be a cannot-answer flag
(\Cref{sec:app-brief}). Reviewers were blind to the \maru{} label, to which panel member flagged an
item, and to the provenance of each highlight, but \emph{not} to the highlights themselves, so we do
not claim the candidate set was presented without cues. Hiding provenance does not fully remove
possible anchoring from the highlights; the brief is written to counter it, but we did not measure whether
highlighted senses were chosen more often than an unbiased reviewer would choose them. A
highlight-free re-review of a sample would settle the question and is the clean test we have not yet
run.

\paragraph{We evaluate inventory-constrained WSD, not end-to-end lexical semantics.}
\sensebench{} scores the disambiguation step: it presents the target word with the full set of WordNet
candidate senses for its lemma and part of speech and asks for an index (\Cref{sec:sensebench}). This
assumes the target span is already identified and the candidate senses are supplied --- the standard
all-words WSD setting, but a narrower task than an end-to-end lexical-semantics system. It does not
require a model to detect target spans, retrieve or generate candidate senses from raw text, recognize
that the correct sense is missing from the inventory, or produce a dictionary-quality definition: the
relaxations that Word Sense Linking adds to standard WSD \citep{bejgu2024}, and that tasks such as
Word-in-Context \citep{pilehvar2019wic}, lexical substitution \citep{mccarthy2009lexsub}, and definition modeling \citep{noraset2017defmod} pursue by moving away from a fixed
inventory or by generating rather than selecting. Selecting among supplied senses is also easier than
open-ended generation --- a system that could not articulate a sense unprompted may still rank the
right gloss highest --- so our accuracy numbers upper-bound open-generation performance and are not
directly comparable to generative WSD settings. All our claims, including the coarse human-agreement
results, are about this inventory-constrained task.

\paragraph{Proprietary endpoints drift.}
Several leaderboard systems are proprietary APIs whose behavior changes without notice. We mitigate
this by pinning the resolved model version, the evaluation date, and the artifact hashes for every
reported run (\Cref{tab:t12}), so that any number can be located and re-verified, but we cannot
guarantee that re-querying the same endpoint today reproduces the same outputs. The frozen artifacts,
not the live endpoints, are the reproducible object.

\paragraph{Reasoning-effort coverage is uneven.}
Reasoning effort is a vendor-specific control, and only 8 of the 52 models run under prompt \texttt{p001} expose more
than one tier (\Cref{tab:teffort}). The headline leaderboard therefore reports each model at its own
best available tier, which mixes tiers across rows; we give a matched-tier comparison and a paired
bootstrap (\Cref{sec:results-noise}) precisely because a single best-tier number can conflate
capability with how much effort a vendor happens to offer. The per-tier curves are single runs, not
replicates, so small non-monotonicities (e.g.\ Grok 4.3 declining slightly with effort) should be read
as ``effort does not help here,'' not as a precise effect size.

\paragraph{The prompt--context ablation is narrow.}
The controlled single-variable ablation that isolates the contribution of context, synonyms, and the
other prompt components (\Cref{sec:results-noise}) was run on \textbf{two open Gemma models only}.
The leaderboard-wide \texttt{p001}-over-\texttt{p002} advantage is broad, but the decomposition of
\emph{why} --- that the 5+1 context window is the single largest driver and synonyms the best
quality-per-cost addition --- is established on those two models and should not be assumed to
transfer unchanged to larger or proprietary systems.

\paragraph{No random control sample.}
Because every reviewed item was model-flagged, this release does not estimate the residual error rate
among the 4{,}554 unreviewed items that inherit their \maru{} labels. We can report the correction
yield where models contest the label (211 of 363 items corrected), but not how often correction would
fire on items the models accept. A held-out random sample sent through the identical review protocol
would supply that base rate, and we intend to collect one.

\paragraph{SemCor relabeling is model disagreement, not a verified error rate.}
The 21\% (GPT-5.5) and 22\% (Gemma) disagreement rates we report for the SemCor training corpus
(\Cref{sec:results-training}) measure how often a frontier relabeler departs from the original gold,
not how often that gold is wrong: we did not adjudicate the relabeled training items with the
lexicographer panel. The evidence that the changes are, in aggregate, corrections is indirect---the
retrained systems improve on independent test sets---and the two worked examples (\Cref{sec:rt-relabel})
are author-judged illustrations, not three-reviewer adjudications. A held-out, lexicographer-adjudicated
sample of the relabeled training items would turn the disagreement signal into an error rate; we have
not collected one.

\paragraph{The retraining result is read from test sets the relabeling never touched.}
Of the three test sets in \Cref{tab:tretrain}, only Raganato \textsc{all} and Maru \textsc{all\_new}
are independent of the relabeling; the \lexen{} column shares a model with the relabeled training
corpus via triage, so a system trained on relabeled data and scored on \lexen{} could gain
partly from train--test alignment rather than from genuine correction. We therefore base every
training-side claim on the Raganato and Maru columns and treat \lexen{} as confirmatory. The same rule
applies with full force to \lens{} (\Cref{sec:rt-encoder}), which is trained on GPT-5.5-relabeled data:
its \lexen{} score shares a labeling function with its training corpus, and every claim we make for it
is anchored to Raganato and Maru. Relatedly, our
reproduced supervised baselines track their published scores (e.g.\ ESCHER 79.57 vs.\ 80.7 reported,
within the standard reproduction band) but are single-seed runs rather than averages over seeds.

\paragraph{Bi-encoder seed variance and recipe provenance.}
The \lens{} relabeled-training cells of \Cref{tab:tbienc} are means over three training seeds
(largest per-surface range 0.6 F1); the original-SemCor control is a single run, and the component
attribution of \Cref{tab:tbiencabl} is scored on \lexen{} fine only and is single-seed per cell,
except the pooling remove-one cell, which was remeasured in the canonical seeded run.
The \lens{} training recipe was developed with \lexen{}-fine feedback, so recipe choices may be
partially adapted to that surface --- one more reason the architecture-axis claims are anchored to
Raganato \textsc{all} and Maru \textsc{all\_new}, which played no role in development.
 \section*{Ethics Statement}

The human component of this work is the lexicographer review. The three reviewers are professional
lexicographers who participated with informed consent, were compensated at professional rates for
their work, and consented to the public release of their names, sense judgments, and written
rationales as part of the dataset; the body refers to them by initials (one reviewer, PH, is also a
co-author of this paper --- see Author Contributions below). No personal or sensitive data about the
reviewers is collected or published beyond this consented professional contribution. The source text comes from the
established Raganato \citep{raganato2017} and \maru{} \citep{maru2022} WSD evaluation corpora, which
are themselves built on publicly released SemEval and Senseval data; \lexen{} adds an auditable
correction layer and does not introduce new source documents or scrape new text. We see no
foreseeable dual-use harm: the artifact is a sense-disambiguation benchmark and an auditable label
layer, and disambiguating word senses against a public dictionary carries no plausible avenue for
misuse beyond that of the underlying lexical resources.

\paragraph{Author contributions and reviewer independence.}
\journalonly{The Glite authors designed the benchmark, built \sensebench{} and \lens{}, and ran the
computational studies. The three-lexicographer review (\Cref{sec:lexen}) was carried out by the
reviewers denoted RF, PW, and PH under a shared written brief, each reviewer blind to the \maru{}
source label, to which panel member flagged an item, and to the other reviewers' choices. Reviewer
PH (Penny Hands) is a co-author of this paper; RF and PW are external professional lexicographers
with no other role in the work. Retained labels were fixed by the frozen two-of-three adjudication
rule (\Cref{sec:lexen}), so no single reviewer --- including the co-author --- could determine a
label on their own.}\aclonly{The three-lexicographer review (\Cref{sec:lexen}) was carried out by the reviewers denoted
RF, PW, and PH under a shared written brief, each reviewer blind to the \maru{} source label, to
which panel member flagged an item, and to the other reviewers' choices. One reviewer (PH) is a
co-author of this paper; the other two are external professional lexicographers with no other role
in the work. Retained labels were fixed by the frozen two-of-three adjudication rule
(\Cref{sec:lexen}), so no single reviewer --- including the co-author --- could determine a label on
their own. Author identities and affiliations appear in the de-anonymized version.}

\paragraph{Competing interests.}
\aclonly{We disclose a competing interest that bears on how the coarse-granularity results should be
read. The coarse sense inventory used for the coarsening layer (\Cref{sec:lexen}) is one the authors
developed for a commercial language-learning product, where coarse senses are the unit of
instruction, so the coarse-granularity results are reported under our grouping rather than under a
third-party standard; the fine-granularity results rest on the public WordNet inventory. We release the coarsening map in full over the evaluation set, so every coarse
number is reproducible from the released artifacts, and to show the finding does not depend on our
grouping we re-grade the same predictions under three public coarsenings --- CSI \citep{lacerra2020},
the WordNet supersenses, and WordNet Domains --- in \Cref{sec:results-noise}. The de-anonymized version names the
inventory and states the relationship explicitly.}\journalonly{We disclose a competing interest that bears on how the coarse-granularity results should
be read. The \glite{} coarse sense inventory used for the coarsening layer (\Cref{sec:lexen}) is one
developed by the authors for Glite's language-learning product, where coarse senses are the unit of
instruction, so the coarse-granularity results are reported under our grouping rather
than under a third-party standard; the fine-granularity results rest on the public WordNet
inventory. We release the coarsening map in full over the evaluation set, so
every coarse number is reproducible from the released artifacts, and to show the finding does not
depend on our grouping we re-grade the same predictions under three public coarsenings ---
CSI \citep{lacerra2020}, the WordNet supersenses, and WordNet Domains --- in \Cref{sec:results-noise}.}

\section*{Reproducibility Statement}\label{sec:repro}

\sensebench{} is built to be re-run and re-verified rather than taken on trust. Every reported run is
pinned: the immutable registered prompt (by its \texttt{p001}/\texttt{p002} identifier), the resolved
model version, the evaluation date, and the artifact hashes that let a result be located and
re-checked from the raw API responses. \Cref{tab:t12}, \shortlong{introduced in \Cref{sec:sensebench}}{already shown in \Cref{sec:sensebench}},
gives the artifact-to-result map for four representative runs --- the top three leaderboard families
plus a Llama-3.1-8B cross-check --- linking each headline number to its run identifier, model version,
reasoning setting, and date.
\aclonly{For this anonymous submission, the immutable prompt registry, the scoring and verification
scripts, the \lexen{}-v1 dataset hash with its full artifact manifest, and anonymized raw-call
snapshots (\texttt{run.json}, \texttt{predictions.jsonl}, \texttt{calls.jsonl.gz}) for the runs in
\Cref{tab:t12} are provided as anonymized supplementary material accompanying this submission, so that
a reviewer can re-derive the reported accuracies from the raw responses rather than take their
existence on trust. The public leaderboard and the two code-and-data repositories holding the full
leaderboard (192 runs across 57 models) and the complete \lexen{}-v1 dataset --- with its per-item correction lineage and
contamination canary --- are linked in the de-anonymized version. The \lexen{} data is released under
CC-BY-NC; the \sensebench{} harness code under Apache-2.0.}\journalonly{The full leaderboard, the immutable prompt registry, the run artifacts
(\texttt{run.json}, \texttt{predictions.jsonl}, \texttt{calls.jsonl.gz}), and the \lexen{} dataset
with its per-item correction lineage and contamination canary are released through the public
leaderboard at \url{https://glitetech.github.io/sensebench/} and the two code-and-data repositories,
\url{https://github.com/GliteTech/lexen} and \url{https://github.com/GliteTech/sensebench}. The
\lexen{} data, including the \lexen{}-v1 labels and reviewer rationales, is released under CC-BY-NC; the
\sensebench{} harness code is released under Apache-2.0.} Model freeze dates and release hashes are
recorded in the artifacts so that proprietary-endpoint drift does not silently invalidate a reported
figure: the frozen artifacts, not the live endpoints, are the reproducible object.

\paragraph{Relabeled training data.}
The training-side experiments (\Cref{sec:results-training}) use two relabeled versions of the SemCor
training corpus, which we release as \textbf{SemCor-GPT5.5} (\texttt{\seqsplit{semcor-gpt55-low-p003}}) and
\textbf{SemCor-Gemma} (\texttt{\seqsplit{semcor-gemma4-31b-reasoning-p003}}), each produced with the same immutable
\texttt{p003} prompt used for evaluation. Each ships as a drop-in layer over SemCor---the original
\texttt{semcor.data.xml} unchanged, with a relabeled gold-key file---alongside a complete record
per instance (original gold key, candidate set, chosen sense, relabeling model, and prompt hash), so the
relabeling is auditable. The relabeled-label layers
are released under CC\,BY-NC. This relabeling study (\Cref{sec:results-training}) was produced with \shortlong{an in-house autonomous-research framework (named in the de-anonymized version)}{the \emph{Glite ARF} autonomous-research framework \citep{philippov2026arf}}. \aclonly{The two datasets are linked in the de-anonymized version.}\journalonly{Both are available at \url{https://github.com/GliteTech/research-semcor-relabeling}.}

\paragraph{Bi-encoder release.}
The \lens{} bi-encoder of \Cref{sec:rt-encoder} is released with its full training and inference code
and the trained checkpoints, together with the per-instance prediction
key files behind every reported cell, so each number in \Cref{tab:tbienc} can be re-derived from the
released predictions or reproduced from the released recipe. The model is also added to the
\sensebench{} leaderboard as a supervised reference baseline, scored under the same harness and
schemes as BEM, ESCHER, and ConSeC. The code is released under Apache-2.0; the trained weights under CC\,BY-NC.
\aclonly{The repository is linked in the de-anonymized version.}\journalonly{Code, weights, and predictions are available at
\url{https://github.com/GliteTech/research-semcor-relabeling} --- the complete research record of
the relabeling-and-retraining study, which also holds the reproduced BEM/ESCHER/ConSeC baselines
and their retrained checkpoints from \Cref{sec:rt-retrain}.}

\paragraph{AI-usage disclosure.}
Large language models are the object of study in this paper, and several were used as instruments
within the method: the triage panel that selected items for human review is built from model
predictions (\Cref{sec:lexen}), and models appear throughout the leaderboard (\Cref{sec:results-noise})
and agreement analyses (\Cref{sec:results-ceiling}); and two models (GPT-5.5 and Gemma 4 31B) relabeled the SemCor training
corpus for the controlled retraining study and for training the released \lens{} bi-encoder
(\Cref{sec:results-training}). AI agents also conducted research itself: the retraining study of
\Cref{sec:results-training} was executed with \shortlong{an in-house autonomous-research framework
(named in the de-anonymized version)}{the \emph{Glite ARF} autonomous-research framework
\citep{philippov2026arf}}, in which LLM coding agents implement and run experiments under automated
verification; the authors reviewed the resulting code, data, and numbers against the released
artifacts. No model set a \lexen{}-v1
label: models influenced triage and candidate highlighting, while human adjudication under the frozen
two-of-three rule determined retained reviewed labels. In preparing the manuscript, the authors used LLM-based assistance for drafting
and editing prose; all claims, numbers, and interpretations were checked by the authors against the
released artifacts, and the authors take full responsibility for the content.

\appendix
\section{Full Agreement Report}\label{sec:app-agreement}

This appendix reports the agreement breakdowns that the main text summarizes only at the aggregate level. All three views describe the same $363$ reviewed items and the same three lexicographers (RF, PW, PH), scored at two granularities: the fine WordNet sense key and the coarse \glite{} concept. Read together, they make one point concrete --- the granularity finding of \Cref{sec:results-ceiling}: expert agreement is low at the fine level and recovers sharply under coarsening, and that recovery is not an artifact of one corpus, one part of speech, or one band of polysemy.

\paragraph{By source corpus.}
\Cref{tab:t13} gives the all-three-agree rate per source. The fine rate ranges from $29.5\%$ (senseval2) to $46.2\%$ (semeval2015); the coarse rate ranges from $59.1\%$ to $73.1\%$. Every source roughly doubles its unanimity rate under coarsening, and the ordering of sources is broadly preserved across granularities. The fine rate is lowest on the two senseval corpora, whose annotation predates the later SemEval campaigns; we do not read this as a quality verdict on any one corpus, since the reviewed subset is model-enriched and small per source, but the consistency of the coarse recovery across all four sources is the load-bearing observation.

\begin{table}[t]
\centering
\small
\caption{All-three-agree rate (\%) by source corpus, fine vs.\ coarse.}
\label{tab:t13}
\begin{tabular}{lrrr}
\toprule
\textbf{Source} & \textbf{$n$} & \textbf{Fine} & \textbf{Coarse} \\
\midrule
senseval2 & 132 & 29.5 & 59.1 \\
senseval3 & 109 & 31.2 & 62.4 \\
semeval2013 & 70 & 45.7 & 64.3 \\
semeval2015 & 52 & 46.2 & 73.1 \\
\bottomrule
\end{tabular}
\end{table}

\begin{figure}[t]
\centering
\includegraphics[width=\linewidth]{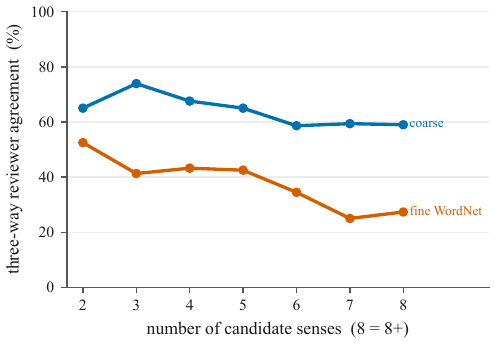}
\caption{Three-way reviewer agreement (the all-three-agree rate over RF/PW/PH) versus polysemy degree (number of candidate WordNet senses for the target), at fine WordNet and \glite{} coarse granularity. Fine agreement falls as the candidate set widens; coarse agreement stays high and roughly flat --- the reviewer-side shadow of the granularity ceiling.}
\label{fig:f9}
\end{figure}

\paragraph{By polysemy degree.}
\Cref{fig:f9} plots the three-way reviewer agreement rate against the number of candidate senses offered for the target. Fine-grained agreement falls steadily as the candidate set widens --- from \textbf{52.5\%} (Fleiss $\kappa=0.65$) on two-candidate targets to \textbf{27\%} ($\kappa=0.48$) on the most polysemous (8+) --- while agreement under the \glite{} coarsening stays high and roughly flat (59--74\%, $\kappa$ 0.70--0.80) across every bin. The fine decline is the item-level shadow of the granularity ceiling: the words carrying many closely spaced WordNet senses are exactly the ones on which the lexicographers themselves divide, and coarsening recovers agreement at every polysemy level. The low fine agreement on the hard items is therefore concentrated on high-polysemy targets rather than spread uniformly across the reviewed set.

\begin{figure}[t]
\centering
\includegraphics[width=\linewidth]{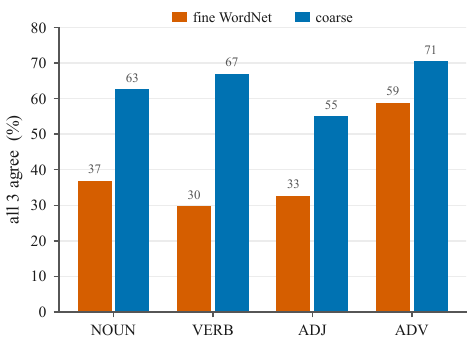}
\caption{All-three-agree rate (\%) per part of speech, fine versus coarse. Every part of speech recovers substantially under coarsening; verbs recover the most, adverbs start the highest.}
\label{fig:f10}
\end{figure}

\paragraph{By part of speech.}
\Cref{fig:f10} gives the all-three-agree rate per part of speech at both granularities. The fine rates are uniformly low and the coarse rates uniformly higher: NOUN $36.9 \rightarrow 62.6$, VERB $29.7 \rightarrow 67.0$, ADJ $32.7 \rightarrow 55.1$, ADV $58.8 \rightarrow 70.6$ (all-three-agree \%, fine $\rightarrow$ coarse). Two patterns are worth noting. Verbs show the largest absolute recovery ($+37.3$ points), consistent with WordNet's verb inventory being the most finely split and the coarse layer collapsing the most redundant distinctions there. Adverbs start highest at the fine level ($58.8\%$), reflecting their smaller and flatter sense inventories, and so have the least room to gain. In every part of speech the coarse rate clears $55\%$ while no fine rate reaches $60\%$ --- the coarsening effect is general, not carried by a single category.

Across all three cuts the message is the one \Cref{sec:results-ceiling} states in aggregate: fine-grained WordNet agreement is low for reasons that are stable across corpora, polysemy bands, and parts of speech, and the \glite{} coarsening recovers agreement everywhere it is measured. The ceiling on fine-grained scoring is a property of the sense inventory, not of any one slice of the data.
 \section{Extended Examples}\label{sec:app-examples}

These boxes extend the worked examples of \Cref{sec:lexen} and \Cref{sec:results-ceiling} with further reviewed items, in the same format and chosen to span the full range of dispositions --- corrections, items kept at the Maru2022 label, the model-agreement patterns, and the genuinely unanswerable items. Every item id resolves in \texttt{items.jsonl} (or, for removed items, \texttt{reviews.jsonl}); sense keys are written \sk{lemma\%pos:lexfile:lexid::} and are verifiable against \wn{}. Reviewer quotes are taken from \texttt{reviews.jsonl}; ``RF/PW/PH'' name the three lexicographers.

\medskip
\noindent\textit{Further corrections and retained items} (extending \Cref{sec:results-ceiling:easyhard}).

\begin{examplebox}[\emph{see} --- unanimous Maru correction (VERB)]
\textit{``\ldots{}less vomiting was \tw{seen} in dogs that received the medicine than in dogs that received a placebo\ldots{}''} \;(\texttt{semeval2015.d003.s016.t008})\\[3pt]
\sk{see\%2:39:02::} ``perceive or be contemporaneous with'' (chosen by all three) \;vs.\;
\sk{see\%2:31:03::} ``be careful or certain to do something; make certain'' (Maru2022).\\[3pt]
\emph{Fine: all three agree.} Maru's ``make certain'' reading is a clear error; the reviewers unanimously chose the plain ``perceive'' sense.
\end{examplebox}

\begin{examplebox}[\emph{solemn} --- a one-satellite shade (ADJ)]
\textit{``Ringers\ldots{} are `filled with the \tw{solemn} intoxication that comes of intricate ritual faultlessly performed.'\,''} \;(\texttt{senseval2.d000.s031.t003})\\[3pt]
\sk{solemn\%5:00:00:serious:00} ``dignified and somber in manner'' (chosen by all three) \;vs.\;
\sk{solemn\%5:00:01:serious:00} ``a firm and humorless belief in the validity of one's opinions'' (Maru2022).\\[3pt]
\emph{Fine: all three agree.} The two senses differ by a single satellite; the reviewers unanimously read ``dignified/somber,'' not the ``humorless conviction'' shade.
\end{examplebox}

\begin{examplebox}[\emph{argument} --- kept at Maru (NOUN)]
\textit{``\ldots{}court filings that sketch out appeals \tw{arguments} that are likely to occupy the courts for years.''} \;(\texttt{semeval2013.d005.s005.t004})\\[3pt]
\sk{argument\%1:10:02::} ``a fact or assertion offered as evidence'' (PW, RF; kept) \;vs.\;
\sk{argument\%1:09:01::} ``a course of reasoning'' (PH).\\[3pt]
PH: \textit{```sketch out'\ldots{} suggests they explained their line of reasoning, not simply listed assertions.''}
\emph{Fine: exactly two agree $\rightarrow$ Coarse: all three agree.} The 2-of-3 majority kept the Maru label; the item is retained unchanged.
\end{examplebox}

\medskip
\noindent\textit{Model-agreement patterns} (extending \Cref{sec:results-ceiling:envelope}).

\begin{examplebox}[\emph{country} --- model$\leftrightarrow$expert convergence (NOUN)]
\textit{``But the mood varies widely from \tw{country} to country\ldots{}''} \;(\texttt{semeval2013.d008.s019.t001})\\[3pt]
\sk{country\%1:15:00::} ``the territory occupied by a nation'' (PW, RF; and the model panel's modal pick) \;vs.\;
\sk{country\%1:14:00::} ``a politically organized body under a single government'' (PH; Maru2022).\\[3pt]
PH: \textit{``Sense 2 [territory] is also possible\ldots{}''}
\emph{Fine: exactly two agree $\rightarrow$ Coarse: all three agree.} The models tracked the expert majority to ``territory''; only PH kept the Maru ``political body'' reading.
\end{examplebox}

\begin{examplebox}[\emph{underlying} --- a shared blind spot (ADJ)]
\textit{``\ldots{}while addressing the \tw{underlying} causes of the vomiting.''} \;(\texttt{semeval2015.d003.s019.t012})\\[3pt]
\sk{underlying\%5:00:00:implicit:00} ``in the nature of something though not readily apparent'' (chosen by all three) \;vs.\;
\sk{underlying\%5:00:00:basic:00} ``being or involving basic facts or principles'' (the model panel's modal pick).\\[3pt]
\emph{Fine: all three agree (label kept).} The reviewers unanimously read ``implicit,'' but the model panel's modal choice was the ``basic'' satellite --- the models agree and are jointly off.
\end{examplebox}

\begin{examplebox}[\emph{local} --- a human split the panel mirrors (ADJ)]
\textit{``\ldots{}translate such clout into relatively more \tw{local} benefits for their respective constituencies\ldots{}''} \;(\texttt{senseval3.d001.s029.t010})\\[3pt]
\sk{local\%3:01:01::} ``of or characteristic of a particular locality'' (PH, PW) \;vs.\;
\sk{local\%3:00:01::} ``concerned with the administration of a city or town'' (RF; Maru2022).\\[3pt]
PH: \textit{``the text is not talking about local government\ldots{} but members of Congress bringing benefits back to their communities.''}
\emph{Fine: exactly two agree $\rightarrow$ Coarse: all three agree.} Reviewers split 2--1 (RF: municipal); the model panel split the same way.
\end{examplebox}

\medskip
\noindent\textit{Unanswerable items, by flag type} (extending \Cref{sec:results-ceiling:typology}).

\begin{examplebox}[\emph{receipt} --- missing common sense (\emph{no sense applies}, NOUN)]
\textit{``\ldots{}a net contribution (\tw{receipts} less expenditure) of an average of 150,000 Euros to public budgets\ldots{}''} \;(\texttt{semeval2013.d011.s022.t008})\\[3pt]
Closest listed sense: \sk{receipt\%1:10:00::} ``an acknowledgment that payment has been made''.\\[3pt]
All three cannot-answer. RF: \textit{``the amount of money received\ldots{} usually or always plural.''} PW: \textit{``money received.''} The plural ``revenue'' sense is absent from \wn{}.
\end{examplebox}

\begin{examplebox}[\emph{smile} --- missing construction (\emph{no sense applies}, VERB)]
\textit{``\ldots{}glanced at the table next to mine, \tw{smiled} that guilty smile\ldots{}''} \;(\texttt{senseval3.d002.s016.t006})\\[3pt]
Closest sense: \sk{smile\%2:32:00::} ``express with a smile''.\\[3pt]
PH/PW cannot-answer. PH: \textit{``missing: `to give a particular type of smile,' e.g.\ to smile a wry smile.''} PW: \textit{```to smile a smile' means just `to smile.'\,''} The cognate-object construction has no \wn{} sense.
\end{examplebox}

\begin{examplebox}[\emph{cost} --- inventory inadequate (VERB)]
\textit{``It won't \tw{cost} you a cent, Phil.''} \;(\texttt{senseval3.d000.s042.t000})\\[3pt]
Closest sense: \sk{cost\%2:42:01::} ``require to lose, suffer, or sacrifice''.\\[3pt]
PW/RF cannot-answer. RF: \textit{```cost nothing'\ldots{} means `to have a monetary cost, to have a price' --- not exactly this sense.''} PW: \textit{``to oblige someone to pay.''} The ``have a price'' reading is only loosely covered.
\end{examplebox}

\begin{examplebox}[\emph{have} --- inventory inadequate (VERB)]
\textit{``Now he wondered if it was worth it, \tw{having} a screwball for company.''} \;(\texttt{senseval3.d000.s029.t001})\\[3pt]
Closest sense: \sk{have\%2:40:00::} ``have or possess''.\\[3pt]
PH/PW cannot-answer. PH: \textit{``missing: `to be with someone'\ldots{} `I had a screwball for company.'\,''} PW: \textit{``the phrase is `to have sb for company.'\,''} The ``be in the company of'' sense is not in the inventory.
\end{examplebox}

\begin{examplebox}[\emph{continental} --- inventory inadequate (ADJ)]
\textit{``\ldots{}the carillons of \tw{continental} Europe\ldots{} fit only for foreigners.''} \;(\texttt{senseval2.d000.s017.t007})\\[3pt]
Closest sense: \sk{continental\%3:01:01::} ``of or typical of Europe''.\\[3pt]
PW/RF cannot-answer. RF: \textit{``this common sense\ldots{} means `excluding Britain' --- Europe seen from the British perspective.''} PW: \textit{```relating to Europe' but explicitly excluding Britain and Ireland.''} The British-perspective ``mainland Europe'' sense is not distinguished.
\end{examplebox}

\begin{examplebox}[\emph{time} --- inventory inadequate (\emph{fixed phrase}, NOUN)]
\textit{``\ldots{}the 2012 presidential race, which will be in full swing by the \tw{time} the court's decision is released.''} \;(\texttt{semeval2013.d009.s004.t008})\\[3pt]
Closest sense: \sk{time\%1:11:00::} ``an instance or single occasion for some event''.\\[3pt]
PH/RF cannot-answer. RF: \textit{```by the time' is an adverbial equivalent to `when'. None of the senses cover it.''} PH: \textit{``the text refers to a point in time, not an extended period.''} ``By the time'' is a fixed phrase no single sense covers.
\end{examplebox}

\begin{examplebox}[\emph{spirit} --- defective source text (\emph{input defective}, NOUN)]
\textit{``One of the advantages\ldots{} include the sport \tw{spirit}\ldots{} that is to say, there was no violent game.''} \;(\texttt{semeval2013.d006.s011.t002})\\[3pt]
Closest sense: \sk{spirit\%1:26:00::} ``the general atmosphere of a place or situation''.\\[3pt]
PW/PH cannot-answer. PW: \textit{``Input text is non-natural English.''} PH: \textit{``Sense 3 is close\ldots{} but the author is referring to something else.''} The source sentence is non-native English; the intended meaning is not recoverable.
\end{examplebox}

\begin{examplebox}[\emph{cycle} --- defective source text (\emph{input defective}, NOUN)]
\textit{``\ldots{}the players fulfil a sanction game and go to the second \tw{cycle} of cards\ldots{}''} \;(\texttt{semeval2013.d006.s019.t011})\\[3pt]
Closest sense: \sk{cycle\%1:11:01::} ``a single complete execution of a periodically repeated phenomenon''.\\[3pt]
PW/RF cannot-answer. PW: \textit{``non-natural English, probably written by a Spanish or French speaker (`ciclo/cycle').''} RF: \textit{``not about `an interval'\ldots{} but the specific series of events.''} From the same defective document --- a translation artifact.
\end{examplebox}

These extended items reinforce the three guards of the review. Corrections (\emph{see}, \emph{solemn}) replace a Maru2022 label the experts judged wrong, while kept items (\emph{argument}) show that an available candidate was declined when the original label fit --- which is why only 211 of the 363 flagged items were corrected. The model-agreement boxes give the texture behind \Cref{sec:results-ceiling:envelope}: convergence (\emph{country}), a shared blind spot (\emph{underlying}), and a human split the panel mirrors (\emph{local}). The unanswerable boxes show the other guard: when the inventory has no fitting sense (\emph{receipt}, \emph{smile}), when its senses do not cover a fixed usage (\emph{cost}, \emph{have}, \emph{continental}, \emph{time}), or when the source text is itself defective (\emph{spirit}, \emph{cycle}), the item leaves \lexen{} rather than being assigned a contestable \lexen{}-v1 label.
 \section{Bi-encoder Configuration and Component Attribution}\label{sec:app-biencoder}

\paragraph{Configuration.}
\lens{} (\Cref{sec:rt-encoder}) is a dual-tower gloss bi-encoder in the mould of BEM
\citep{blevins2020}, rebuilt from current parts. Two independent ModernBERT-base towers
\citep{warner2024modernbert} of 149M parameters each (298M in total): the context tower encodes the
target word in a window of five preceding sentences plus one following sentence and pools the
target's subword representations; the gloss tower encodes each candidate sense as a structured text
--- headword, part of speech, definition, synonyms, and example sentences, the same candidate
presentation the \texttt{p003} prompt gives the LLMs --- and the two are scored by a dot product.
Training uses a current contrastive recipe with in-batch negatives over the union of the batch's
candidate glosses, for three epochs over the 226{,}036 instances of SemCor-GPT5.5
(\Cref{sec:rt-relabel}). Exact hyperparameters, training scripts, and the trained checkpoints are in
the released repository (\hyperref[sec:repro]{Reproducibility Statement}); a single training run
takes about two and a half GPU-hours (under \$2 of GPU time) on one consumer GPU. The reported
numbers are means over three training seeds; the largest per-surface range is 0.6 F1
(\Cref{tab:tbienc}).

\paragraph{Component attribution.}
\Cref{tab:tbiencabl} decomposes the gap between a BEM-configured anchor and the full \lens{} recipe,
both trained on the same relabeled corpus and scored on \lexen{}-v1 fine, along six axes: each
component is either enabled alone on the anchor (add-one) or disabled from the full recipe
(remove-one). Two caveats apply: the grid is scored on \lexen{} fine only, and every cell is a
single seed except the pooling remove-one cell, which we remeasured in the canonical seeded run
(against its own seed-42 baseline of 90.6 rather than the exploratory grid's 90.0). Three components carry essentially the whole gap --- the modern backbone, the
updated training recipe, and the wider context window; their remove-one costs sum to the full
difference. The structured gloss and the in-batch negatives contribute little at the margin, and the
add-one column shows why the two readings disagree: components interact (the structured gloss
\emph{hurts} badly when added alone to the anchor), so the recipe behaves as a co-adapted stack
rather than a sum of parts. Pooling is the clearest case of a foundation rather than a
modernization: replacing target-span pooling with sentence-level CLS pooling collapses the
BERT-base anchor by nine points, and the canonical seeded remeasurement shows the full ModernBERT
recipe collapses almost as far ($90.6 \to 81.7$ on \lexen{} fine; $-6.0$ on Raganato
\textsc{all}). The exploratory single-seed grid had read ModernBERT as insensitive to the choice;
the seeded re-run refutes that reading --- target-span pooling is required in both stacks.

\begin{table*}[t]
\centering
\small
\caption{\textbf{Component attribution for \lens{} (appendix).} Add-one: the component enabled alone on the BEM-configured anchor (85.5 \lexen{} fine); remove-one: the component disabled from the full \lens{} recipe (90.0); positive remove-one means removing it costs accuracy. \lexen{} fine F1 only; single seed per cell except the pooling remove-one, the canonical seeded remeasurement ($90.6 \to 81.7$ under CLS pooling).}
\label{tab:tbiencabl}
\begin{tabular}{lrrl}
\toprule
\textbf{Component} & \textbf{Add-one $\Delta$} & \textbf{Remove-one $\Delta$} & \textbf{Reading} \\
\midrule
ModernBERT backbone (replaces BERT-base) & $+0.94$ & $+1.79$ & portable \\
Updated training recipe & $+0.80$ & $+1.25$ & portable \\
Context window (5 prev.\ + 1 next sent.) & $-0.29$ & $+1.44$ & emergent in the full stack \\
Structured gloss (synonyms + examples) & $-4.53$ & $+0.43$ & coupled; within seed noise \\
In-batch gloss-union negatives & $-0.29$ & $+0.02$ & neutral \\
Mean-target context pooling & $-9.32$ & $+8.93$ & required in both stacks \\
\bottomrule
\end{tabular}
\\[2pt]\footnotesize\textit{The gap over the BEM anchor decomposes onto the backbone, recipe, and context axes (their remove-one deltas sum to the full $+4.5$); glosses and negatives add little marginally. Pooling is different: replacing target-span pooling with sentence-level CLS pooling collapses the BERT-base anchor by $9.3$ points \emph{and} the full ModernBERT recipe by $8.9$ --- required in both stacks, not absorbed by the modern backbone. Add-one and remove-one disagree because the components interact; the recipe behaves as a co-adapted stack, not a sum of parts.}
\end{table*}

\paragraph{Serving-cost measurement.}
The \$0.126-per-million-items figure quoted in \Cref{sec:rt-encoder} and \Cref{sec:discussion-cost}
is a steady-state measurement, not an estimate: a single RTX~5090 at bf16, the gloss-embedding
gallery precomputed once for the full candidate inventory, batched inference with an asynchronous
dataloader, and model-load time excluded; the GPU is priced at market hourly rental rates. Under
these conditions the model disambiguates at roughly 0.9\,ms per item. The measurement conditions
matter for comparison: the cloud rows of \Cref{tab:t4} are metered API list prices, so cross-column
comparisons are indicative rather than like-for-like, and self-hosted throughput depends on batch
size and hardware. The structural point is hardware-independent: the gloss side of a bi-encoder is
context-free and therefore cacheable, so inference is one context encoding plus dot products
regardless of the inventory's size.
 \section{Reviewer Brief}\label{sec:app-brief}

This appendix reproduces the core of the brief given to the three lexicographers (RF, PW, PH) before they reviewed the $363$ triaged items. It is taken from the frozen, content-addressed reviewer-facing protocol\aclonly{ (SHA-256 \texttt{fb115149\ldots d82d80})}\journalonly{ {\ttfamily\seqsplit{marureview-brief-2026-05-26.md}} (SHA-256 \texttt{fb115149\ldots d82d80}), the stable reference copy of the live {\ttfamily\seqsplit{marureview.com/brief}} page}; the same brief was used for all three same-protocol reviews. We condense the worked examples and omit the per-item submission checklist for length, but reproduce the task definition and the three pieces of guidance that most shaped the labels --- granularity, noun adjuncts and phrasal verbs, and the three cannot-answer flags --- in the reviewers' own terms.

\paragraph{The task.}
Each reviewer saw $363$ English sentences with a single target word marked, and decided, from the WordNet sense inventory shown for that word, which sense the word carries in that sentence. The verdicts feed an automated WSD evaluation, which fixes one decision the reviewers make repeatedly: the granularity to work at. The brief is explicit that this is not lexicography for a learner's dictionary, where two near-identical definitions would be merged --- the reviewer identifies, as precisely as the inventory allows, which sense node the author's meaning belongs under. To keep the three reviews independent, reviewers worked independently of one another and were told not to seek out other verdicts before finishing.

\paragraph{The core rule: one sense by default.}
For each item, the reviewer selects the single sense that best fits the meaning in context, treating the sentence as written by a competent author who had one meaning in mind. In the large majority of items one sense is clearly the best fit even when others are loosely related. When two senses both seem to fit, the reviewer reasons to the better one, picks it, and records lower confidence; closeness in the inventory is not a reason to select both. Multiple senses are selected only when the sentence \emph{itself} is genuinely ambiguous --- a competent reader cannot tell which of two distinct meanings the author intended, and each is fully supported by the context --- in which case the reviewer adds a comment explaining the irreducible ambiguity. This is for true ambiguity in the text, not closeness in the inventory.

\paragraph{Highlighted candidates are not hints.}
To save reviewers from scanning long sense lists, the tool highlighted the candidate senses that other sources (an earlier annotation and the automated triage systems) had considered plausible --- two or more per item when those sources disagreed. Which source proposed which sense was deliberately not shown. The brief stresses that the highlights are a starting point for attention, not a hint that the answer is among them: the correct sense may be unhighlighted, the correct verdict may instead be a cannot-answer flag, and a lone highlight means only that the sources agreed. Reviewers were told to do their own analysis first and then check whether it landed on a highlighted sense --- either outcome being equally valid.

\paragraph{Noun adjuncts --- pick the standard noun sense.}
A noun placed before another noun to modify it (``U.S. troops'', ``water bottle'', ``government policy'') is a noun adjunct, not an adjective, and carries the same meaning as the noun used on its own. Reviewers were told to pick the standard noun sense at high confidence and \emph{not} to flag a part-of-speech mismatch. The single exception is a word that has genuinely lexicalised into a distinct adjective with a new meaning --- \emph{plastic} meaning ``fake'' or ``superficial'' rather than the material --- and only then does a separate sense apply. This rule prevents a large class of spurious cannot-answer flags on ordinary attributive nouns.

\paragraph{Phrasal verbs and multiword expressions.}
When the target is part of a phrasal verb or fixed expression that WordNet does not list as a separate lexical unit --- \emph{keep} in ``she kept back the truth'', or \emph{even} in the fixed phrase ``even as'' --- the plain senses of the headword do not cover the usage, and the reviewer flags \texttt{inventory\_inadequate} with a note identifying the expression. These are inventory gaps, not reviewer indecision, and they are a recurring reason items leave \lexen{} (see \Cref{sec:results-ceiling} and \Cref{sec:app-examples}).

\paragraph{The three cannot-answer flags.}
When a reviewer cannot pick a sense, the reason matters, and the brief requires the correct one of three flags, each with a free-text note rather than a single catch-all:
\begin{itemize}\setlength{\itemsep}{1pt}
  \item \texttt{no\_sense\_applies} --- the inventory is complete enough but none of its senses matches this usage; the meaning is clear, WordNet simply lacks it. The reviewer notes what the correct sense \emph{would} be.
  \item \texttt{inventory\_inadequate} --- WordNet has roughly the right area but cannot express the distinction needed: the word is used in a part of speech no listed sense allows, the definitions are too obscure to choose between, or the target is part of an unlisted phrasal verb. The reviewer notes the specific inadequacy.
  \item \texttt{input\_defective} --- the problem is the source sentence, not the senses: it is ungrammatical, appears machine-translated, or is missing the context needed to disambiguate. The reviewer notes what is wrong.
\end{itemize}
The flag and a best-effort sense pick are not mutually exclusive: if a reviewer can still make a guarded pick despite a flagged problem, the brief asks them to do so, set the flag, and lower their confidence. These three reasons are the typology reported in \Cref{sec:results-ceiling} and decide which items \lexen{} removes.

\paragraph{Confidence and practical notes.}
Every verdict carries a confidence rating --- high (one sense clearly fits), medium (a best answer with a defensible competitor), or low (close to guessing, or a flagged problem) --- on the principle that a low-confidence verdict is an honest verdict, not a failure. Reviewers were reminded that WordNet definitions are sometimes terse and that the example sentences WordNet supplies for each sense often disambiguate better than the definition text, to use the full surrounding context rather than the target sentence alone, and that a one-line comment is welcome on any verdict whenever the reasoning is non-obvious.

\medskip
\noindent\textit{Caveat preserved from the source protocol: the brief asks reviewers to set a confidence rating, but the exported verdict schema does not persist a \texttt{confidence} field. The released review records preserve the fields the tool actually exported --- selected sense keys, typed cannot-answer flags, cannot-answer notes, and free-text comments.}
 \aclonly{\section{Additional Tables and Figures}\label{sec:app-floats}

\begin{table}[H]
\centering
\small
\caption{Dataset composition of lexEN-v1.}
\label{tab:t1}
\begin{tabular}{lr}
\toprule
\textbf{Quantity} & \textbf{Count} \\
\midrule
Maru2022 source instances & 4917 \\
lexEN-v1 retained & 4861 \\
Unreviewed, kept with Maru2022 label & 4554 \\
Model-triaged (reviewed) & 363 \\
— retained, three-way exact agreement & 124 \\
— retained, two-of-three agreement & 183 \\
— removed, $\geq$2 reviewers cannot-answer & 27 \\
— removed, three-way no-consensus & 29 \\
Gold labels changed (of 307 retained) & 211 \\
\bottomrule
\end{tabular}
\\[2pt]\footnotesize\textit{The 363 reviewed items split into 307 retained (124 three-way + 183 two-of-three agreement) and 56 removed (27 with $\geq$2 reviewers cannot-answer + 29 three-way disagreement); 211 of the 307 retained had their gold corrected. The other 4{,}554 source items were kept with the Maru2022 label, unreviewed.}
\end{table}

\begin{table}[H]
\centering
\small
\caption{Unanswerable and inventory-gap counts among the 363 reviewed items: reason flags, reviewer agreement, and removal outcome.}
\label{tab:t10}
\begin{tabular}{lr}
\toprule
\textbf{Category} & \textbf{Items} \\
\midrule
Cannot-answer reason: no sense applies & 57 \\
Cannot-answer reason: inventory inadequate & 34 \\
Cannot-answer reason: input defective & 8 \\
\midrule
Flagged cannot-answer by exactly 1 reviewer & 50 \\
Flagged cannot-answer by exactly 2 reviewers & 22 \\
Flagged cannot-answer by all 3 reviewers & 5 \\
\midrule
Removed: $\geq$2 reviewers cannot-answer & 27 \\
Removed: fine three-way no-consensus & 29 \\
\bottomrule
\end{tabular}
\\[2pt]\footnotesize\textit{Reason flags (top) count items for which at least one reviewer cited that reason; an item may carry more than one, so they do not sum to the item totals. 77 items received $\geq$1 cannot-answer flag (middle); the 27 with $\geq$2 are removed, together with 29 removed for fine three-way disagreement (56 removed in all).}
\end{table}

\begin{figure}[t]
\centering
\includegraphics[width=\shortlong{\linewidth}{0.85\linewidth}]{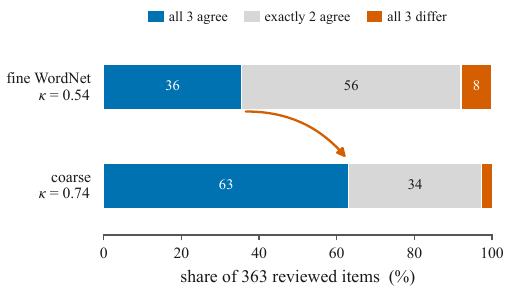}
\caption{Inter-annotator agreement over the 363 reviewed items, fine WordNet versus \glite{} coarse. Three-way exact agreement rises from 35.5\% to 63.1\% and Fleiss $\kappa$ from 0.537 to 0.740 under coarsening; 42.7\% of non-unanimous fine items become unanimous. The underlying pairwise and Fleiss figures are in \Cref{tab:t3}.}
\label{fig:f5}
\end{figure}

\begin{table*}[t]
\centering
\small
\caption{Label-noise triple-score at the \glite{} \emph{coarse}-concept level (companion to Table~\ref{tab:t6}): the same fixed predictions on the same 4{,}861 items, now comparing coarse concepts instead of fine sense keys. The three $\Delta$ columns are as in Table~\ref{tab:t6}.}
\label{tab:t6b}
\begin{tabular}{lrrrrrr}
\toprule
\textbf{System} & \textbf{Raganato} & \textbf{Maru2022} & \textbf{lexEN} & \textbf{$\Delta$\,R$\to$M} & \textbf{$\Delta$\,M$\to$L} & \textbf{$\Delta$\,R$\to$L} \\
\midrule
GPT-5.5 & 94.7 & 97.4 & 98.7 & +2.7 & +1.3 & +4.0 \\
Gemini-3.1-Pro & 94.8 & 97.6 & 98.6 & +2.8 & +1.0 & +3.8 \\
Claude-Fable-5 & 94.8 & 97.6 & 98.7 & +2.8 & +1.1 & +3.9 \\
GPT-5-mini & 93.5 & 96.1 & 97.0 & +2.6 & +0.9 & +3.5 \\
gemma-4-26B & 92.5 & 95.0 & 95.6 & +2.5 & +0.6 & +3.1 \\
GPT-4o-mini & 89.4 & 91.7 & 92.3 & +2.3 & +0.6 & +2.9 \\
\midrule
ConSeC & 93.3 & 93.9 & 93.9 & +0.6 & +0.0 & +0.6 \\
ESCHER & 90.6 & 91.2 & 91.1 & +0.6 & -0.1 & +0.5 \\
BEM & 90.5 & 91.4 & 91.0 & +0.9 & -0.4 & +0.5 \\
SANDWiCH & 94.9 & 93.9 & 93.4 & -1.0 & -0.5 & -1.5 \\
MFS & 75.2 & 77.0 & 76.6 & +1.8 & -0.4 & +1.4 \\
\bottomrule
\end{tabular}
\\[2pt]\footnotesize\textit{Coarsening collapses exactly the fine WordNet senses lexEN corrects, so accuracies rise to 92--98\% and the label-noise gaps shrink: the lexEN$-$Raganato gain falls from up to $+10$ points (fine) to about $+4$ (coarse).}
\end{table*}

\begin{table*}[t]
\centering
\small
\caption{Full-set vs.\ hard-subset accuracy (\%) with 95\% bootstrap confidence intervals.}
\label{tab:t7}
\begin{tabular}{lccc}
\toprule
\textbf{Model} & \textbf{Full set} & \textbf{Hard, fine} & \textbf{Hard, coarse} \\
\midrule
GPT-5.5 & 95.6 [95.0, 96.2] & 66.1 [60.9, 71.3] & 87.0 [83.1, 90.6] \\
Gemini-3.1-Pro & 94.9 [94.3, 95.5] & 64.8 [59.6, 70.0] & 86.6 [82.7, 90.2] \\
Claude-Fable-5 & 95.2 [94.6, 95.8] & 68.1 [62.9, 73.3] & 87.3 [83.4, 90.9] \\
\bottomrule
\end{tabular}
\\[2pt]\footnotesize\textit{The hard subset is the 307 reviewer-adjudicated items. Coarse accuracy uses \glite{} concepts and is always $\geq$ the fine accuracy on the same items.}
\end{table*}

\begin{table*}[t]
\centering
\small
\caption{Reproducibility map: a representative p001 run per top-three system (plus Llama-3.1-8B as a harness cross-check). These are the exact runs shipped in the supplementary package, so accuracy re-derives from released artifacts; the GPT-5.5 row is its medium-effort run (the xhigh run is the Table~\ref{tab:t4} champion). Run-ids are shown in full; all four use prompt p001 on the lexEN-v1 dataset (4{,}861 items, content-hash \texttt{sha256:5fd4382b}\ldots) and were executed on 2026-06-14. Each run.json additionally records the git commit, sampling and decoding policy, token usage, and cost.}
\label{tab:t12}
\resizebox{\linewidth}{!}{%
\begin{tabular}{llll}
\toprule
\textbf{Run ID} & \textbf{Resolved model} & \textbf{Reasoning} & \textbf{Date} \\
\midrule
{\scriptsize\ttfamily gpt-5.5-medium-reasoning-p001-lexen-v1-20260614} & {\scriptsize\ttfamily gpt-5.5-2026-04-23} & medium & 2026-06-14 \\
{\scriptsize\ttfamily gemini-3.1-pro-high-reasoning-p001-lexen-v1-20260614} & {\scriptsize\ttfamily gemini-3.1-pro-preview} & high & 2026-06-14 \\
{\scriptsize\ttfamily claude-fable-5-xhigh-reasoning-p001-lexen-v1-20260701-fallback-claude-opus-4-8} & {\scriptsize\ttfamily claude-fable-5} & xhigh & 2026-07-01 \\
{\scriptsize\ttfamily vllm-llama-3.1-8b-bf16-a100-p001-lexen-v1-20260614} & {\scriptsize\ttfamily meta-llama/Llama-3.1-8B-Instruct} & n/a & 2026-06-14 \\
\bottomrule
\end{tabular}
}
\end{table*}

} 
\end{document}